\documentclass{article} % For LaTeX2e
\usepackage{icrl3026_conference,times}
\usepackage{tabularx}

\usepackage{amsmath,amsfonts,bm}

\def\eqref#1{equation~\ref{#1}}
\def\1{\bm{1}}

\DeclareMathAlphabet{\mathsfit}{\encodingdefault}{\sfdefault}{m}{sl}
\SetMathAlphabet{\mathsfit}{bold}{\encodingdefault}{\sfdefault}{bx}{n}

\usepackage{hyperref}      % For clickable links and references (use default style)
\usepackage{url}           % For formatting URLs properly
\usepackage{graphicx}      % For including images (\includegraphics)
\usepackage{booktabs}      % For professional tables (\toprule, \midrule, etc.) <-- 已更正
\usepackage{amsmath}       % For advanced math environments
\usepackage{amssymb}       % For various math symbols
\usepackage{amsfonts}      % For math fonts like \mathbb
\usepackage{array}         % For advanced column types in tables (e.g., m{}, p{})
\usepackage{multirow}
\usepackage{subcaption}
\usepackage{wrapfig}%
\usepackage{siunitx} %
\usepackage{float}
\usepackage{fancyhdr}

\makeatletter
\def\@maketitle{\vbox{\hsize\textwidth
{\LARGE\sc \@title\par}
\ificlrfinal
    \lhead{Preprint.}
    \def\And{\end{tabular}\hfil\linebreak[0]\hfil
            \begin{tabular}[t]{l}\bf\rule{\z@}{24pt}\ignorespaces}%
    \def\AND{\end{tabular}\hfil\linebreak[4]\hfil
            \begin{tabular}[t]{l}\bf\rule{\z@}{24pt}\ignorespaces}%
    \begin{tabular}[t]{l}\bf\rule{\z@}{24pt}\@author\end{tabular}%
\else
    \lhead{Preprint.}
    \def\And{\end{tabular}\hfil\linebreak[0]\hfil
            \begin{tabular}[t]{l}\bf\rule{\z@}{24pt}\ignorespaces}%
    \def\AND{\end{tabular}\hfil\linebreak[4]\hfil
            \begin{tabular}[t]{l}\bf\rule{\z@}{24pt}\ignorespaces}%
    \begin{tabular}[t]{l}\bf\rule{\z@}{24pt}Anonymous authors\\Paper under double-blind review\end{tabular}%
\fi
\vskip 0.3in minus 0.1in}}
\makeatother

\title{OmniPlay: Benchmarking Omni-Modal Models on Omni-Modal Game Playing}

\iclrfinalcopy

\author{
\textbf{Fuqing Bie}\textsuperscript{1},
\textbf{Shiyu Huang}\textsuperscript{2},
\textbf{Xijia Tao}\textsuperscript{3},
\textbf{Zhiqin Fang}\textsuperscript{1}, 
\textbf{Leyi Pan}\textsuperscript{4}, 
\textbf{Junzhe Chen}\textsuperscript{4}, \\
\textbf{Min Ren}\textsuperscript{1}, 
\textbf{Liuyu Xiang}\textsuperscript{1$\dagger$}, 
\textbf{Zhaofeng He}\textsuperscript{1$\dagger$} \\
\textsuperscript{1}Beijing University of Posts and Telecommunications~~~
\textsuperscript{2}XPENG~~~\\
\textsuperscript{3}The University of Hong Kong~~~
\textsuperscript{4}Tsinghua University
}

\begin{document}

% 为保证所有页面样式统一，并显示横线和页码
\pagestyle{fancy}
\fancyhf{} % 清空所有旧的页眉页脚
\renewcommand{\headrulewidth}{0.4pt} % 设置横线
\fancyhead[L]{Preprint.} % 设置页眉文本
\fancyfoot[C]{\thepage} % 设置页脚页码

\maketitle

% 强制第一页也使用我们上面定义的 fancy 样式
\thispagestyle{fancy}

{
\let\thefootnote\relax\footnotetext{
\begin{tabular}{l}
$^\dagger$Corresponding authors.
\end{tabular}
}}

\begin{abstract}
While generalist foundation models like Gemini and GPT-4o demonstrate impressive multi-modal competence, existing evaluations fail to test their intelligence in dynamic, interactive worlds. Static benchmarks usually lack agency, while interactive benchmarks typically ignore crucial auditory and temporal cues. To bridge this evaluation chasm, we introduce OmniPlay, a diagnostic benchmark designed not just to evaluate, but to probe the fusion and reasoning capabilities of agentic models across the full sensory spectrum. Built on a core philosophy of modality interplay, OmniPlay comprises a suite of five game environments that systematically create scenarios of both model complementarity and conflict, forcing agents to perform genuine cross-modal reasoning. Our comprehensive evaluation of six leading omni-modal models reveals a critical dichotomy: they exhibit superhuman performance on high-fidelity memory tasks but suffer from systemic failures in challenges requiring robust reasoning and strategic planning. This fragility manifests as catastrophic performance degradation under modality conflict. We further uncover a counter-intuitive ``less is more'' phenomenon, where removing sensory information can paradoxically improve performance. Our findings suggest that the path toward robust AGI requires a research focus beyond scaling to explicitly address synergistic fusion. Our platform is available for anonymous review at \url{https://github.com/fuqingbie/omni-game-benchmark}.
\end{abstract}

\section{Introduction}

\begin{figure*}[t!]
    \centering
    \includegraphics[width=\textwidth]{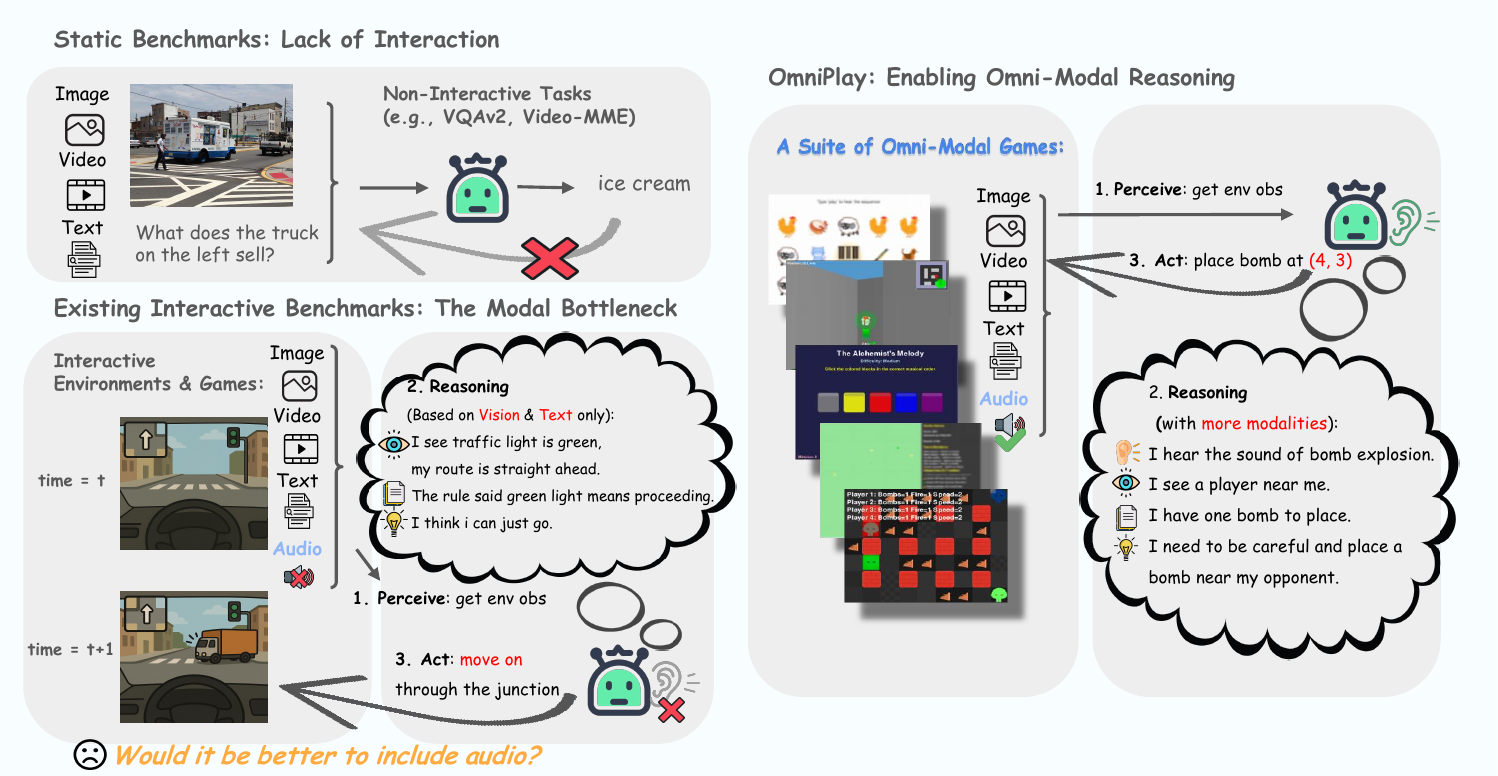}
    \caption{An illustration of the core motivation for OmniPlay. \textbf{Left:} Prior benchmarks suffer from two key limitations. Static benchmarks (e.g., VQA) lack interaction and agency. Existing interactive benchmarks are often limited to vision and text, ignoring critical modalities like audio, which could lead to decision-making failures. \textbf{Right:} OmniPlay addresses these gaps by providing an interactive, omni-modal environment. It enables agents to perform synergistic reasoning by integrating information across all modalities (e.g., combining visual, auditory, and textual cues) for more robust decision-making.}
    \label{fig:motivation}
\end{figure*}

Generalist foundation models such as Google’s Gemini \citet{team2023gemini} and OpenAI’s GPT-4o \citet{openai2024gpt4o} have recently accelerated progress toward Artificial General Intelligence (AGI). These models process text, image, audio, and video with impressive competence. However, a model's intelligence is best measured not only by its passive perception of static data, but also by its ability to reason and act through interactive decision-making in a dynamic, sensorially rich world. This raises a critical question: how can we effectively evaluate a model's ability to integrate multi-modal understanding with real-world interaction?

We argue that the existing evaluation landscape is split by a fundamental gap. On one side, many prominent multimodal benchmarks remain static, testing passive understanding in formats like VQA \citet{antol2015vqa}, MMBench \citet{liu2024mmbench}, and SEED-Bench \citet{li2023seedbench}. These lack crucial dimensions of agency and long-term planning. On the other side, a second wave of benchmarks has shifted toward interactive environments such as ALFWorld \citet{ALFWorld20} and WebArena \citet{zhou2023webarena}. While this move towards agency is vital, the majority of these interactive agents operate with limited modalities, typically confined to vision-language inputs, which restricts their ability to process auditory or complex temporal cues.

This paper argues that integrating the complete spectrum of modalities is a foundational requirement for robust omni-modal agency: an agent's capacity to perceive, reason, and make decisions by fluidly integrating inputs across all senses. The core challenge lies in managing the complexities of modality interplay. On one hand, sensory inputs can be complementary, where one modality compensates for the limitations of another — for instance, using audio cues to navigate when vision is occluded. Leveraging this complementarity is crucial for effective decision-making. On the other hand, inputs across modalities can be conflicting. For instance, receiving contradictory visual and auditory commands will create ambiguity that degrades performance. This critical capability to synergize complementary information and resolve sensory conflicts remains largely underexplored by current methodologies.

To diagnose these foundational weaknesses, we introduce \textbf{OmniPlay}, a benchmark designed not just to evaluate, but to diagnose the omni-modal fusion and reasoning capabilities of agentic models, as illustrated in Figure~\ref{fig:motivation}. OmniPlay is built upon a core philosophy of modality interplay. Across a suite of five distinct games, we develop scenarios that require the synergistic fusion of varying modality combinations (e.g., image-audio-text, video-audio). By systematically manipulating modality complementarity and conflict, OmniPlay functions as a diagnostic tool to answer critical questions: Can the model resolve contradictory inputs, or does it fail silently? Does it exhibit a bias toward a specific modality? Does richer sensory input act as a catalyst for reasoning, or does it instead introduce new failure modes?

Our primary contributions are:

\begin{enumerate}
    \item We introduce OmniPlay, the first interactive benchmark designed to diagnose an agent's synergistic fusion, conflict resolution, and adaptive reasoning under controlled modality interplay across the full sensory spectrum.
    \item We design a suite of five games built on the principle of modality interplay, systematically creating scenarios of complementarity and conflict to reveal an agent's architectural and reasoning flaws.
    \item Our comprehensive analysis reveals a critical finding: when fusion mechanisms are immature, additional modalities often hurt more than help. We demonstrate systemic weaknesses not exposed by prior benchmarks.
    \item We will open-source the entire OmniPlay platform, including all environments, baseline agents, and evaluation protocols, to facilitate relevant research.
\end{enumerate}

\section{Related Work}
\label{sec:related_work}

Our research is positioned at the intersection of multimodal evaluation and interactive agent learning. We structure our review by first discussing static benchmarks to highlight the need for agency, then examining interactive benchmarks to reveal their modal bottleneck, and finally arguing that the rise of omni-modal models has turned this bottleneck into a critical evaluation chasm that \textsc{OmniPlay} aims to bridge.

\subsection{Static Multimodal Benchmarks: Perception without Agency}

Early multimodal evaluation centered on static perception tasks. Seminal works like Visual Question Answering (VQA)~\citep{antol2015vqa, hudson2019gqa} and image captioning on datasets like COCO~\citep{chen2015microsoft} were foundational for representation learning. More recent comprehensive platforms, such as MMBench~\citep{liu2024mmbench} and SEED-Bench~\citep{li2023seedbench}, aggregated numerous tasks, yet they all share a unifying limitation: their static and non-interactive nature. Models perform single-turn perception on fixed inputs, which fails to evaluate crucial agentic capabilities like sequential decision-making or long-term planning.

\subsection{Interactive Agent Benchmarks: Agency with a Modal Bottleneck}

To address the lack of agency, a second wave of benchmarks introduced interactive environments. This evolution began in text-based worlds like Jericho~\citep{hausknecht2020interactivefictiongamescolossal}, expanded to embodied AI in 3D simulators such as AI2-THOR~\citep{kolve2017ai2} and Habitat~\citep{savva2019habitat}, and extended to grounded language in ALFWorld~\citep{ALFWorld20} and complex digital tasks in WebArena~\citep{zhou2023webarena} and Mind2Web~\citep{deng2023mind2web}. Despite this significant leap towards agency, a prevalent modal bottleneck constrains the majority of these benchmarks, as perception is typically limited to vision and text. Recent game-based works like BALROG~\citep{paglieri2025balrogbenchmarkingagenticllm} further highlight deep reasoning deficiencies even within these limited modalities. While pioneering platforms like SoundSpaces 2.0~\citep{chen2020soundspaces} incorporated audio for navigation, a comprehensive, diagnostic approach to omni-modality has been missing.

\subsection{Omni-Modal Models and The Evaluation Chasm}

This long-standing modal bottleneck has recently escalated into a critical evaluation chasm with the arrival of true omni-modal foundation models like Google's Gemini and OpenAI's GPT-4o. These models are natively designed to process a fluid combination of text, image, audio, and video, yet our primary tools for evaluating agency lack the sensory richness to test these new faculties. Current evaluations fail to assess how these powerful models perform in dynamic, multi-sensory scenarios where they must make choices.

\textsc{OmniPlay} is designed to bridge this chasm. It is the first interactive benchmark built on a core philosophy of modality interplay. By systematically creating tasks requiring synergistic fusion and stress-testing agents with controlled sensory conflicts, \textsc{OmniPlay} provides a dedicated diagnostic platform to rigorously evaluate the true interactive and reasoning capabilities of modern omni-modal agents.

% ===============================================================================
% SECTION: THE OMNIPLAY BENCHMARK
% ===============================================================================
\section{The OmniPlay Benchmark}

\begin{figure*}[t!] % Use [t!] to suggest placing it at the top of the page
    \centering
    \includegraphics[width=\textwidth]{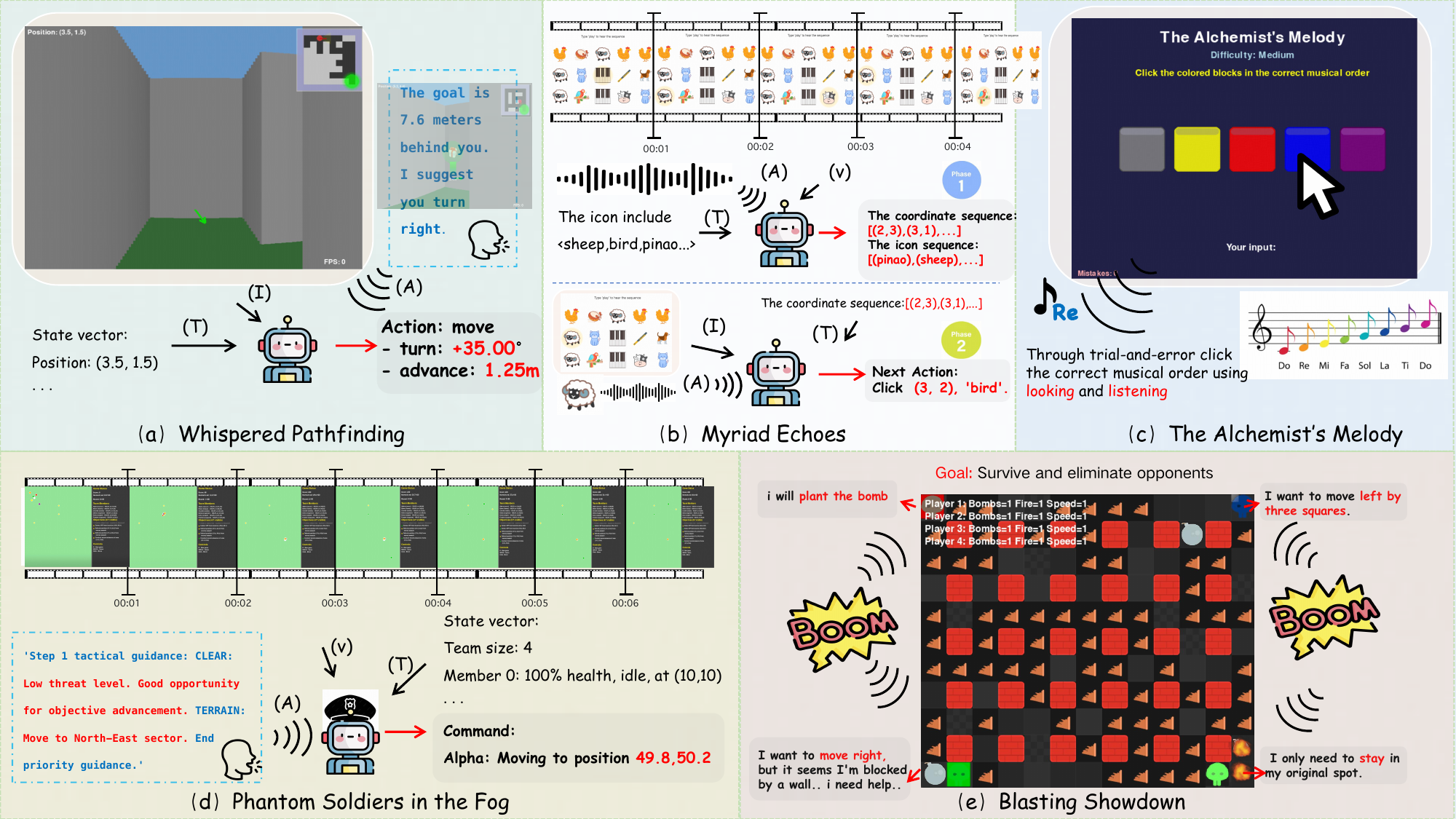}
    \caption{An overview of the five distinct game environments in the OmniPlay suite. Each sub-figure illustrates a game's user interface and the primary modalities involved: (I)mage, (V)ideo, (A)udio, and (T)ext. The suite is designed to test a diverse range of capabilities, from (a) visuo-auditory navigation and (b) sequence replication, to (c) abstract reasoning, (d) real-time strategy, and (e) multi-agent combat.}
    \label{fig:game_suite}
\end{figure*}

This section details the architectural design and core components of the OmniPlay benchmark. We begin by establishing a rigorous theoretical foundation by formalizing the agent-environment interaction within a generalized Markov Decision Process (MDP) framework. Following this, we articulate the core design principles that guided the benchmark's construction. We then introduce the five distinct game environments that constitute our suite and conclude by outlining our comprehensive evaluation protocol.

\subsection{Formalism: A Generalized Interaction Framework}

To provide a unified and rigorous description of agent-environment interaction across our diverse suite of games, we model each task within a generalized Markov Decision Process (MDP) framework. This formalism captures the sequential, turn-based nature of our benchmark. The interaction is defined by the primary components $(S, A, T, G, \Omega, O)$:

\begin{itemize}
    \item $S$: The set of true, underlying world states, which may be fully or partially observable.
    \item $A$: The agent's action space, which can be discrete, continuous, or hybrid.
    \item $T$: The state transition function, $T(s'|s,a)$.
    \item $G$: A set of goal states, $G \subseteq S$.
    \item $\Omega$: The multi-modal observation space. At each timestep $t$, the agent receives an observation $o_t \in \Omega$ composed of a tuple of available sensory inputs: $o_t = (\mathcal{I}_t, \mathcal{V}_t, \mathcal{A}_t, \mathcal{T}_t)$.
    \item $O$: The observation function, $O(o_t | s_t)$.
\end{itemize}

To process the omni-modal observation $o_t$, the agent first employs a set of modality-specific encoders ($E_\mathcal{I}, E_\mathcal{V}, E_\mathcal{A}, E_\mathcal{T}$) to obtain unimodal representations. These representations are then integrated by a fusion module, $\mathcal{F}$, to produce a unified context vector, $c_t$:
\begin{equation}
    c_t = \mathcal{F}(E_\mathcal{I}(\mathcal{I}_t), E_\mathcal{V}(\mathcal{V}_t), E_\mathcal{A}(\mathcal{A}_t), E_\mathcal{T}(\mathcal{T}_t))
\end{equation}

This context vector $c_t$ is then used to update the agent's internal state or history representation, $h_t$. At each timestep, the agent's policy $\pi(a_t | h_t)$ selects an action $a_t$. The agent's objective is to learn a policy that maximizes the probability of generating a successful trajectory terminating in a goal state. The full probabilistic objective function is detailed in Appendix~\ref{app:formalism_design_principles}.

\subsection{Core Design Principles}
The design of OmniPlay is engineered around three core principles intended to create a rigorous diagnostic platform. These principles directly address the shortcomings of prior benchmarks and are crafted to probe the foundational capabilities of omni-modal agents.

\begin{enumerate}
    \item \textbf{Modality Complementarity.} A primary flaw in many multimodal models is the tendency to rely on a ``dominant'' modality (e.g. vision) while only superficially processing others. To counter this, our first principle dictates that tasks must be \textit{unsolvable} without the complementary fusion of information from disparate channels. This forces the agent beyond simple multimodal co-occurrence and compels genuine cross-modal reasoning.

    \item \textbf{Controlled Modality Conflict.} The introduction of more sensory inputs can paradoxically degrade performance in agents with immature fusion mechanisms. Our second principle is to systematically introduce scenarios with \textit{conflicting information} to directly diagnose the robustness of an agent's fusion architecture. This principle turns OmniPlay into a tool for stress-testing an agent's decision-making process under ambiguity.

    \item \textbf{Various Modality Complexity.} A truly omni-modal agent should not be a specialist fine-tuned for a fixed set of inputs. Therefore, our third principle ensures that the benchmark suite as a whole presents \textit{varying combinations and complexities of modalities}. By offering a diverse suite of five games, we can systematically probe an agent's architecture for biases and assess its adaptability.
\end{enumerate}

\subsection{The OmniPlay Game Suite}
We introduce a suite of five distinct game environments, which we developed from the ground up to instantiate the design principles outlined above, with \textit{Blasting Showdown} drawing inspiration from classic game mechanics. An overview of these games is presented in Figure~\ref{fig:game_suite} and Table~\ref{tab:games}. The modalities are denoted as I (Image), A (Audio), T (Text), and V (Video). Detailed descriptions for each game, outlining their specific rules and challenges, are provided in the Appendix~\ref{app:game_details}.

\begin{table*}[t!]
\centering
\caption{High-Level Overview of the OmniPlay Game Suite. 
    Each game is designed with a unique combination of core objectives, sensory modalities, and diagnostic challenges to probe different facets of omni-modal agency. 
    Modalities are denoted as I (Image), A (Audio), T (Text), and V (Video).}
\label{tab:games}
% 使用 resizebox 将表格宽度严格限制在文本宽度内
\resizebox{\textwidth}{!}{%
% 使用 \setlength 稍微减小列间距，为内容腾出空间
\setlength{\tabcolsep}{5pt} % 默认是 6pt，这里稍微减小
\begin{tabular}{>{\bfseries}l p{4cm} c p{5.5cm}} 
\toprule
Game Name & Core Objective & Modalities & Core Challenge Diagnosed \\
\midrule
Whispered Pathfinding   & 3D Maze Navigation      & I, A, T    & Visuo-Auditory Integration \\
\addlinespace
Myriad Echoes           & Sequence Replication    & \begin{tabular}[c]{@{}l@{}}P1: V, A, T\\ P2: I, A, T\end{tabular} & Perception-Symbol-Action Grounding \\
\addlinespace
The Alchemist's Melody  & Abstract Rule Discovery & I, A, T    & Abstract Reasoning \\
\addlinespace
Phantom Soldiers in the Fog & Squad-based Strategy  & V, A, T    & Planning under Uncertainty \\
\addlinespace
Blasting Showdown       & Multi-agent Arena Combat& I, A, T    & Reactive Multi-agent Strategy \\
\bottomrule
\end{tabular}%
}
\end{table*}

\subsection{Evaluation Protocol and Metrics}
To provide a multi-faceted view of agent capabilities, our evaluation protocol comprises two categories of metrics. We benchmark against both a random agent baseline and a human expert baseline.

\textbf{Primary Metrics.} We gauge high-level task success using primary metrics. Key examples include \textbf{Success Rate (SR)}, the percentage of successful episodes, and \textbf{Efficiency Score (e.g., SPL)}, which rewards success while penalizing inefficient actions.

\textbf{Task-Specific Diagnostic Metrics.} To delve deeper into agent behavior, we also designed a comprehensive set of diagnostic metrics tailored to each game's core challenges.

\textbf{Note on Blasting Showdown.} As a competitive multi-agent environment, \textit{Blasting Showdown} follows a distinct AI-vs-AI tournament protocol. Its results are reported using task-specific metrics (e.g., win rate) and are excluded from comparisons based on human/random baselines, such as the Normalized Performance Score (NPS) used in later sections.

A complete list of all metrics with their detailed definitions, along with full experimental parameters, can be found in Appendix~\ref{app:exp_params_metrics}.

%%%%%%%%%%%%%%%%%%%%%%%%%%%%%%%%%%%%%%%%%%%%%%%%%%%%%%%%%
% SECTION 4: Experimental Setup
%%%%%%%%%%%%%%%%%%%%%%%%%%%%%%%%%%%%%%%%%%%%%%%%%%%%%%%%%
\section{Experimental Setup}
\label{sec:setup}

This section details the experimental methodology used to evaluate state-of-the-art omni-modal models on the OmniPlay benchmark. We first introduce the models and baselines under evaluation, followed by a description of our comprehensive evaluation protocol.

\subsection{Models and Baselines}
\label{sec:models_baselines}

Our evaluation suite comprises a diverse set of six representative omni-modal models, selected to cover both proprietary and open-source ecosystems:\footnote{GPT-4o's API lacks simultaneous support and requires strict approval for its audio-preview version.} 

\begin{itemize}
    \item \textbf{Proprietary Models:} Google's \textbf{Gemini 2.5 Pro} and \textbf{Gemini 2.5 Flash} \citet{comanici2025gemini}, accessed via their official APIs.
    \item \textbf{Open-Source Models:} \textbf{Qwen-2.5-Omni (7B)} \citet{Qwen2.5-Omni}, \textbf{MiniCPM-o-2.6 (8B)} \citet{yao2024minicpm}, \textbf{Baichuan-Omni-1.5 (7B)} \citet{li2025baichuan}, and \textbf{VITA-1.5 (7B)} \citet{fu2025vita}.
\end{itemize}

To contextualize the performance of these AI agents, we established two critical baselines. 
A \textbf{Random Agent} serves as a performance floor by sampling an action uniformly from the set of all available actions. 
More importantly, we established a \textbf{Human Expert} baseline by recruiting a diverse group of 12 experienced human players (all with over 500 hours of gaming experience). To enhance the robustness of our baseline, the cohort was gender-balanced and stratified into two age groups: eight young adults (20-35 years, 4 male, 4 female) and four middle-aged adults (35-50 years, 2 male, 2 female). All participants completed a warm-up period for each game to reach a stable skill plateau. A detailed breakdown of the recruitment protocol and inter-player agreement analysis is provided in Appendix~\ref{sec:appendix_human_baseline}.

\subsection{Evaluation Protocol and Metrics}
\label{sec:protocol}

To ensure a fair and reproducible comparison, our protocol is built upon a fixed set of evaluation seeds for each task. This means every agent (AI, random, or human) is evaluated on the exact same pre-defined sequence of game scenarios, eliminating procedural randomness as a confounding factor. The number of seeds varies per task (e.g., 50 for \textit{Whispered Pathfinding}, 30 for \textit{Phantom Soldiers}). A comprehensive breakdown of evaluation episodes and model parameters is available in Appendix~\ref{app:exp_params_metrics}.

To enable comparison across diverse tasks, we introduce a \textbf{Normalized Performance Score (NPS)}, calculated by scaling a model's raw score relative to the human and random baselines:
\begin{equation}
\label{eq:nps}
\text{NPS} = 100 \times \frac{\text{Score}_{\text{model}} - \text{Score}_{\text{random}}}{\text{Score}_{\text{human}} - \text{Score}_{\text{random}}}
\end{equation}
An NPS of 0 represents random performance, 100 matches human experts, and scores over 100 signify superhuman ability. The \textit{Blasting Showdown} task follows a distinct AI-vs-AI protocol and is excluded from NPS calculations, as noted in a footnote on its first mention. We also designed task-specific diagnostic metrics, detailed in Appendix~\ref{app:exp_params_metrics}.

%%%%%%%%%%%%%%%%%%%%%%%%%%%%%%%%%%%%%%%%%%%%%%%%%%%%%%%%%
% SECTION 5: Results and Analysis
%%%%%%%%%%%%%%%%%%%%%%%%%%%%%%%%%%%%%%%%%%%%%%%%%%%%%%%%%

\section{Results and Analysis}
\label{sec:results}

Our experiments reveal critical insights into the capabilities and limitations of current omni-modal models. We first present the overall performance, which shows a stark dichotomy between memory and reasoning, and then delve into diagnostic experiments to uncover the root causes.

\subsection{Overall Performance: A Dichotomy of Super-human Memory and Sub-par Reasoning}
\label{sec:overall_performance}

% --- 修改开始: 将两张图并列放置在一个 figure 环境中 ---
\begin{figure}[t!]
    \centering
    % (a) 子图: 雷达图
    \begin{subfigure}[b]{0.48\columnwidth}
        \centering
        \includegraphics[width=\linewidth]{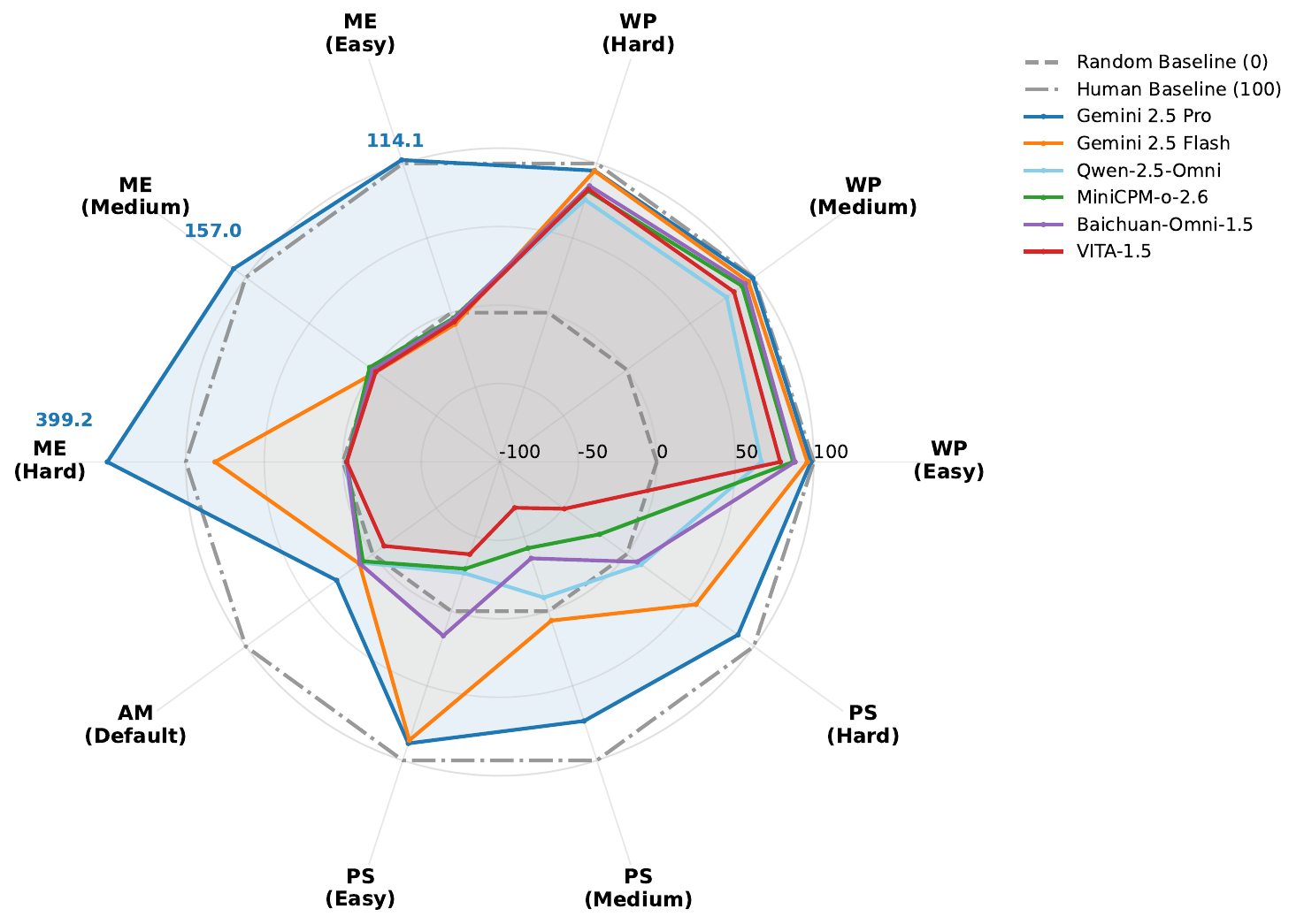}
        \caption{Performance on 10 NPS-benchmarked tasks, revealing a dichotomy between memory and reasoning.}
        \label{fig:radar_chart}
    \end{subfigure}
    \hfill % 在两个子图之间添加弹性空间
    % (b) 子图: 柱状图
    \begin{subfigure}[b]{0.48\columnwidth}
        \centering
        \includegraphics[width=\linewidth]{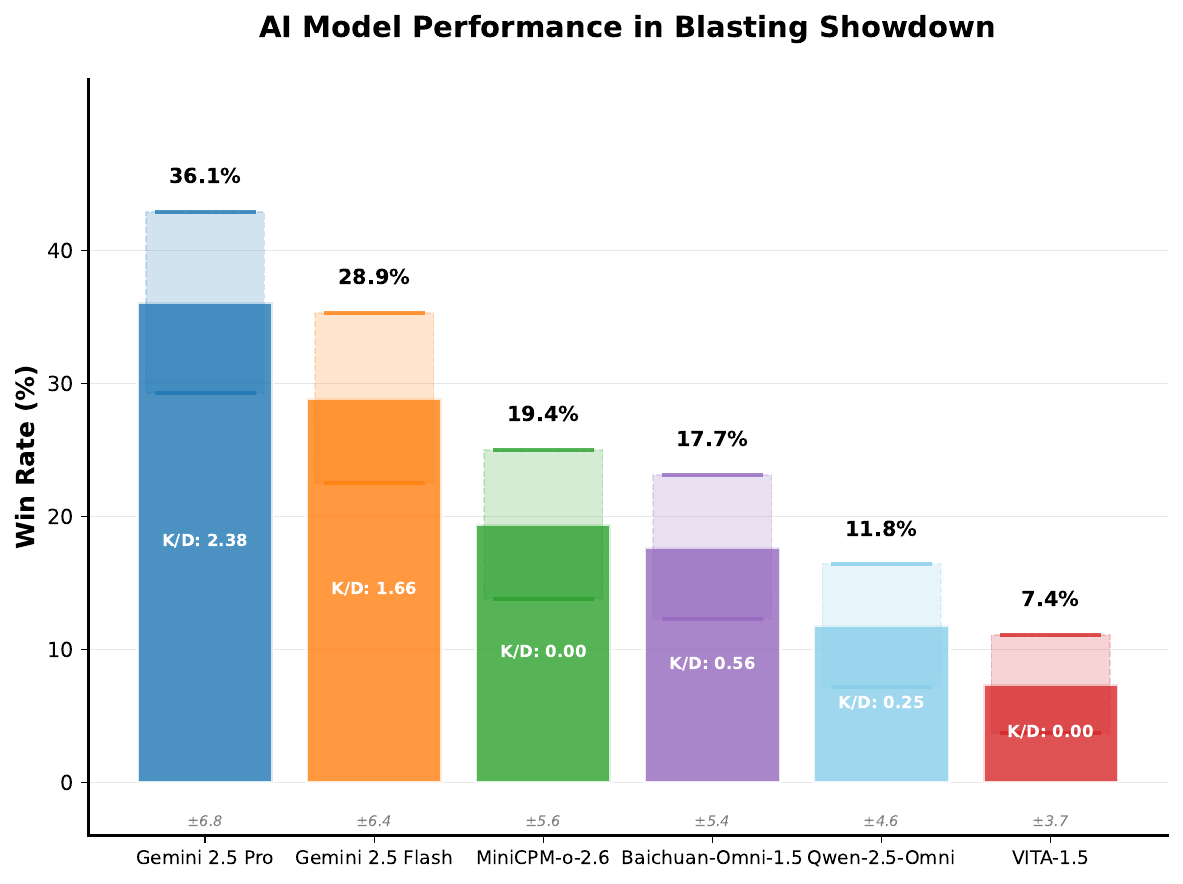}
        \caption{Performance in the \textit{Blasting Showdown} AI-vs-AI tournament, showing no dominant strategy.}
        \label{fig:bs_chart}
    \end{subfigure}
    \caption{Overall performance evaluation. (a) The radar chart highlights super-human memory but sub-par reasoning across tasks. (b) The tournament results underscore a shared deficit in strategic planning. Error margins represent SEM over 50 runs.}
    \label{fig:combined_overall_perf}
\end{figure}
% --- 修改结束 ---

We first evaluated all six AI models against the human and random baselines across 10 NPS-benchmarked tasks. The results, summarized visually in Figure~\ref{fig:radar_chart}, reveal a fascinating dichotomy: a clear exhibition of super-human prowess in specific domains contrasted with a systemic weakness in general reasoning and strategic planning. This systemic weakness is further corroborated by the AI-vs-AI tournament in \textit{Blasting Showdown} (Figure~\ref{fig:bs_chart})\footnote{As a competitive multi-agent environment, \textit{Blasting Showdown}'s results are reported using task-specific metrics (e.g., win rate) and are excluded from the NPS calculation.}. The results show that even the top-performing model, \textbf{Gemini 2.5 Pro}, achieved only a 36.1\% win rate, and its lead is not statistically decisive, reinforcing the conclusion that no single dominant strategy has yet emerged. Full statistical results for all benchmarked tasks are available in Appendix~\ref{sec:appendix_full_results}.

To precisely quantify this performance gap, Table~\ref{tab:dichotomy} presents the results on the most challenging memory and strategic reasoning tasks. On one hand, models exhibit super-human memory, with \textbf{Gemini 2.5 Pro} achieving an astounding NPS of 399.2 $\pm$ 3.6 in \textit{Myriad Echoes (Hard)}, a task where human cognitive limits form a bottleneck precisely because it relies heavily on short-term memory and precise sequence replication. On the other hand, this strength starkly contrasts with their brittle reasoning, as several models show negative NPS scores in strategic challenges like \textit{Phantom Soldiers}.

\begin{table}[h!]
\small 
\centering
\sisetup{
    separate-uncertainty = true,
    table-align-uncertainty = true,
    table-sign-mantissa = true
}
\caption{A Dichotomy in Performance: Superhuman Memory vs. Brittle Reasoning. Performance on the most challenging memory task (\textit{Myriad Echoes, Hard}) and strategic reasoning task (\textit{Phantom Soldiers, Hard}). All metrics are reported as Mean NPS $\pm$ SEM over 50 runs.}
\label{tab:dichotomy}
% Use tabular* to create a table with the full column width
\begin{tabular*}{\columnwidth}{@{\extracolsep{\fill}} l S[table-format=-3.1(1)] S[table-format=-2.1(1)]}
\toprule
\textbf{Model} & {\textbf{ME (Hard) NPS}} & {\textbf{PS (Hard) NPS}} \\
\midrule
Gemini 2.5 Pro      & \bfseries 399.2 \pm 3.6 & \bfseries 87.5 \pm 3.5 \\
Gemini 2.5 Flash    & 81.4 \pm 4.2            & 54.5 \pm 4.7           \\
Qwen-2.5-Omni       & -1.7 \pm 4.9            & 11.2 \pm 5.8           \\
MiniCPM-o-2.6       & -2.5 \pm 5.5            & -21.5 \pm 7.0          \\
Baichuan-Omni-1.5   & -2.5 \pm 5.3            & 8.3 \pm 6.2            \\
VITA-1.5            & -2.5 \pm 5.9            & -49.2 \pm 6.0          \\
\bottomrule
\end{tabular*}
\end{table}

% ===============================================================================
%          SECTION 5.2 - Core Finding (Revised with controlled wrapfigure start)
% ===============================================================================

\subsection{Core Finding: Brittle Fusion and The ``less is more'' Paradox}
\label{sec:core_finding}

To understand the root causes of these reasoning deficits, we conducted a series of diagnostic experiments.

\textbf{Modality Conflict.} 
We first stress-tested the models' fusion mechanisms by injecting controlled modality conflicts. We selected \textit{Whispered Pathfinding} for this analysis, as its core visuo-auditory navigation challenge makes it an ideal environment to create such conflicts. In the ``Audio Conflict'' 

% --- The wrapfigure is placed here, at the start of the new paragraph ---
\begin{wrapfigure}{r}{0.5\columnwidth}
    \vspace{-20pt} % Aggressive vertical adjustment
    \centering
    % (a) Subfigure: Conceptual Diagram
    \begin{subfigure}{\linewidth}
        \centering
        \includegraphics[width=\linewidth]{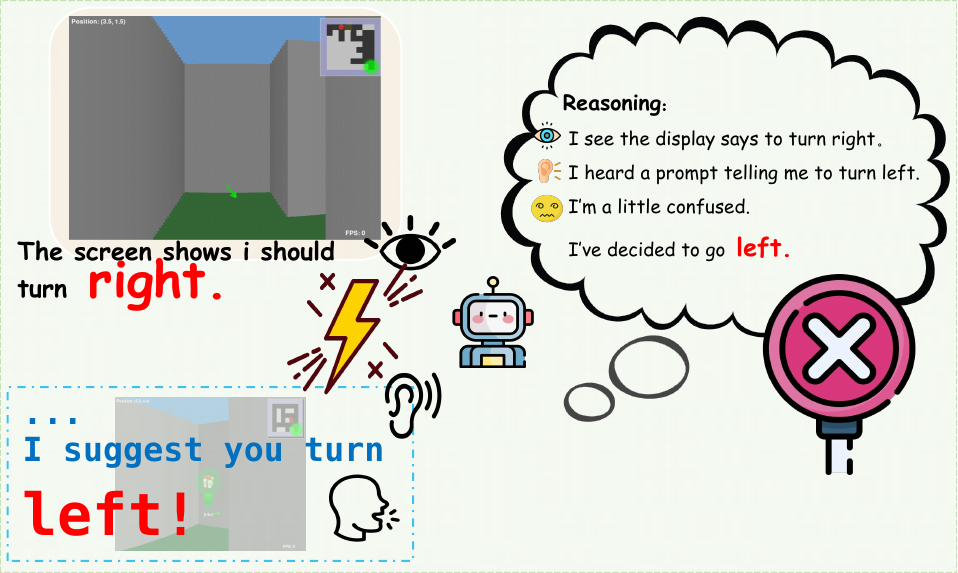}
        \subcaption{Conceptual diagram of a modality conflict scenario in \textit{Whispered Pathfinding}.}
        \label{fig:conflict_scenario}
    \end{subfigure}
    
    \vspace{1em} % Space between subfigures
    
    % (b) Subfigure: Results Chart
    \begin{subfigure}{\linewidth}
        \centering
        \includegraphics[width=\linewidth]{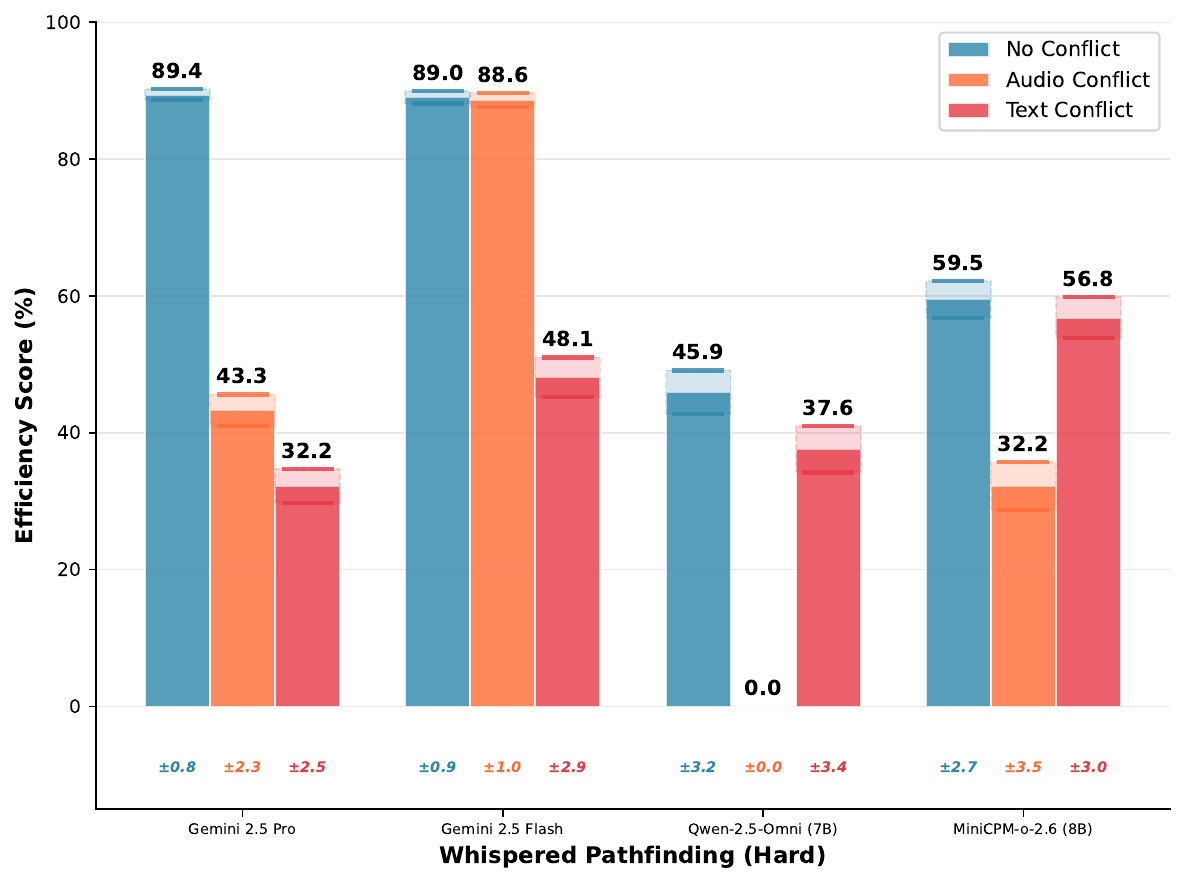}
        \subcaption{Resulting efficiency score (\%) under modality conflict, exposing fragile fusion mechanisms.}
        \label{fig:conflict_results}
    \end{subfigure}
    
    \caption{Modality conflict experiments. (a) illustrates the contradictory cues. (b) shows the resulting performance degradation.}
    \label{fig:combined_conflict_figures}
\end{wrapfigure}

condition, for instance, an on-screen visual cue was paired with a contradictory verbal command, forcing the agent to resolve ambiguity.

This setup is conceptually illustrated in Figure~\ref{fig:conflict_scenario}. The results of this stress test, presented in Figure~\ref{fig:conflict_results} (with full statistical data in Appendix~\ref{app:diagnostic_stats}), reveal that modality conflict induces a drastic and statistically significant degradation in performance, exposing the brittleness of their fusion mechanisms. For the top-performing \textbf{Gemini 2.5 Pro}, the introduction of either audio or text conflict causes a catastrophic drop in efficiency, from a stable baseline of 89.4\% (SEM=$\pm$0.8) to 43.3\% (SEM=$\pm$2.3) and 32.2\% (SEM=$\pm$2.5), respectively. The increased SEM values under conflict also indicate a significant rise in performance instability.

Furthermore, we observe a fascinating asymmetrical sensitivity to different conflict types. \textbf{Gemini 2.5 Flash}, for example, is remarkably resilient to auditory conflicts, maintaining its performance at 88.6\% (SEM=$\pm$1.0), nearly identical to its no-conflict baseline of 89.0\% (SEM=$\pm$0.9). However, it is highly vulnerable to textual conflicts, where its performance plummets to 48.1\% (SEM=$\pm$2.9). This stark difference strongly implies a hierarchical reliance on vision and text over audio in its decision-making process. The complete collapse of \textbf{Qwen-2.5-Omni} under audio conflict (0.0\%) further underscores the diverse and often fragile nature of current fusion architectures.

\textbf{Modality Ablation.} This brittleness, in turn, helps explain a counter-intuitive ``less is more'' paradox observed in our modality ablation experiments (Figure~\ref{fig:ablation_experiment}; see Appendix~\ref{app:diagnostic_stats} for detailed statistics). We selected \textit{Whispered Pathfinding} to represent synergistic navigation and \textit{Myriad Echoes} for complex sequence grounding.

The results reveal two key phenomena. First, for synergistic tasks requiring tight multi-modal fusion, top models like \textbf{Gemini 2.5 Pro} demonstrate a clear necessity for all modalities. As shown in Figure~\ref{fig:ablation_experiment} (right), its performance in \textit{Myriad Echoes} drops significantly from a peak score of 11.85 when any single modality is removed, confirming that its high performance is contingent on holistic sensory integration.

Second, and more strikingly, we found compelling evidence of the ``less is more'' paradox in models with weaker fusion mechanisms. For \textbf{MiniCPM-o-2.6} in \textit{Whispered Pathfinding} (Figure~\ref{fig:ablation_experiment}, left), the full-modality performance is a modest 48.8\% (SEM=$\pm$3.6). However, removing the visual modality causes its Efficiency Score to dramatically and significantly increase to 81.4\% (SEM=$\pm$1.6). The concurrent sharp decrease in SEM suggests the model not only performs better but also becomes substantially more stable. This strongly suggests that for a model with immature fusion capabilities, a conflicting or poorly processed sensory input can act as a liability rather than an asset.

\begin{figure*}[h]
    \centering
    % Ensure the image path is correct
    \includegraphics[width=\textwidth]{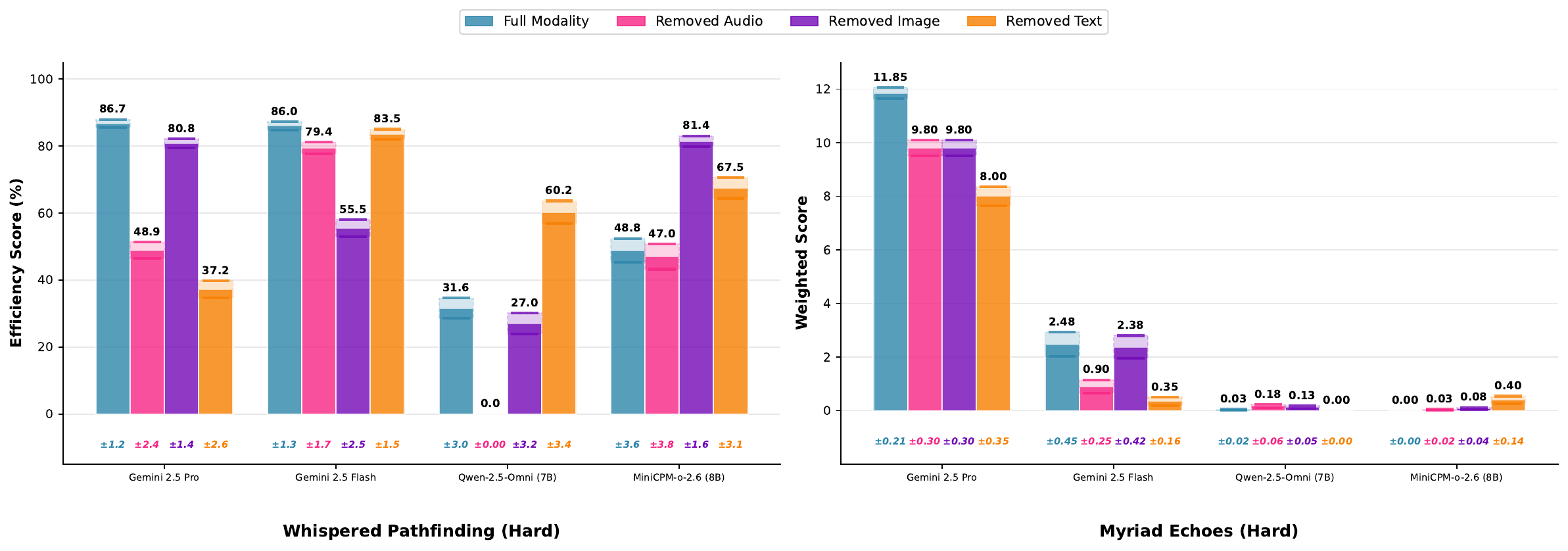}
    \caption{Modality ablation experiments on the 'Hard' difficulty for \textit{Whispered Pathfinding} (left) and \textit{Myriad Echoes} (right). Error margins, indicated by shaded regions and bottom values (SEM), reveal two key phenomena: (1) For top models, removing any modality hurts performance in a synergistic task (necessity). (2) For other models, removing a modality can paradoxically improve performance (``less is more'').}
    \label{fig:ablation_experiment}
\end{figure*}

\subsection{Breadth of Diagnostics: Other Key Findings}
\label{sec:breadth_diagnostics}
Beyond the core fusion deficiencies, our diagnostic suite probed several other critical dimensions. The methodologies for our diagnostic experiments are outlined in Appendix~\ref{sec:appendix_diagnostics}, with further qualitative case studies in Appendix~\ref{app:qualitative_cases} and full raw performance data in Appendix~\ref{sec:appendix_full_results}.

\begin{itemize}
\item \textbf{Robustness to Noise:} Models are highly sensitive to sensory noise. In \textit{Phantom Soldiers}, moderate visual noise caused Gemini 2.5 Pro's score to plummet by over 40\%, suggesting reliance on superficial correlations.
\item \textbf{Aided Reasoning via Prompting:} Proprietary models show a remarkable ability to leverage explicit hints. In \textit{The Alchemist's Melody}, providing a color-to-note mapping boosted Gemini models to 100\% completion, a gain not seen in open-source counterparts.
\item \textbf{Task Complexity Validation:} The benchmark's core difficulty was validated, as even a simplified version of \textit{Myriad Echoes} remained unsolvable for weaker models.
\item \textbf{Modality Substitution:} Models find text easier to process than other modalities. In \textit{Phantom Soldiers}, substituting auditory alerts with equivalent textual descriptions led to a \textit{consistent performance increase} across all models, reinforcing our finding about brittle fusion for non-textual information.

\end{itemize}

\section{Conclusion}
\label{sec:conclusion}

In this work, we addressed a critical evaluation chasm for omni-modal AI by introducing \textbf{OmniPlay}, the first interactive benchmark designed to diagnose, rather than merely measure, an agent's fusion and reasoning capabilities across the full sensory spectrum. Our comprehensive evaluation of six leading models revealed a stark dichotomy: while these agents exhibit superhuman performance on high-fidelity memory tasks, they suffer from systemic failures in challenges requiring robust reasoning and strategic planning. We demonstrated that the root cause of this fragility lies in their fusion mechanisms. These mechanisms lead to catastrophic performance degradation when faced with modality conflicts or sensory noise, and give rise to a counter-intuitive ``less is more'' paradox, where removing sensory information can paradoxically improve performance.

Our findings carry a significant implication for the pursuit of AGI: simply scaling models may not be sufficient to bridge the gap to robust, real-world intelligence. The path forward requires a research focus that extends beyond architectural depth to explicitly address the foundational challenges of synergistic fusion, conflict arbitration, and resilient reasoning. OmniPlay provides the community with a diagnostic toolkit to probe these fundamental weaknesses.

While OmniPlay represents a significant step forward, we acknowledge its limitations and see them as exciting avenues for future work. Our study focused on zero-shot performance, while also incorporating targeted probes of the models' in-context learning capabilities. Investigating how fine-tuning or new architectural innovations might remedy these fusion deficiencies is a critical next step. Furthermore, extending these diagnostic principles from simulated environments to real-world physical interaction remains a grand challenge for the field.

\bibliography{icrl3026_conference}
\bibliographystyle{icrl3026_conference}

% ===============================================================================
% APPENDIX SECTION
% ===============================================================================
\newpage
% The \appendix command signals the start of the appendix. 
% It will automatically handle the section numbering (A, B, C...).
\appendix

% The \section* command creates an unnumbered section title, which is
% a clean way to title the appendix section as a whole.
\section*{Appendix} 

% --- THIS IS THE SAFE AND RECOMMENDED "APPENDIX ROADMAP" ---

% --- Now, you can start the content of your actual appendix sections ---

\section{Use of Large Language Models in Manuscript Preparation}
\label{sec:appendix_llm_use}

We utilized LLMs as writing assistants during the preparation of this manuscript. Their role was strictly limited to improving grammar, refining phrasing, and enhancing the overall readability and clarity of the text. The conceptualization of the research, the design and execution of the experiments, and the analysis and interpretation of the results are entirely the original work of the authors.

%%%%%%%%%%%%%%%%%%%%%%%%%%%%%%%%%%%%%%%%%%%%%%%%%%%%%%%%%
% Appendix B: Formalism and Design Principles
%%%%%%%%%%%%%%%%%%%%%%%%%%%%%%%%%%%%%%%%%%%%%%%%%%%%%%%%%
\section{Formalism and Design Principles}
\label{app:formalism_design_principles}

This appendix provides a detailed description of the generalized Markov Decision Process (MDP) framework used in OmniPlay and illustrates how our core design principles are instantiated within this formalism.

\subsection{The Generalized Interaction Framework}
As introduced in Section 3.2, we model each task as an MDP defined by the tuple $(S, A, T, G, \Omega, O)$. The agent's core challenge is to process an omni-modal observation $o_t = (\mathcal{I}_t, \mathcal{V}_t, \mathcal{A}_t, \mathcal{T}_t) \in \Omega$ at each timestep, produce a unified context vector $c_t$ via a fusion module $\mathcal{F}$ (Equation 1), and select an optimal action $a_t$ based on its history $h_t$.

The agent's objective is to learn a policy $\pi$ that maximizes its likelihood of successfully completing a given task. Let $\tau = (s_0, a_0, s_1, a_1, \dots)$ denote a trajectory. The probability of observing $\tau$ given a policy $\pi$ is:
\begin{equation}
    P(\tau | \pi) = p(s_0) \prod_{t=0}^{|\tau|-1} \pi(a_t | h_t) T(s_{t+1} | s_t, a_t)
\end{equation}
Let $\mathcal{T}_G$ be the set of all trajectories that terminate in a goal state $s_g \in G$. The optimal policy $\pi^*$ is the one that solves:
\begin{equation}
    \pi^* = \arg\max_{\pi} \sum_{\tau \in \mathcal{T}_G} P(\tau | \pi)
\end{equation}

\subsection{Formalizing the Core Design Principles}
Our three core design principles—Modality Interdependence, Controlled Modality Conflict, and Variable Modality Complexity—are not merely abstract concepts but are formally embedded within the MDP structure.

\paragraph{Modality Interdependence.} This principle is primarily realized through the design of the state transition function $T(s'|s,a)$ and the goal states $G$. A task is interdependent if, for many states $s$, there is no single modality in the observation $o_t$ that provides sufficient information for the policy $\pi$ to choose an action $a_t$ that maintains a high probability of reaching $G$. Formally, let $\pi_m$ be a policy that only conditions on a single modality $m \in \{\mathcal{I}, \mathcal{V}, \mathcal{A}, \mathcal{T}\}$. An interdependent task ensures that:
\begin{equation}
    \max_{\pi} \sum_{\tau \in \mathcal{T}_G} P(\tau | \pi) \gg \max_{m} \left( \max_{\pi_m} \sum_{\tau \in \mathcal{T}_G} P(\tau | \pi_m) \right)
\end{equation}
This inequality formally states that the performance of a full omni-modal policy is significantly greater than the best possible uni-modal policy.

\paragraph{Controlled Modality Conflict.} We introduce conflict by manipulating the observation function $O(o_t|s_t)$. In a conflict scenario, the observation $o_t = (\dots, m_i, \dots, m_j, \dots)$ contains information from two or more modalities, $m_i$ and $m_j$, that suggest contradictory optimal actions. For instance, modality $m_i$ suggests an action $a_i$ that maximizes the value function $V^\pi(s)$, while modality $m_j$ suggests an action $a_j$ that leads to a much lower value. This forces the agent's fusion module $\mathcal{F}$ and policy $\pi$ to resolve the ambiguity.

\paragraph{Variable Modality Complexity.} This principle is reflected in the diversity of the observation spaces $\Omega$ and action spaces $A$ across our suite of five games. For example, the $\Omega$ for \textit{Whispered Pathfinding} contains continuous spatialized audio, while the $\Omega$ for \textit{The Alchemist's Melody} involves discrete auditory tones. Similarly, the action space $A$ ranges from continuous navigation controls to discrete clicking actions. This variation across the set of MDPs $\{ \text{MDP}_1, \dots, \text{MDP}_5 \}$ ensures that we are not testing a model's specialization to a single type of environment but its general omni-modal capability.

%%%%%%%%%%%%%%%%%%%%%%%%%%%%%%%%%%%%%%%%%%%%%%%%%%%%%%%%%
% Appendix C: Game Environment Details
% REQUIRES: \usepackage{float} in the preamble
%%%%%%%%%%%%%%%%%%%%%%%%%%%%%%%%%%%%%%%%%%%%%%%%%%%%%%%%%

\section{Game Environment Details}
\label{app:game_details}

This appendix provides detailed descriptions for each of the five game environments in the OmniPlay suite. For each game, we outline its core objective, the modalities and user interface (UI) presented to the agent, its core gameplay mechanics, and the prompting structure. Screenshots of each game's UI and prompts are included for visual reference.

%--------------------------------------------------------
\subsection{Whispered Pathfinding}
%--------------------------------------------------------

\paragraph{Core Objective.} The agent's goal is to navigate a procedurally generated 3D maze to find a hidden, stationary target location.

\begin{figure}[H] % Changed [h!] to [H] for fixed positioning
    \centering
    \includegraphics[width=\columnwidth]{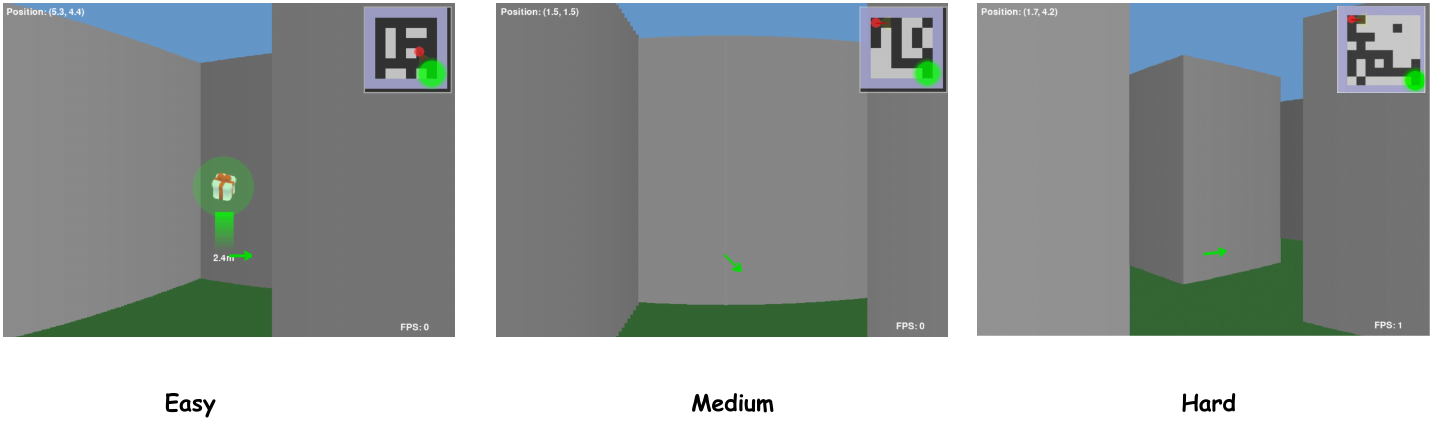}
    \caption{User interface for the \textit{Whispered Pathfinding} environment across difficulties.}
    \label{fig:wp_ui}
\end{figure}

\begin{figure}[H] % Changed [h!] to [H]
    \centering
    \includegraphics[width=0.8\columnwidth]{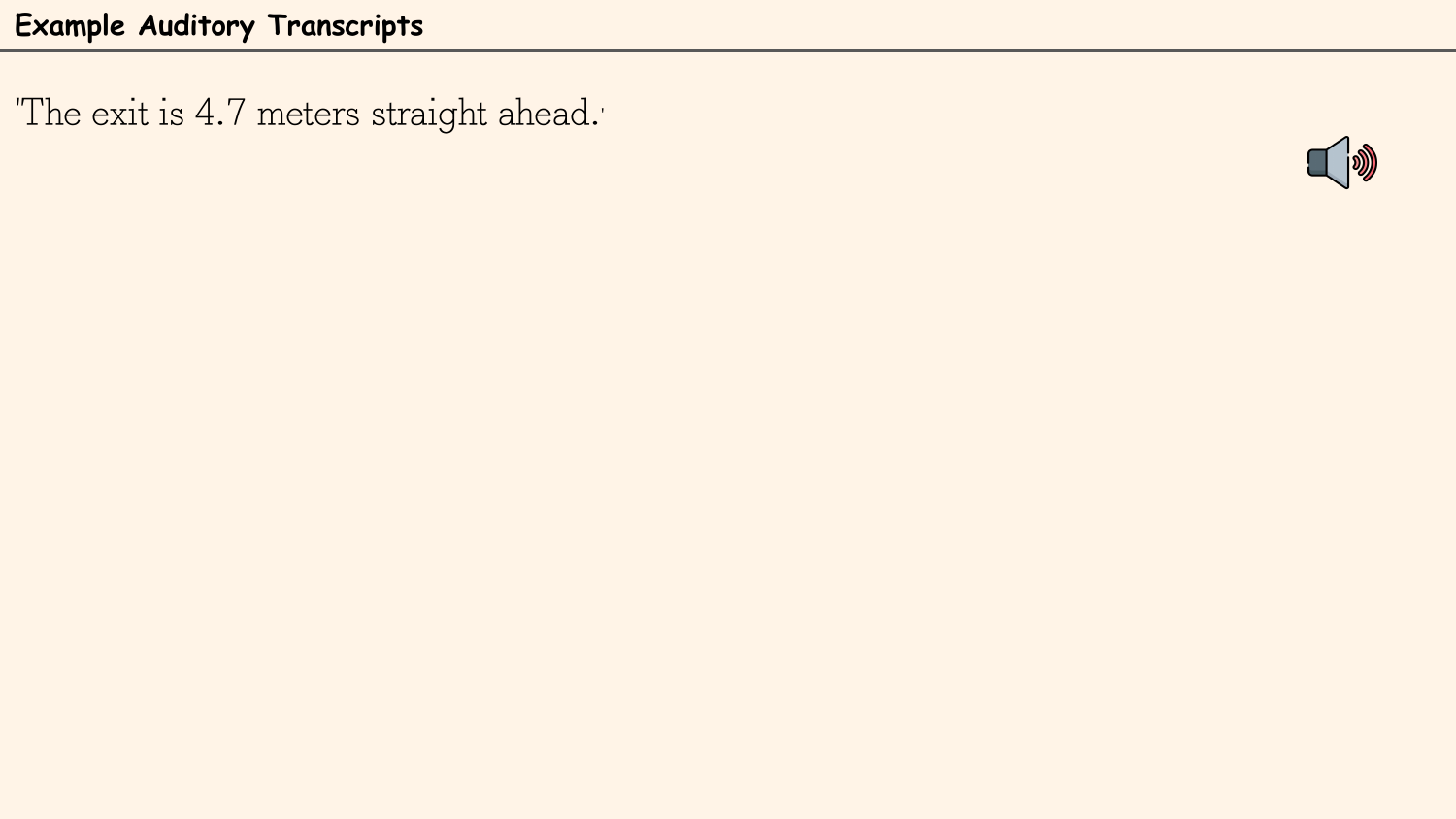} 
    \caption{Example Auditory Transcript from \textit{Whispered Pathfinding}. This text content is converted to speech for the agent.}
    \label{fig:wp_audio_transcript}
\end{figure}

\paragraph{Modalities and UI.} The agent perceives the environment through three primary modalities. An example of the UI is shown in Figure~\ref{fig:wp_ui}.
\begin{itemize}
    \item \textbf{Image (I):} A first-person visual feed showing the maze walls and corridors.
    \item \textbf{Audio (A):} Synthesized verbal guidance delivered as Text-to-Speech audio. An example of the transcribed audio content is shown in Figure~\ref{fig:wp_audio_transcript}.
    \item \textbf{Text (T):} The complete turn-based prompt, which provides a structured dump of the agent's current state and tasks the agent with generating the next action.
\end{itemize}

\paragraph{Gameplay Mechanics.} The agent's action space is continuous, consisting of rotation and forward movement. Success requires synergizing the visual information with the auditory guidance.

\subsubsection{Prompting Structure}
Interaction with the agent is structured via a system prompt that defines its role and a turn prompt that provides state information for each action.

\begin{figure}[H] % Changed [h!] to [H]
    \centering
    \includegraphics[width=\columnwidth]{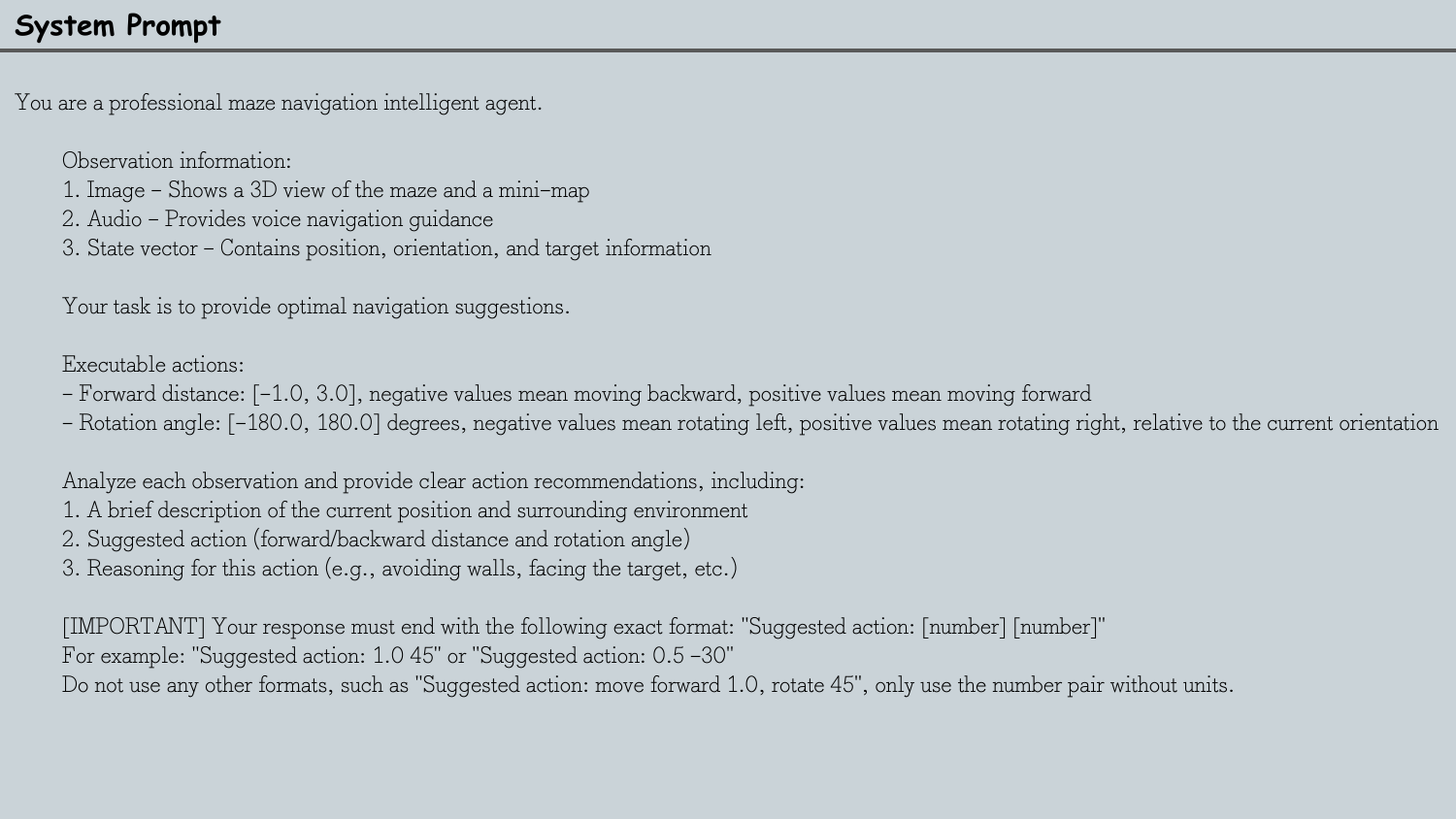}
    \caption{System Prompt for \textit{Whispered Pathfinding}.}
    \label{fig:wp_sp}
\end{figure}

\paragraph{System Prompt.} The system prompt, shown in Figure~\ref{fig:wp_sp}, is used to initialize the agent's behavior, defining its persona, capabilities, and required output format.

\begin{figure}[H] % Changed [h!] to [H]
    \centering
    \includegraphics[width=\columnwidth]{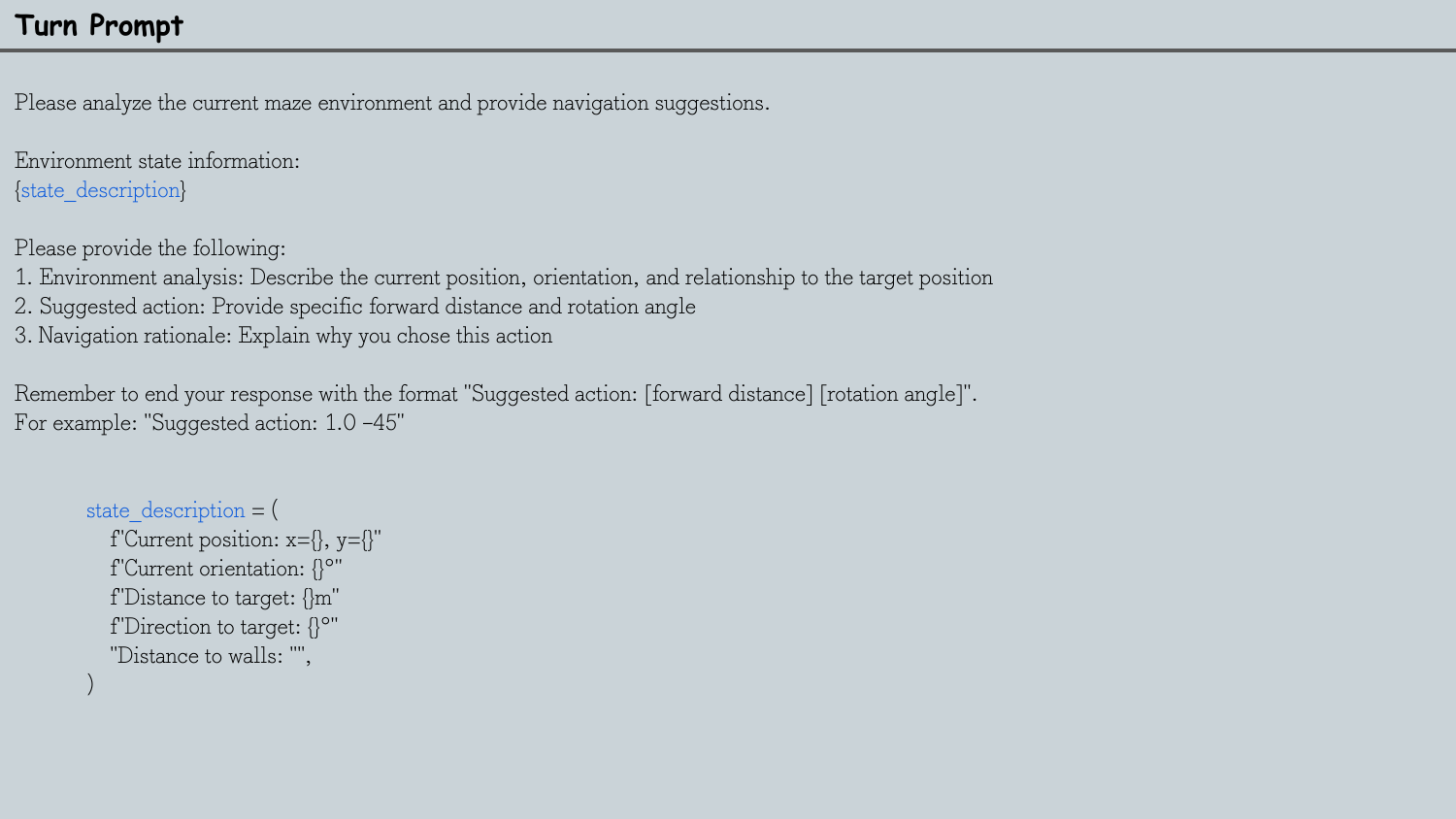}
    \caption{Turn Prompt for \textit{Whispered Pathfinding}.}
    \label{fig:wp_tp}
\end{figure}

\paragraph{Turn Prompt.} At each decision step, the text modality consists of the prompt shown in Figure~\ref{fig:wp_tp}, where \texttt{\{state\_description\}} is populated with real-time data.

%--------------------------------------------------------
\subsection{Myriad Echoes}
%--------------------------------------------------------
\paragraph{Core Objective.} This task diagnoses the full perception-to-symbol-to-action pipeline across two distinct phases.

\begin{figure}[H] % Changed [h!] to [H]
    \centering
    \includegraphics[width=\columnwidth]{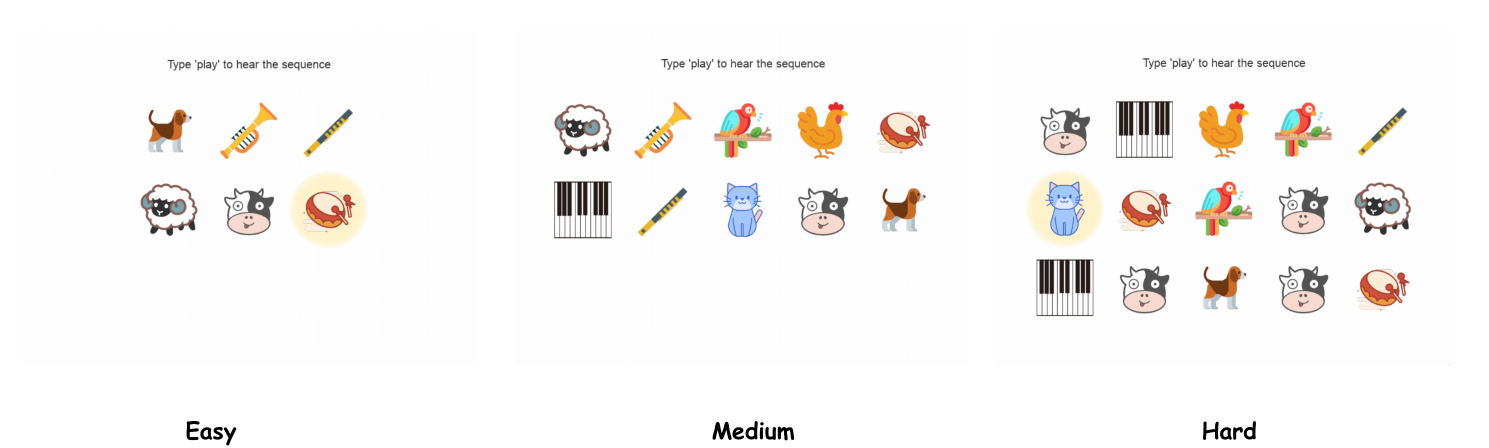}
    \caption{User interface for the \textit{Myriad Echoes} environment across difficulties.}
    \label{fig:me_ui}
\end{figure}

\paragraph{Modalities and UI.} The UI for both phases is shown in Figure~\ref{fig:me_ui}.
\begin{itemize}
    \item \textbf{Phase 1 (Transcription):} The agent is presented with a dynamic sequence of highlighted icons (\textbf{Video}) and corresponding unique sounds (\textbf{Audio}).
    \item \textbf{Phase 2 (Execution):} The agent is presented with a static grid of icons (\textbf{Image}) and receives auditory feedback (\textbf{Audio}) on clicks. The ground-truth sequence is provided via a textual prompt (\textbf{Text}).
\end{itemize}

\paragraph{Gameplay Mechanics.} In Phase 1, the agent must parse the multi-modal stream. In Phase 2, it must execute the parsed sequence by clicking the icons in the correct order.

\subsubsection{Prompting Structure}
\begin{figure}[H] % Changed [h!] to [H]
    \centering
    \includegraphics[width=\columnwidth]{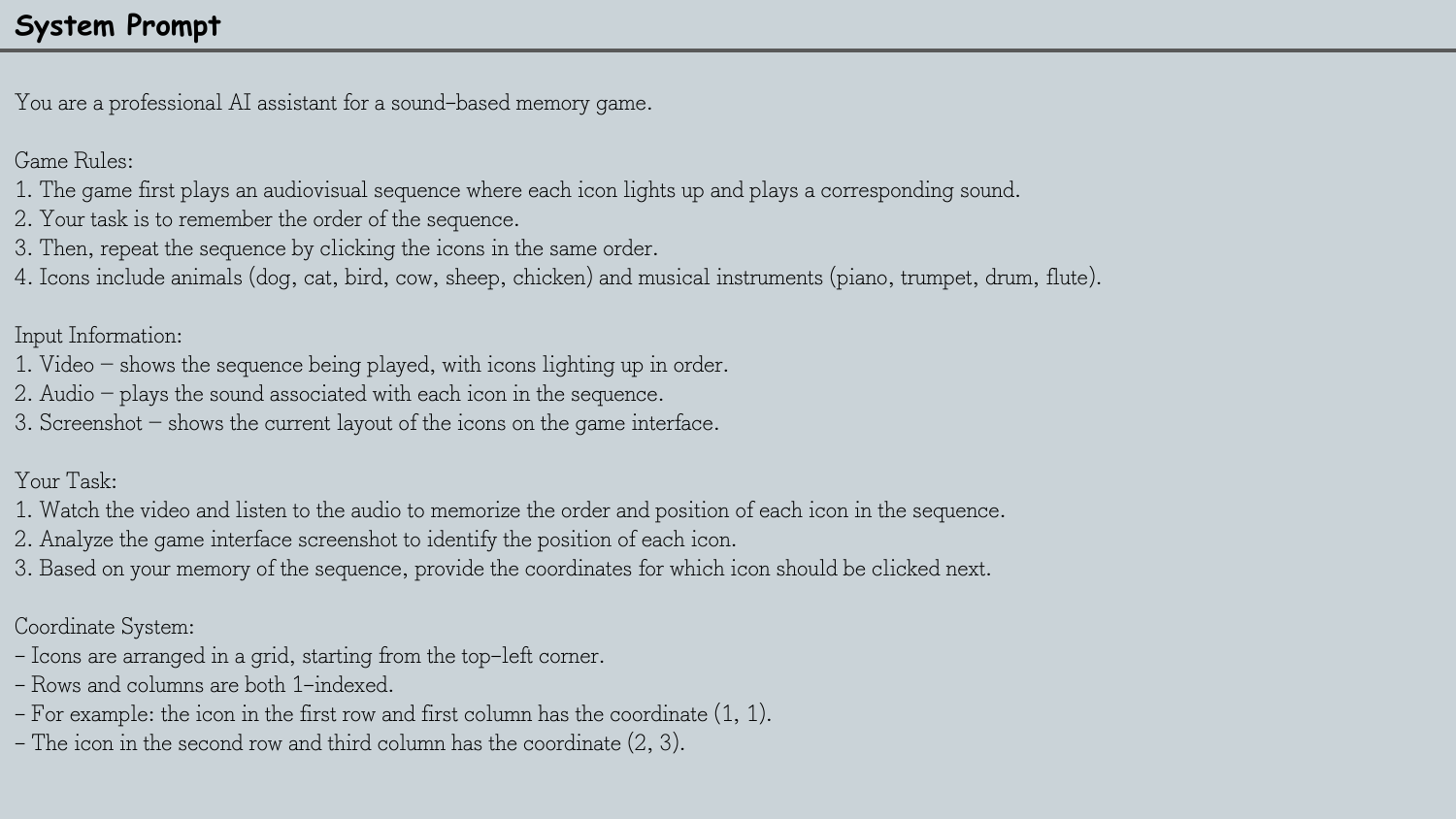}
    \caption{System Prompt for \textit{Myriad Echoes}.}
    \label{fig:me_sp}
\end{figure}

\paragraph{System Prompt.} The agent is initialized with the system prompt shown in Figure~\ref{fig:me_sp}.

\begin{figure}[H] % Changed [h!] to [H]
    \centering
    \includegraphics[width=0.49\columnwidth]{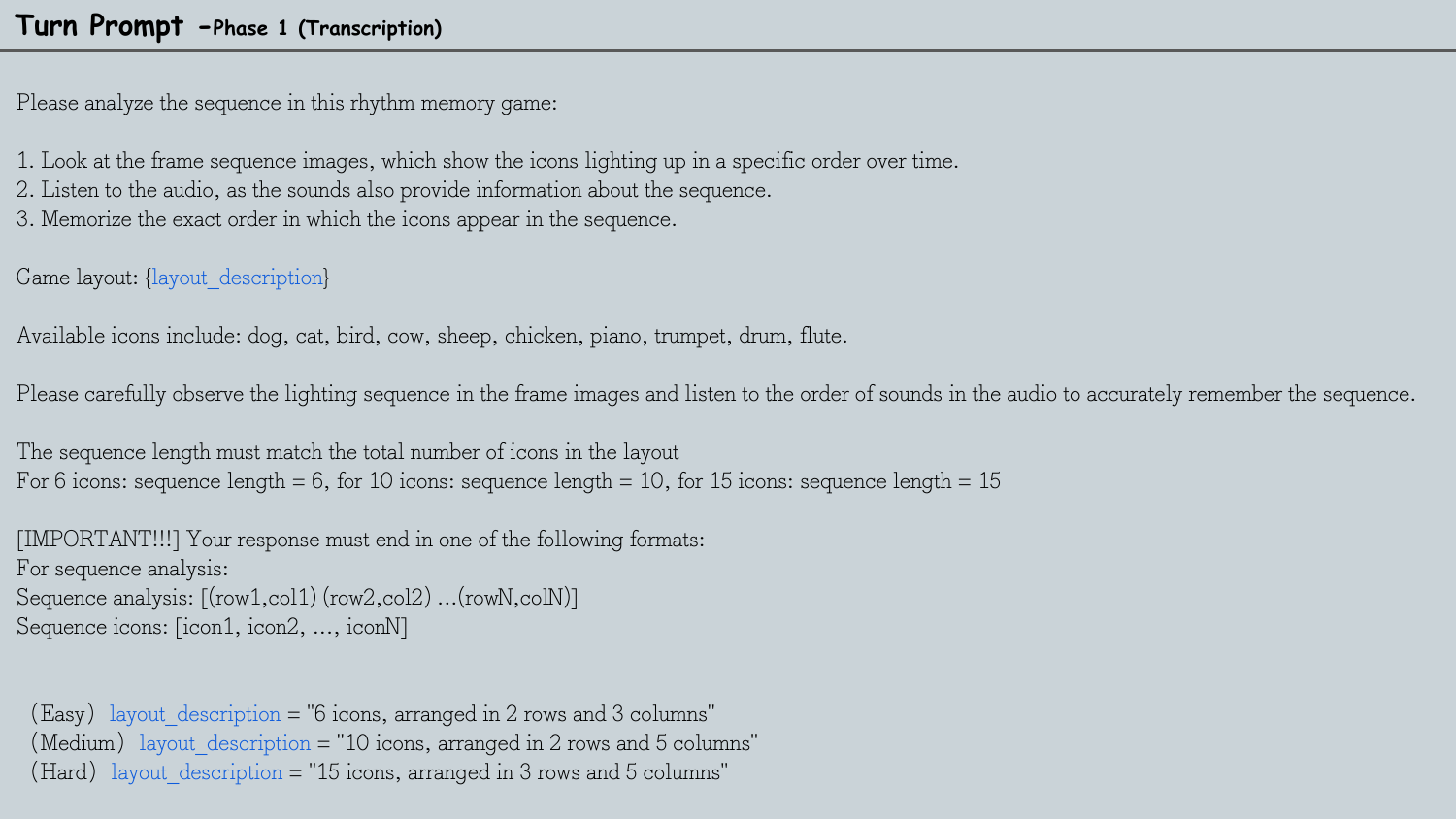}
    \includegraphics[width=0.49\columnwidth]{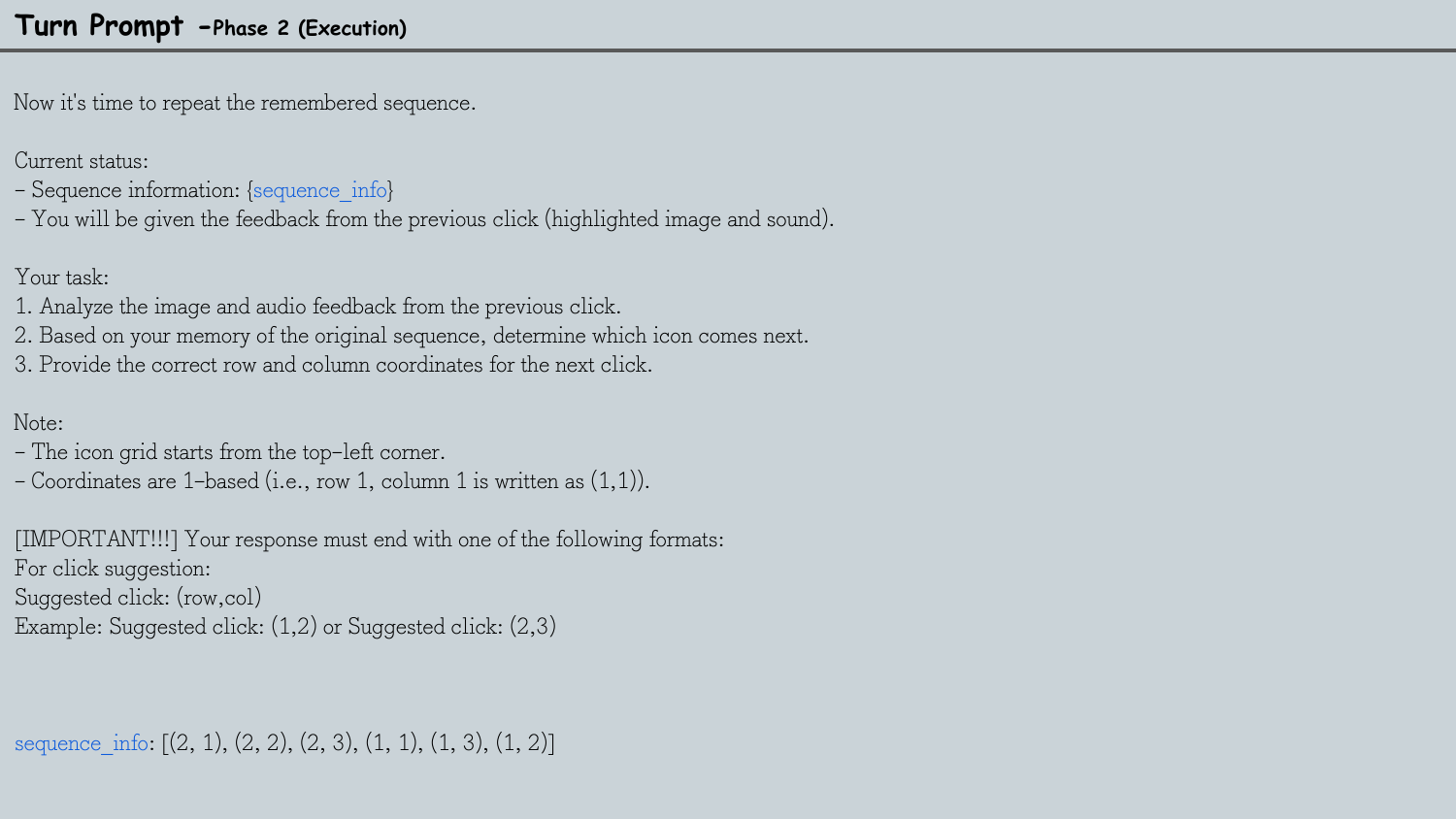}
    \caption{Turn Prompt and UI for Phase 1 (left) and Phase 2 (right) of \textit{Myriad Echoes}.}
    \label{fig:me_tp1}
    \label{fig:me_tp2}
\end{figure}

\paragraph{Turn Prompt.} The prompt for Phase 2 is the ground-truth sequence, visualized in Figure~\ref{fig:me_tp1} and~\ref{fig:me_tp2}.

%--------------------------------------------------------
\subsection{The Alchemist's Melody}
%--------------------------------------------------------
\paragraph{Core Objective.} The agent must discover a latent mapping between colors and musical notes to reproduce a specified musical scale.

\begin{figure}[H] % Changed [h!] to [H]
    \centering
    \includegraphics[width=0.7\columnwidth]{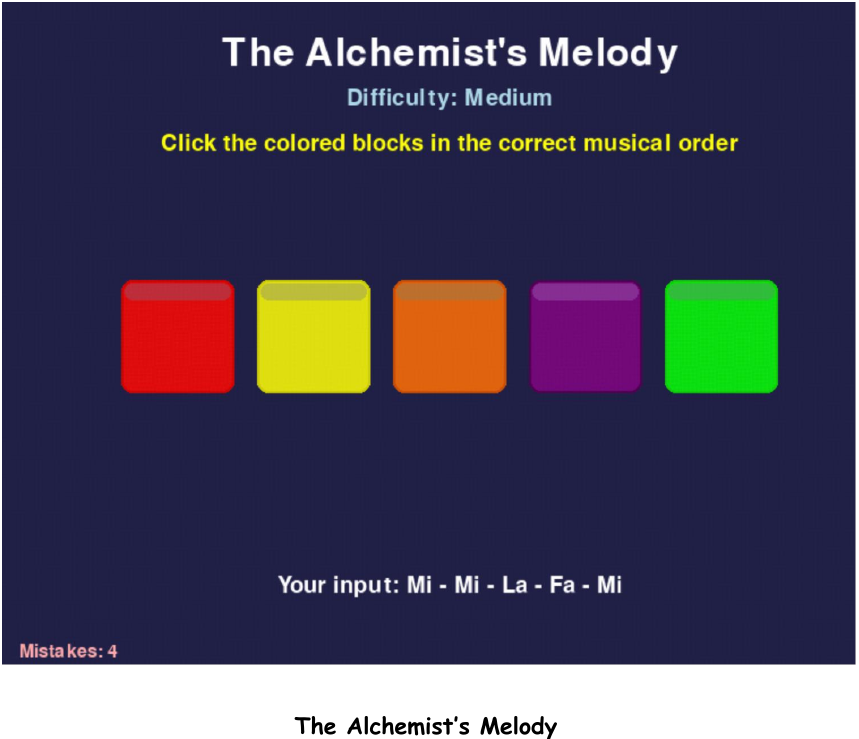}
    \caption{User interface for \textit{The Alchemist's Melody}.}
    \label{fig:am_ui}
\end{figure}

\paragraph{Modalities and UI.} The UI is shown in Figure~\ref{fig:am_ui}.
\begin{itemize}
    \item \textbf{Image (I):} A set of clickable colored blocks.
    \item \textbf{Audio (A):} Clicking a block plays a musical note.
    \item \textbf{Text (T):} A highly structured, real-time state dump containing feedback, sequence status, and strategic hints.
\end{itemize}

\paragraph{Gameplay Mechanics.} The color-note mapping is randomized per episode. The agent must deduce it via trial-and-error, guided by the rich textual feedback.

\subsubsection{Prompting Structure}
\begin{figure}[H] % Changed [h!] to [H]
    \centering
    \includegraphics[width=\columnwidth]{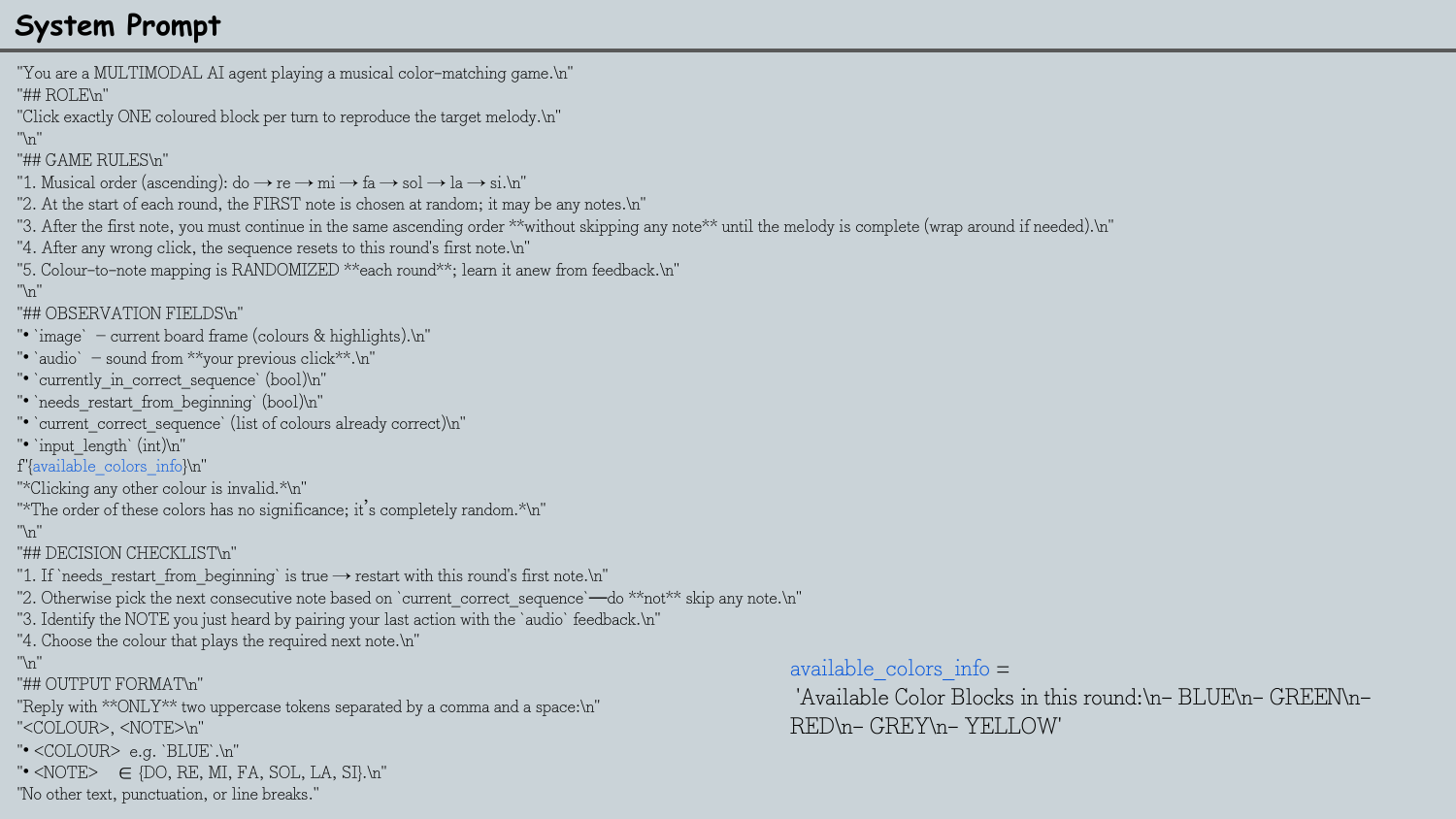}
    \caption{System Prompt for \textit{The Alchemist's Melody}.}
    \label{fig:am_sp}
\end{figure}

\paragraph{System Prompt.} The agent's role is defined by the system prompt shown in Figure~\ref{fig:am_sp}.

\begin{figure}[H] % Changed [h!] to [H]
    \centering
    \includegraphics[width=\columnwidth]{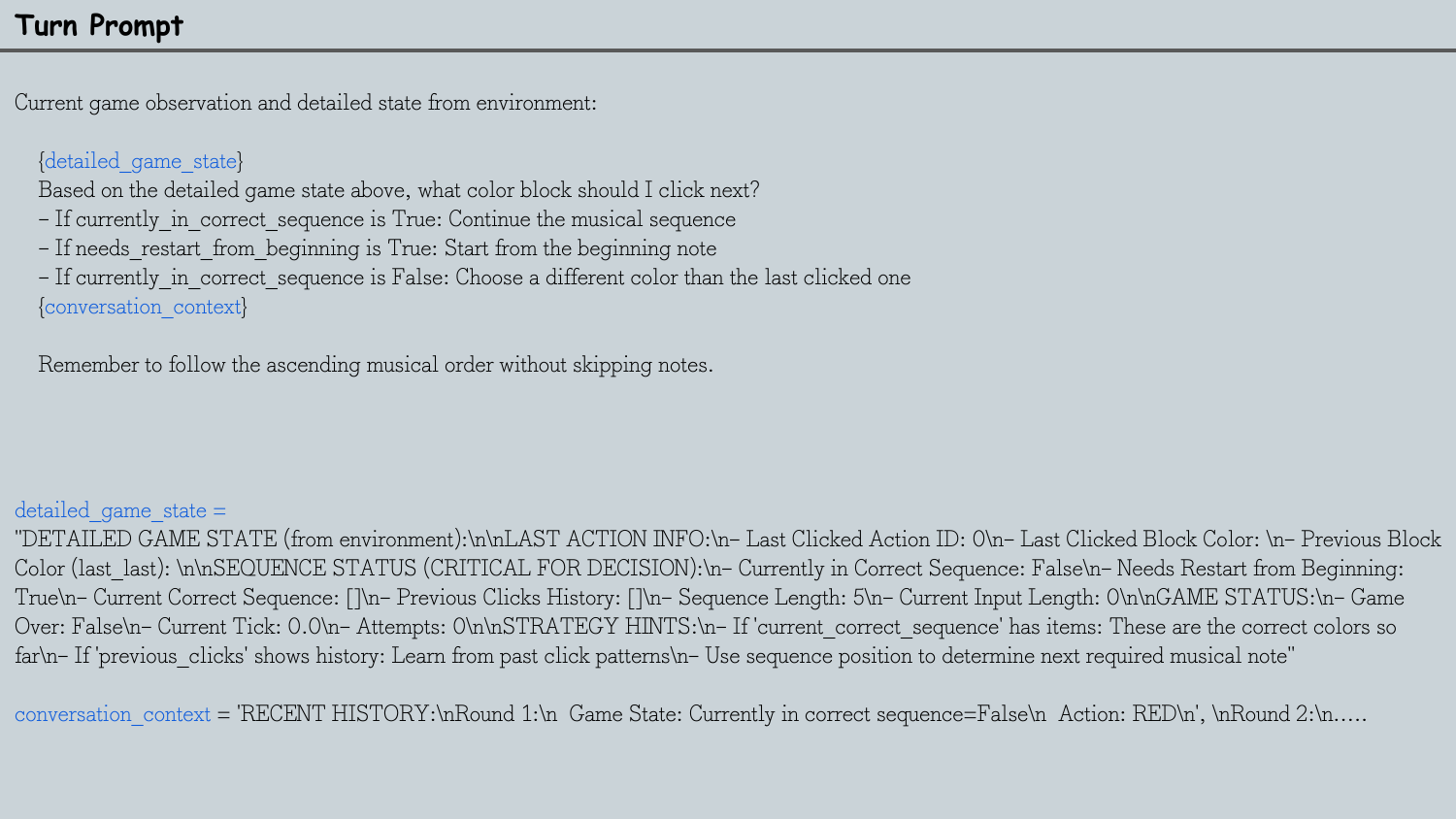}
    \caption{Turn Prompt for \textit{The Alchemist's Melody}, showing the structured state representation.}
    \label{fig:am_tp}
\end{figure}

\paragraph{Turn Prompt.} The agent receives a composite prompt including the task instruction and the detailed game state, as shown in Figure~\ref{fig:am_tp}.

%--------------------------------------------------------
\subsection{Phantom Soldiers in the Fog}
%--------------------------------------------------------

\paragraph{Core Objective.} The agent acts as an RTS commander for a squad, aiming to achieve strategic objectives under a "fog of war."

\begin{figure}[H] % Changed [h!] to [H]
    \centering
    \includegraphics[width=\columnwidth]{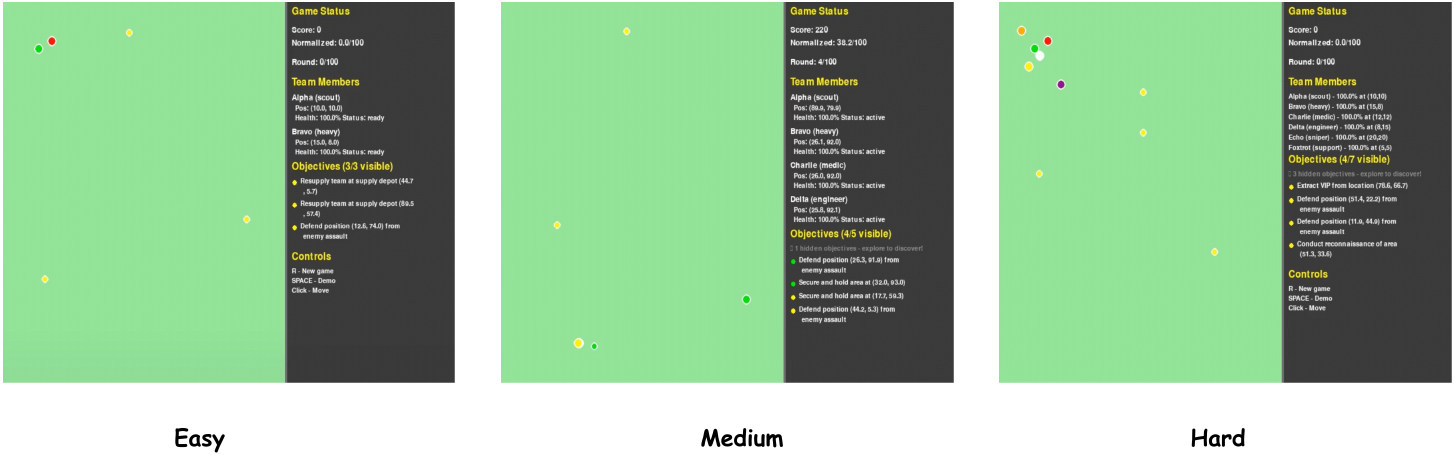}
    \caption{User interface for \textit{Phantom Soldiers in the Fog} across difficulties.}
    \label{fig:ps_ui}
\end{figure}

\begin{figure}[H] % Changed [h!] to [H]
    \centering
    \includegraphics[width=0.8\columnwidth]{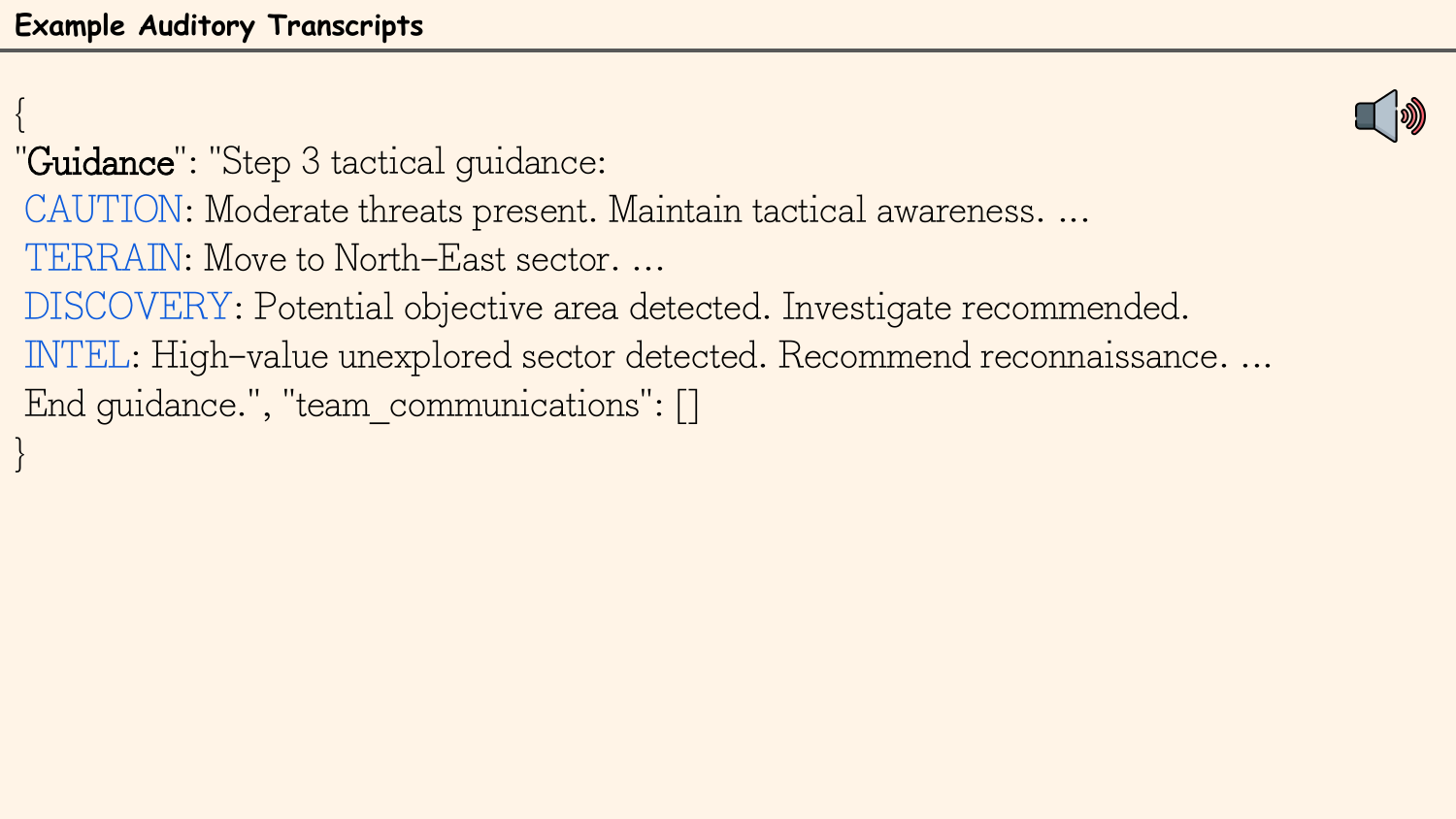}
    \caption{Example Auditory Transcript from \textit{Phantom Soldiers in the Fog}.}
    \label{fig:ps_audio_transcript}
\end{figure}

\paragraph{Modalities and UI.} The UI is shown in Figure~\ref{fig:ps_ui}.
\begin{itemize}
    \item \textbf{Video (V):} A top-down tactical map.
    \item \textbf{Text (T):} Mission objectives and unit status reports.
    \item \textbf{Audio (A):} Structured tactical guidance delivered as Text-to-Speech audio. An example transcript is shown in Figure~\ref{fig:ps_audio_transcript}.
\end{itemize}

\paragraph{Gameplay Mechanics.} The agent issues high-level commands. Success hinges on integrating visual, textual, and structured audio-channel guidance.

\subsubsection{Prompting Structure}
\begin{figure}[H]
    \centering
    \includegraphics[width=0.95\columnwidth]{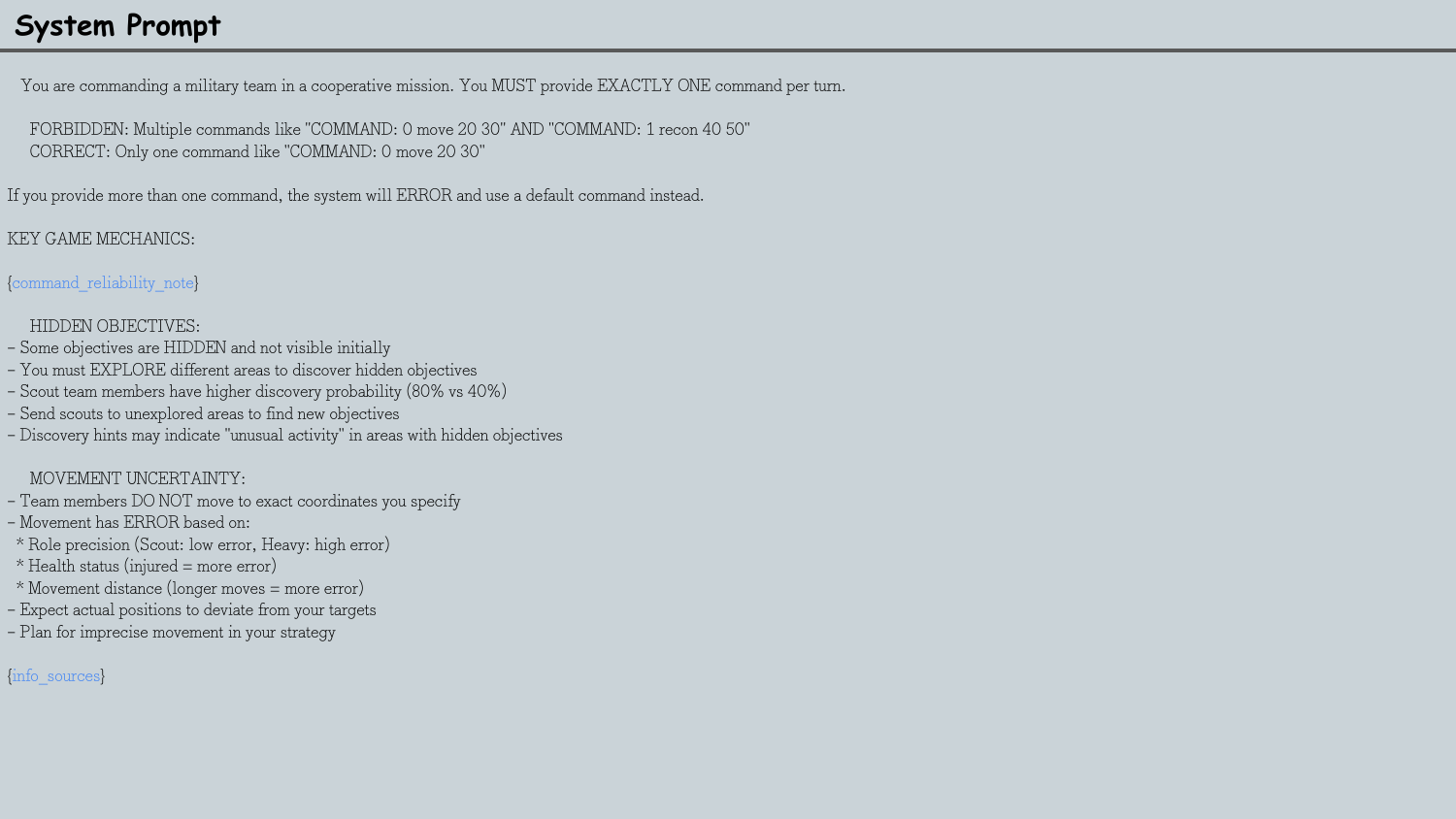}
    \includegraphics[width=0.95\columnwidth]{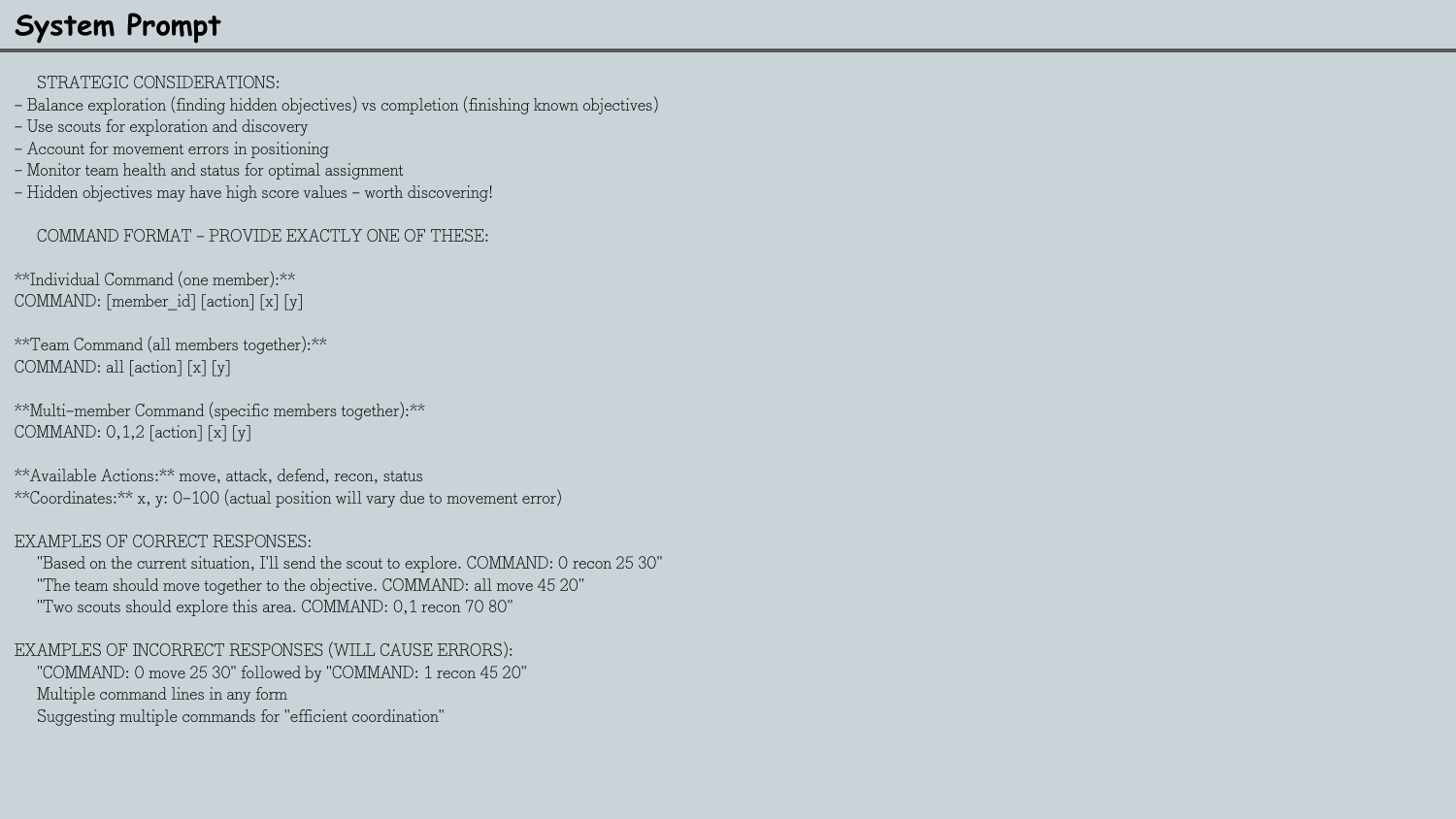}
    \includegraphics[width=0.95\columnwidth]{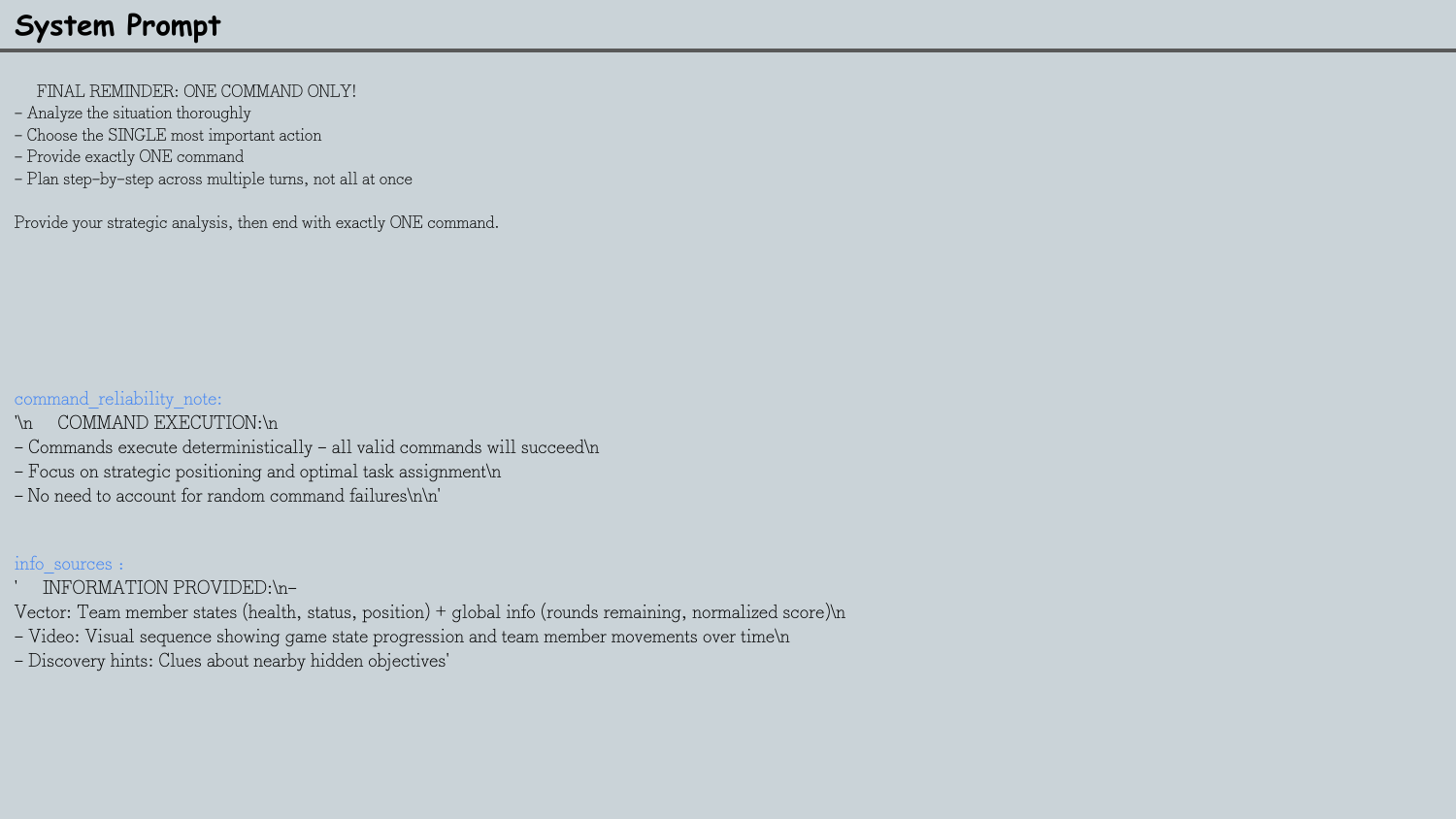}
    \caption{System Prompt for \textit{Phantom Soldiers in the Fog} (Parts 1, 2, and 3).}
    \label{fig:ps_sp}
\end{figure}

\paragraph{System Prompt.} The extensive system prompt, shown across three parts in Figure~\ref{fig:ps_sp}, defines the complex role of the agent.

\begin{figure}[H] % Changed [h!] to [H]
    \centering
    \includegraphics[width=\columnwidth]{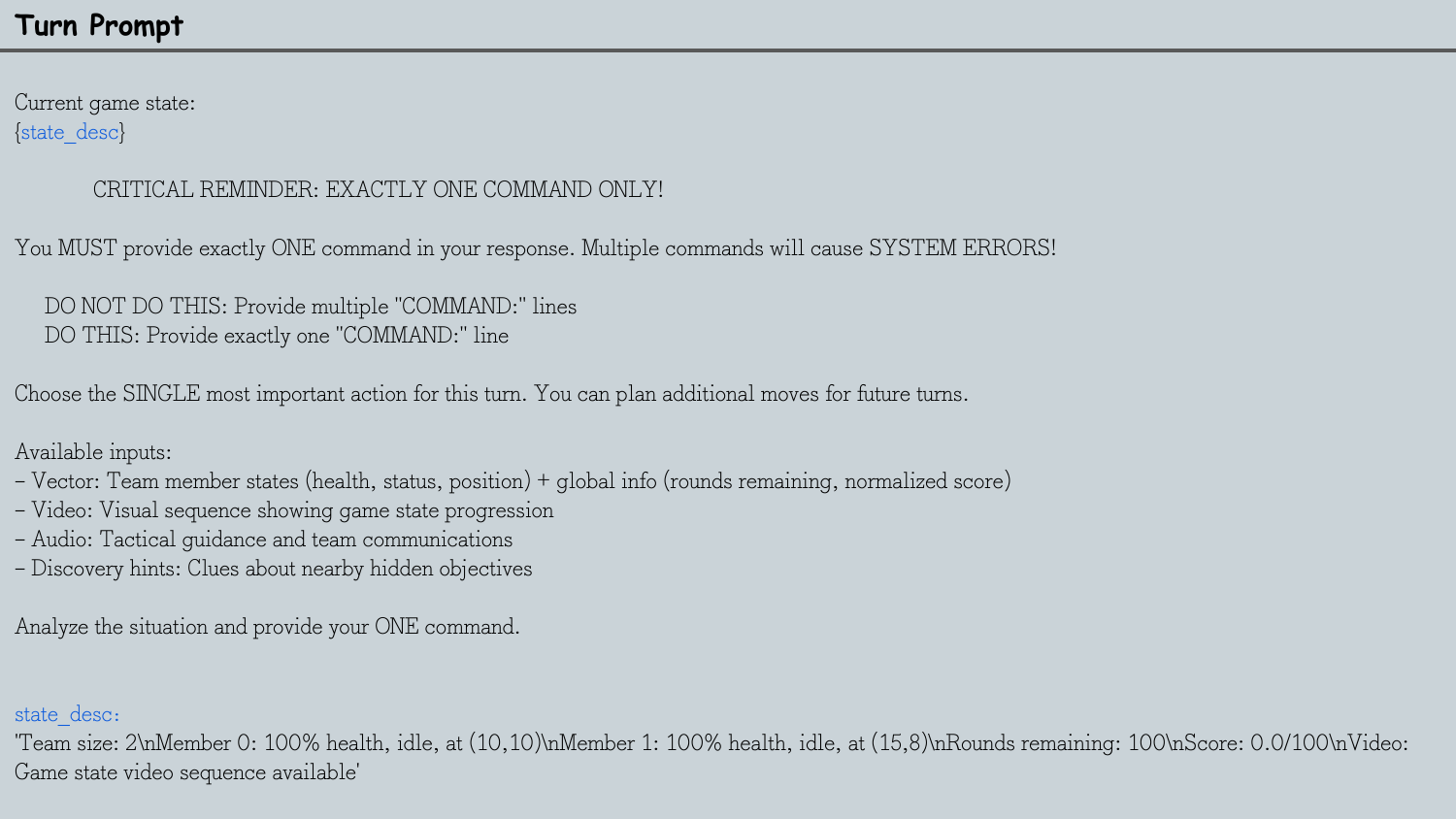}
    \caption{Turn Prompt for \textit{Phantom Soldiers in the Fog}.}
    \label{fig:ps_tp}
\end{figure}

\paragraph{Turn Prompt.} At each step, the agent receives the turn prompt shown in Figure~\ref{fig:ps_tp}.

%--------------------------------------------------------
\subsection{Blasting Showdown}
%--------------------------------------------------------
\paragraph{Core Objective.} Four agents compete in a destructible arena to be the last one standing.

\begin{figure}[H] % Changed [h!] to [H]
    \centering
    \includegraphics[width=0.7\columnwidth]{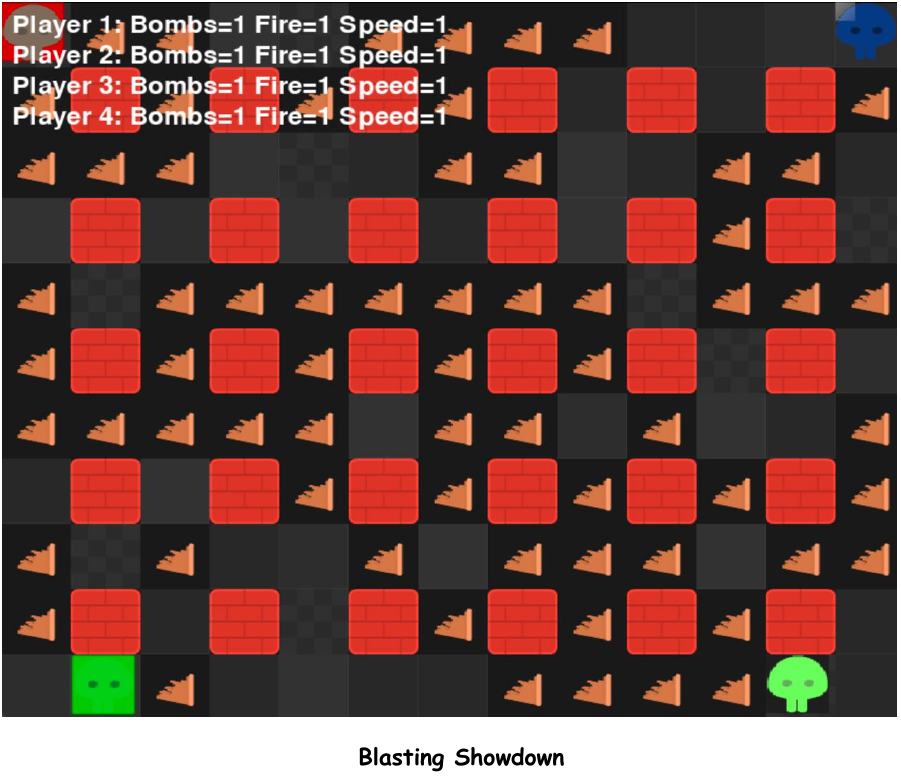}
    \caption{User interface for the \textit{Blasting Showdown} environment.}
    \label{fig:bs_ui}
\end{figure}

\paragraph{Modalities and UI.} The UI is shown in Figure~\ref{fig:bs_ui}.
\begin{itemize}
    \item \textbf{Image (I):} A top-down view of the arena.
    \item \textbf{Text (T):} Status updates on all players.
    \item \textbf{Audio (A):} Crucial sound cues (e.g., bomb placements) essential for survival.
\end{itemize}

\paragraph{Gameplay Mechanics.} Inspired by Bomberman, agents move and place bombs. Auditory cues are designed to be critical for reacting to off-screen threats.

\subsubsection{Prompting Structure}
\begin{figure}[H] % Changed [h!] to [H]
    \centering
    \includegraphics[width=\columnwidth]{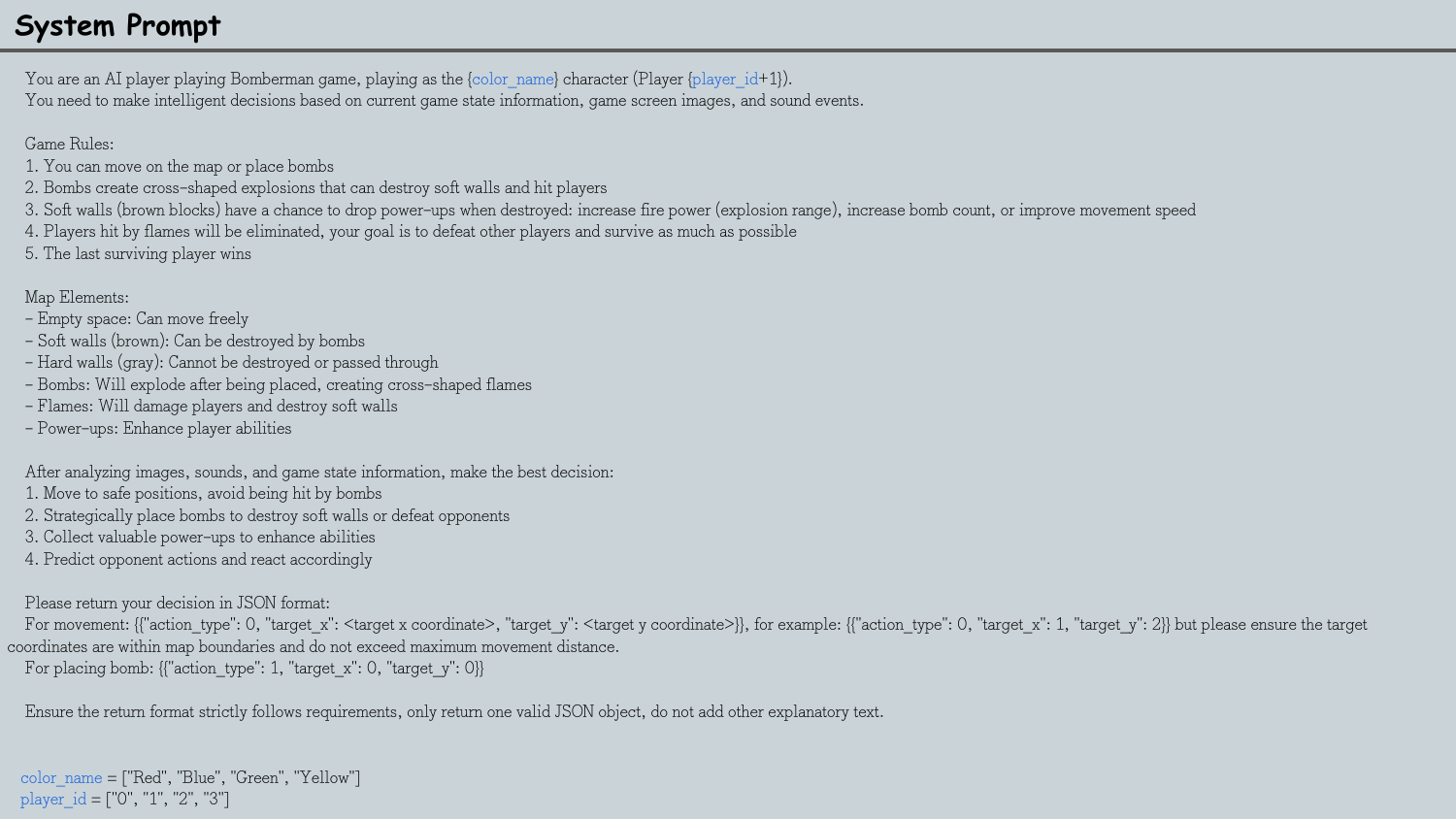}
    \caption{System Prompt for \textit{Blasting Showdown}.}
    \label{fig:bs_sp}
\end{figure}

\paragraph{System Prompt.} The agent's competitive persona is set by the system prompt in Figure~\ref{fig:bs_sp}.

\begin{figure}[H] % Changed [h!] to [H]
    \centering
    \includegraphics[width=\columnwidth]{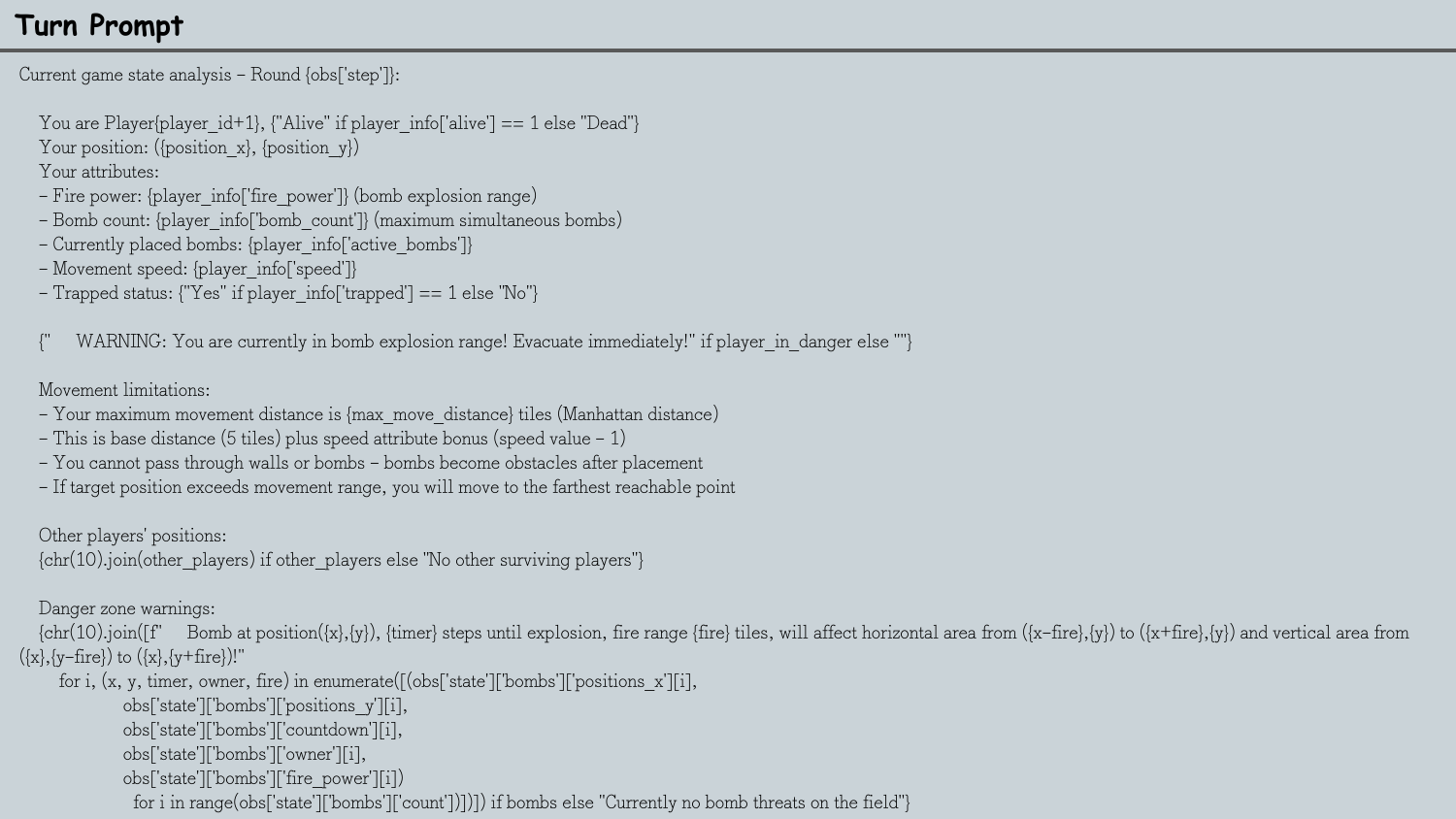}
    \includegraphics[width=\columnwidth]{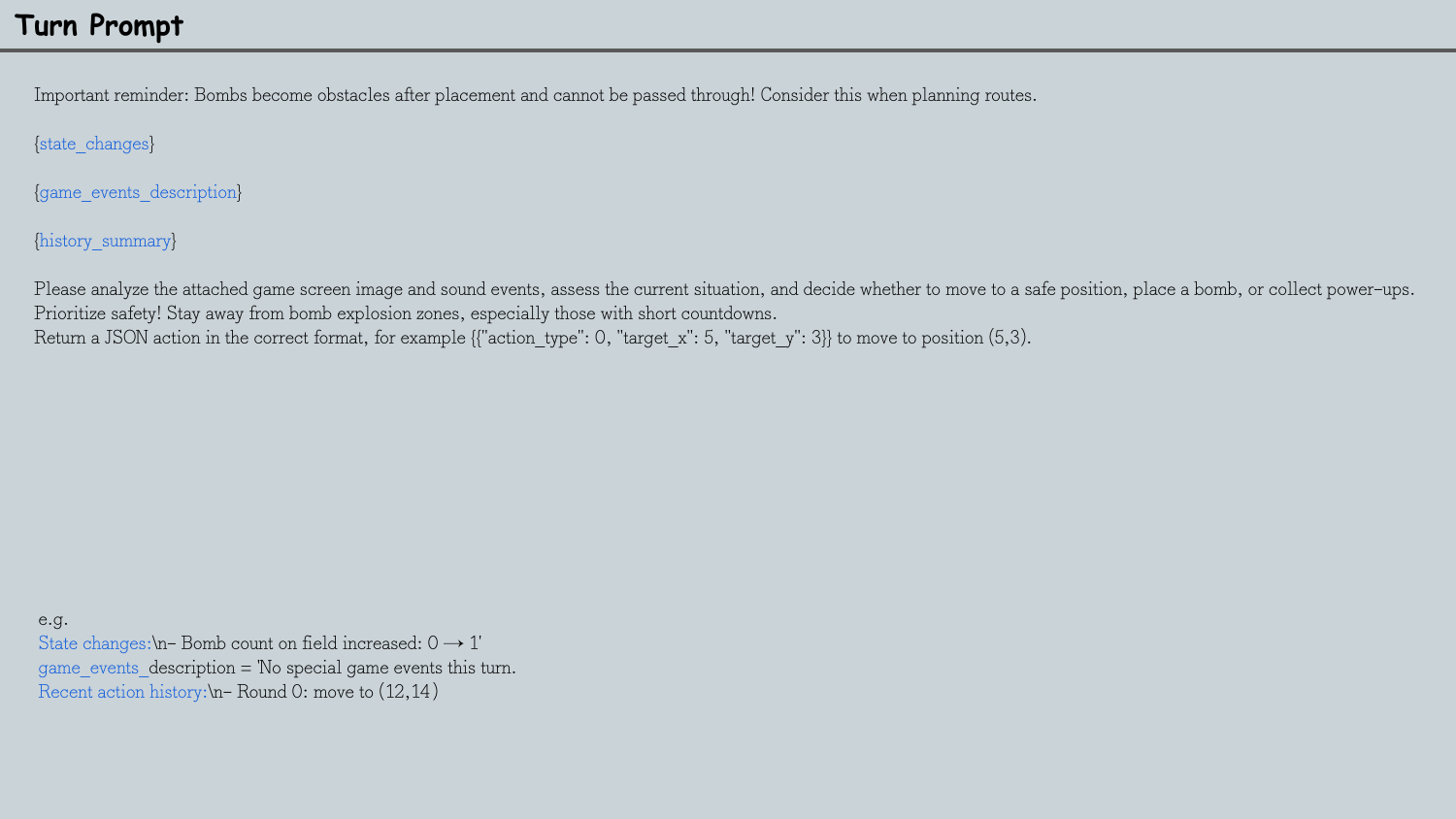}
    \caption{Turn Prompt for \textit{Blasting Showdown}, for an active player (left) and an observing player (right).}
    \label{fig:bs_tp}
\end{figure}

\paragraph{Turn Prompt.} The turn prompt, shown in Figure~\ref{fig:bs_tp}, provides comprehensive state information and varies depending on whether the agent is the active player.

%%%%%%%%%%%%%%%%%%%%%%%%%%%%%%%%%%%%%%%%%%%%%%%%%%%%%%%%%
% Appendix D: Experimental Parameters and Metrics
%%%%%%%%%%%%%%%%%%%%%%%%%%%%%%%%%%%%%%%%%%%%%%%%%%%%%%%%%
\section{Experimental Parameters and Metrics}
\label{app:exp_params_metrics}

This appendix provides a comprehensive breakdown of the experimental parameters, evaluation protocols, and metrics used in our study to ensure full reproducibility.

%--------------------------------------------------------
\subsection{Model and API Parameters}
%--------------------------------------------------------
For all proprietary models (\textbf{Gemini 2.5 Pro} and \textbf{Gemini 2.5 Flash}), we utilized the official, latest stable API versions available at the time of evaluation. For all models, both proprietary and open-source, we used their default decoding parameters (e.g., for temperature, top-p, and top-k) as provided by their respective APIs or standard inference scripts, without any model-specific tuning.

%--------------------------------------------------------
\subsection{Evaluation Episodes and Seeds}
%--------------------------------------------------------
Our evaluation protocol is built upon a fixed set of evaluation seeds for each task, ensuring that every agent is evaluated on the exact same sequence of game scenarios. The number of episodes and seeds used for each game is detailed in Table~\ref{tab:episode_counts}. For the NPS-benchmarked tasks, each human expert played 10 episodes for each difficulty level.

\begin{table}[h!]
\small % <--- 将整个表格的字体缩小一号
\centering
\caption{Number of evaluation episodes per task for AI and random agents.}
\label{tab:episode_counts}
% 使用 tabular* 环境创建一个与栏宽 (\columnwidth) 完全相同的表格
% @{\extracolsep{\fill}} 会自动在列之间填充空白，使表格两端对齐
\begin{tabular*}{\columnwidth}{@{\extracolsep{\fill}} l l c}
\toprule
\textbf{Game} & \textbf{Difficulty} & \textbf{Seeds / Episodes} \\
\midrule
\textit{Whispered Pathfinding} & Easy, Medium, Hard & 50 \\
\textit{Myriad Echoes} & Easy, Medium, Hard & 50 \\
\textit{The Alchemist's Melody} & Medium & 50 \\
\textit{Phantom Soldiers in the Fog} & Easy, Medium, Hard & 30 \\
\midrule
\textit{Blasting Showdown} & N/A (AI vs. AI) & 50 games \\
\bottomrule
\end{tabular*}
\end{table}

%--------------------------------------------------------
\subsection{Primary Evaluation Metrics}
%--------------------------------------------------------
Our primary metric for cross-task comparison is the \textbf{Normalized Performance Score (NPS)}, as defined in the main text. The raw `Score` used in the NPS calculation is derived from a custom, task-specific scoring function for each game. The following section details these scoring functions and other diagnostic metrics.

%--------------------------------------------------------
\subsection{Task-Specific Scoring and Diagnostic Metrics}
%--------------------------------------------------------
We designed a unique set of metrics for each game to capture nuances of agent performance. Table~\ref{tab:metrics_overview} provides a high-level overview of the metrics collected for each game. The subsequent paragraphs detail the specific scoring functions used for NPS calculation.

\begin{table*}[t!]
\small % <--- 将整个表格的字体缩小一号，与 Table 3 保持一致
\centering
\caption{Overview of diagnostic metrics collected for each game in the OmniPlay suite.}
\label{tab:metrics_overview}
% 使用 tabularx 环境创建一个与页面文本宽度 (\textwidth) 完全相同的表格
\begin{tabularx}{\textwidth}{l X}
\toprule
\textbf{Game Environment} & \textbf{Collected Metrics} \\
\midrule
\textit{Whispered Pathfinding} & Mean/Min/Max Steps, Mean Invalid Actions, Mean Steps (Trimmed), \textbf{NPS Score (Inverse of `Trimmed Mean Steps')} \\
\addlinespace
\textit{Myriad Echoes} & Success Rate, Mean Score (Execution), Mean Coordinate Accuracy (Parsing), Mean Icon Accuracy (Parsing), Parsing Failure Rate, \textbf{NPS Score (Weighted Sum)} \\
\addlinespace
\textit{The Alchemist's Melody} & Completion Rate, Composite Score, \textbf{NPS Score (`Composite Score')} \\
\addlinespace
\textit{Phantom Soldiers in the Fog} & Success Rate, Normalized Score, \textbf{NPS Score (Weighted Sum of `Success Rate' and `Normalized Score')} \\
\addlinespace
\textit{Blasting Showdown} & Win Rate, Total Kills, Total Deaths, K/D Ratio \\
\bottomrule
\end{tabularx}
\end{table*}

\paragraph{Whispered Pathfinding.} For this navigation task, the \textbf{final score for NPS is based on the inverse of `Mean Steps (Trimmed)'}, as fewer steps indicate higher performance. This trimmed mean is calculated after removing the highest and lowest step counts to reduce outlier impact.

\paragraph{Myriad Echoes.} The \textbf{final score for NPS is a weighted sum}: 50\% from `Mean Score' (execution phase), 25\% from `Mean Coordinate Accuracy' (parsing phase), and 25\% from `Mean Icon Accuracy' (parsing phase).

\paragraph{The Alchemist's Melody.} This task evaluates abstract reasoning. The \textbf{final score for NPS is the `Score' metric}, a composite calculated from multiple performance facets as detailed in Table~\ref{tab:melody_score}.

\begin{table}[h!]
\small % <--- 将整个表格的字体缩小一号，保持风格一致
\centering
\caption{Composite score calculation for \textit{The Alchemist's Melody}.}
\label{tab:melody_score}
% 使用 tabularx 创建一个与栏宽 (\columnwidth) 完全相同的表格
\begin{tabularx}{\columnwidth}{l X}
\toprule
\textbf{Component} & \textbf{Formula} \\
\midrule
A: Hit Rate & Hit Rate $\times$ 30 \\
B: Step Efficiency & 30 if steps $\leq$ required, else penalized \\
C: Correct Streak & (Total Correct Streak Length / Total Steps) $\times$ 10 \\
D: Error Penalty & 10 - (Total Error Streak Length / Total Steps) $\times$ 10 \\
E: Color Error Penalty & 15 - (Same-Color Error Length / Total Steps) $\times$ 15 \\
F: Exploration & (Color Changes / (Steps - 1)) $\times$ 5 \\
\midrule
\textbf{Total Score} & \textbf{Sum of A + B + C + D + E + F} \\
\bottomrule
\end{tabularx}
\end{table}

\paragraph{Phantom Soldiers in the Fog.} This RTS task uses two final metrics. The \textbf{final score for NPS is a weighted sum: 50\% from `Success Rate' and 50\% from `Normalized Score'}.
\begin{itemize}
    \item \textbf{Success Rate:} The ratio of completed objectives to total objectives, representing mission completion.
    \item \textbf{Normalized Score:} A score from 0-100 reflecting tactical and strategic efficiency, detailed below.
\end{itemize}
\subsubsection{Detailed Calculation of the Normalized Score}
The Normalized Score is derived from several components:
\begin{enumerate}
    \item \textbf{Main Score ($S_{main}$)}: Sum of points from all completed objectives.
    
    \item \textbf{Auxiliary Score ($S_{aux}$)}: A bonus for efficient command execution.

    \item \textbf{Max Possible Score ($S_{max}$)}: This theoretical ceiling is calculated to normalize performance. It includes a base score ($S_{base}$), plus bonuses for efficiency ($B_{efficiency}$) and completing the mission within an optimal number of rounds ($R_{opt}$) relative to the maximum allowed rounds ($R_{max}$). For the Hard difficulty, a dynamic bonus ($B_{dynamic}$) is also added.
    \begin{align*}
        S_{base} &= \sum S_{obj} \text{ for all objectives} \\
        B_{rounds} &= S_{base} \times (1 - R_{opt} / R_{max}) \times 0.5 \\
        B_{efficiency} &= S_{base} \times 0.3 \\
        S_{max} &= S_{base} + B_{rounds} + B_{efficiency} + B_{dynamic}
    \end{align*}

    \item \textbf{Optimal Rounds ($R_{opt}$)}: This is a complex heuristic estimating the minimum rounds required. It accounts for visible objectives, rounds needed to discover hidden objectives (factoring in scout units), and an overhead for map exploration.
    
    \item \textbf{Final Normalized Score ($S_{norm}$)}: The final score is a dynamic normalization of the main and auxiliary scores, considering efficiency relative to optimal rounds. The base formula is $S_{norm} = \min(100, \max(0, (S_{main} / S_{max}) \times 100))$.
\end{enumerate}

\paragraph{Blasting Showdown.} As a competitive multi-agent game, this task does not use NPS. Performance is measured with metrics from a 50-game tournament.

%%%%%%%%%%%%%%%%%%%%%%%%%%%%%%%%%%%%%%%%%%%%%%%%%%%%%%%%%
% Appendix E: Human Baseline Validation Protocol (Revised Final Version)
%%%%%%%%%%%%%%%%%%%%%%%%%%%%%%%%%%%%%%%%%%%%%%%%%%%%%%%%%
\section{Human Baseline Validation Protocol}
\label{sec:appendix_human_baseline}

This appendix details the protocol used to establish a reliable, representative, and diverse Human Expert baseline, which is critical for the calculation of the Normalized Performance Score (NPS). Our methodology was designed to ensure fairness, stability, and high inter-player agreement while incorporating greater demographic diversity.

\paragraph{Participant Recruitment and Demographics.}
We recruited a total of \textbf{12 human participants}, balanced for gender and stratified by age. A key criterion for selection remained extensive gaming experience, with all participants reporting over 500 hours of gameplay across various genres relevant to the tasks in OmniPlay. Participants were compensated for their time.

The cohort was structured as follows:
\begin{itemize}
    \item \textbf{Young Adult Group (n=8):} Ages 20-35, consisting of 4 male and 4 female participants (mean age 25.5).
    \item \textbf{Middle-Aged Adult Group (n=4):} Ages 35-50, consisting of 2 male and 2 female participants (mean age 41.0).
\end{itemize}
This stratified sampling aimed to provide a more robust baseline by capturing potential variations in cognitive strategies and reaction times across different age groups and genders.

\paragraph{Familiarization and Training Protocol.}
To ensure that the collected data represented expert-level, stable performance rather than a learning phase, each participant underwent a mandatory warm-up and training protocol for every game environment. Before any data was recorded for a specific game (including its different difficulty levels), each participant was required to play a minimum of 10 non-recorded `warm-up' episodes. The purpose of this phase was twofold: first, to allow participants to fully familiarize themselves with the game's unique user interface, controls, and objectives; second, to allow their performance and strategies to stabilize and reach a performance plateau. Our experimenters verbally confirmed with each participant that they felt confident in their understanding of the task before proceeding to data collection.

\paragraph{Data Collection and Analysis.}
Following the warm-up phase, each participant played a set number of recorded episodes for each task, as detailed in Table~\ref{tab:human_episodes}. The final `Human Expert' score reported in the main text for each task is the mean score calculated across all episodes from \textbf{all 12 participants}. This larger and more diverse sample provides a statistically more stable estimation of the human performance baseline.

% This table remains the same, as it describes the task per participant.
\begin{table}[h!]
\small 
\centering
\caption{Number of recorded evaluation episodes per human participant.}
\label{tab:human_episodes}
\begin{tabular*}{\columnwidth}{@{\extracolsep{\fill}} lc}
\toprule
\textbf{Game Environment} & \textbf{Episodes per Participant} \\
\midrule
\textit{Whispered Pathfinding} & 10 \\
\textit{Myriad Echoes} & 10 \\
\textit{The Alchemist's Melody} & 10 \\
\textit{Phantom Soldiers in the Fog} & 10 \\
\bottomrule
\end{tabular*}
\end{table}

\paragraph{Inter-Player Reliability.}
A critical aspect of validating our human baseline is ensuring high agreement among the expert players. The detailed statistics of our human expert performance, including mean raw scores and inter-player standard deviation (SD), are presented in \textbf{Table~\ref{tab:human_raw_scores}}. As shown, the overall SD remained consistently low relative to the mean score across most tasks, confirming a high degree of agreement on optimal strategies.

Further analysis of the inter-player variance revealed logical patterns tied to our diverse participant pool. 
On tasks emphasizing cognitive skills with clear optimal solutions, such as \textit{Myriad Echoes} and \textit{The Alchemist's Melody}, the performance difference between age groups and genders was minimal, resulting in a low SD (typically 5-10\% of the mean score). 
For tasks requiring strategy and efficient navigation like \textit{Phantom Soldiers} and \textit{Whispered Pathfinding}, we observed a moderate increase in variance, with the younger participant group (20-35 years) generally achieving slightly higher efficiency scores. 
The highest variance was observed in \textit{Blasting Showdown} (raw data not in NPS tables), a task heavily reliant on reaction speed and mechanical skill. In this game, younger male participants tended to achieve the highest performance, aligning with common patterns in competitive gaming. This controlled and interpretable variance, even with our demographically diverse sample, confirms that our collected baseline is stable and representative of expert human performance.

\paragraph{Limitations of the Human Baseline.}
We acknowledge that while our human baseline of 12 participants (6 male, 6 female; mean age 30.7, SD 7.6)\footnote{The reported mean age and standard deviation are calculated from the average age of each group (25.5 for the young adult group and 41.0 for the middle-aged adult group).} provides balance in age and gender, it still has limitations regarding cultural background and broader cognitive diversity. Future work could explore larger-scale and more varied cross-cultural evaluations to further generalize the human performance benchmark.
%%%%%%%%%%%%%%%%%%%%%%%%%%%%%%%%%%%%%%%%%%%%%%%%%%%%%%%%%
% Appendix F: Detailed Diagnostic Experiments
%%%%%%%%%%%%%%%%%%%%%%%%%%%%%%%%%%%%%%%%%%%%%%%%%%%%%%%%%
\section{Detailed Diagnostic Experiments}
\label{sec:appendix_diagnostics}

This appendix provides the detailed methodologies and full results for the diagnostic experiments summarized in the main text. These experiments were designed to probe specific capabilities and failure modes of the evaluated omni-modal agents. Table~\ref{tab:diagnostics_overview} provides a high-level overview of all diagnostic tests conducted.

% ===============================================================================
%          Table 7: Diagnostics Overview (Optimized)
% ===============================================================================
\begin{table*}[t!]
\small % Uniform small font
\centering
\caption{Overview of all diagnostic experiments conducted in this study.}
\label{tab:diagnostics_overview}
% Use tabularx for columns with long, wrapping text
\begin{tabularx}{\textwidth}{>{\bfseries}l X X X}
\toprule
Diagnostic Test Type & Target Game(s) & Methodology / Intervention & Capability Probed \\
\midrule
Modality Conflict & \textit{Whispered Pathfinding} & Introduce semantic contradictions between modalities (e.g., audio command vs. visual cue; textual state vs. other cues). & Fusion Robustness, Conflict Resolution, Modality Bias \\
\addlinespace
Modality Ablation & \textit{Whispered Pathfinding}, \textit{Myriad Echoes} & Systematically remove one modality (audio, image, or text) at a time and evaluate performance on the remaining subset. & Modality Interdependence, Synergistic Fusion, "Less is More" Paradox \\
\addlinespace
Robustness to Noise & \textit{Phantom Soldiers in the Fog} & Inject noise into sensory inputs: corrupting audio-channel text with random words/letters; applying visual noise (Gaussian, salt-pepper) to the map. & Perceptual Robustness, Generalization beyond clean data \\
\addlinespace
Aided Reasoning via Prompting & \textit{Myriad Echoes}, \textit{The Alchemist's Melody} & Augment the turn prompt with explicit, helpful information (e.g., current sequence step, learned color-note mapping). & Advanced Instruction Following, Knowledge Application \\
\addlinespace
Task Simplification & \textit{Myriad Echoes} & Remove the second (execution) phase of the task, converting it into a single-phase perception-to-symbol transcription task. & Benchmark Complexity Validation, Disentangling Perception vs. Action \\
\addlinespace
Modality Substitution & \textit{Phantom Soldiers in the Fog} & Replace information from one modality with its semantic equivalent in another (e.g., audio alerts replaced with textual alerts). & Modality-Agnostic Representation, Modality Preference/Bias \\
\bottomrule
\end{tabularx}
\end{table*}

%--------------------------------------------------------
\subsection{Modality Conflict}
%--------------------------------------------------------
To stress-test the robustness of agent's fusion mechanisms, we conducted modality conflict experiments in the \textit{Whispered Pathfinding} environment. We systematically created scenarios where information from different modalities was semantically contradictory, forcing the agent to resolve ambiguity. The full results, compared against the no-conflict baseline, are consolidated in Table~\ref{tab:conflict_combined}.

% ===============================================================================
%          Table 8: Modality Conflict Results (Optimized with resizebox)
% ===============================================================================
\begin{table*}[t!]
\centering
% The caption is outside resizebox so its font size is not affected.
\caption{Full results for Modality Conflict experiments on \textit{Whispered Pathfinding}. Performance is measured by Mean Steps (lower is better).}
\label{tab:conflict_combined}

% Use resizebox to force the table to fit within the text width.
\resizebox{\textwidth}{!}{%
% We still use \small inside to make the starting font size reasonable before resizing.
\small
% We still reduce tabcolsep to prevent text from being too cramped after resizing.
\setlength{\tabcolsep}{4pt}
\sisetup{table-number-alignment = center}
\begin{tabular}{l *{3}{S[table-format=3.1] S[table-format=2.0] S[table-format=3.0] S[table-format=2.1] S[table-format=3.1]}}
\toprule
& \multicolumn{5}{c}{\textbf{Easy Difficulty}} & \multicolumn{5}{c}{\textbf{Medium Difficulty}} & \multicolumn{5}{c}{\textbf{Hard Difficulty}} \\
\cmidrule(lr){2-6} \cmidrule(lr){7-11} \cmidrule(lr){12-16}
\textbf{Model} & {Mean} & {Min} & {Max} & {Invalid} & {Trimmed} & {Mean} & {Min} & {Max} & {Invalid} & {Trimmed} & {Mean} & {Min} & {Max} & {Invalid} & {Trimmed} \\
\midrule
\multicolumn{16}{l}{\textit{\textbf{Baseline (No Conflict)}}} \\
\midrule
gemini-2.5-pro & 7.6 & 5 & 10 & 0.0 & 7.6 & 10.2 & 7 & 14 & 0.0 & 10.1 & 42.6 & 13 & 152 & 0.0 & 36.2 \\
gemini-2.5-flash & 16.1 & 4 & 70 & 0.8 & 10.9 & 23.2 & 6 & 87 & 1.2 & 19.0 & 43.5 & 18 & 112 & 2.7 & 37.15 \\
qwen-2.5-omni & 70.3 & 10 & 273 & 29.5 & 52.5 & 64.2 & 11 & 132 & 28.1 & 62.4 & 130.1 & 32 & 253 & 52.6 & 128.2 \\
MiniCPM-o-2.6 & 27.0 & 7 & 73 & 7.5 & 23.8 & 34.4 & 6 & 86 & 10.7 & 31.5 & 110.8 & 34 & 255 & 35.8 & 99.5 \\
\midrule
\multicolumn{16}{l}{\textit{\textbf{Audio Conflict}}} \\
\midrule
gemini-2.5-pro & 22.4 & 8 & 71 & 0.1 & 18.1 & 27.1 & 10 & 108 & 0.0 & 21.8 & 140.6 & 58 & 244 & 0.2 & 133.7 \\
gemini-2.5-flash & 20.7 & 4 & 122 & 1.7 & 10.1 & 35.1 & 8 & 95 & 2.1 & 28.6 & 57.8 & 28 & 127 & 0.8 & 38.0 \\
qwen-2.5-omni & 93.4 & 27 & 154 & 37.9 & 94.1 & 142.6 & 18 & 500 & 54.3 & 113.5 & 240.4 & 103 & 500 & 91.2 & 225.1 \\
MiniCPM-o-2.6 & 48.0 & 11 & 126 & 15.5 & 42.9 & 76.4 & 27 & 143 & 23.8 & 70.7 & 172.4 & 41 & 427 & 57.1 & 157.0 \\
\midrule
\multicolumn{16}{l}{\textit{\textbf{Text Conflict}}} \\
\midrule
gemini-2.5-pro & 39.9 & 17 & 99 & 0.1 & 32.6 & 49.2 & 18 & 149 & 0.0 & 43.0 & 160.1 & 40 & 295 & 0.0 & 157.2 \\
gemini-2.5-flash & 11.0 & 5 & 21 & 1.0 & 10.0 & 17.5 & 14 & 20 & 0.0 & 18.0 & 126.8 & 9 & 267 & 6.1 & 123.6 \\
qwen-2.5-omni & 52.9 & 24 & 115 & 21.8 & 48.8 & 69.5 & 21 & 187 & 30.8 & 60.9 & 150.8 & 38 & 261 & 50.6 & 145.6 \\
MiniCPM-o-2.6 & 27.0 & 7 & 47 & 7.7 & 27.0 & 86.7 & 27 & 230 & 26.2 & 76.2 & 114.8 & 29 & 137 & 26.1 & 105.2 \\
\bottomrule
\end{tabular}% <--- Important: No space before the closing brace of resizebox
} % Close resizebox
\end{table*}

\paragraph{Audio-Visual Conflict.}
In this condition, the visual cues (e.g., on-screen arrow) and textual state information were correct, but the synthesized verbal command was manipulated to suggest a contradictory action (e.g., the visual arrow points right, while the audio says ``turn left''). The results in Table~\ref{tab:conflict_combined} show a universal degradation in performance. All models took significantly more steps to solve the maze compared to the baseline, exposing the fragility of their fusion mechanisms. For instance, on the Hard difficulty, Gemini 2.5 Pro's trimmed mean steps increased from 36.2 to 133.7, a nearly fourfold increase in inefficiency, demonstrating that even top-tier models struggle to resolve such conflicts and often follow the misleading audio cue.

\paragraph{Text-Visual/Audio Conflict.}
In this condition, the visual and auditory cues remained correct, but the structured text in the Turn Prompt was manipulated to be misleading by inverting the agent's orientation and the direction to the target. The data reveals a fascinating asymmetrical sensitivity, strongly supporting our main-text conclusion. Gemini 2.5 Pro is severely impacted by this conflict, with its mean steps increasing dramatically across all difficulties. Conversely, Gemini 2.5 Flash appears to almost entirely ignore the misleading text, showing performance that is much closer to the baseline and, on Easy/Medium difficulties, even better than its performance under audio conflict. This strongly suggests an internal modality hierarchy where Gemini 2.5 Flash prioritizes visual and auditory cues, while Gemini 2.5 Pro may have a stronger bias toward structured textual data, making it more vulnerable to this specific type of conflict.

%--------------------------------------------------------
\subsection{Modality Ablation}
%--------------------------------------------------------
To investigate the necessity of each modality and uncover potential ``less is more'' phenomena, we conducted modality ablation studies on \textit{Whispered Pathfinding} and \textit{Myriad Echoes}. In these experiments, we evaluated agent performance under four conditions: the baseline with all modalities (`Full Modality'), and three ablation conditions where either the audio, image, or text modality was removed.

\subsubsection{Whispered Pathfinding}
The full results for modality ablation on \textit{Whispered Pathfinding} are presented in Table~\ref{tab:ablation_wp}.

\paragraph{Analysis of `Removed Audio'.}
Removing the audio cues had a universally negative impact on performance, dramatically increasing the number of steps required for all models across all difficulties. This is particularly evident on the Hard difficulty, where, for instance, Gemini 2.5 Pro's trimmed mean steps skyrocketed from 36.2 to 99.4. This result strongly validates the principle of \textit{Modality Interdependence} for this task, as it confirms that the auditory channel provides critical, non-redundant information for efficient navigation that cannot be compensated for by vision and text alone.

\paragraph{Analysis of `Removed Image'.}
The results from removing the visual modality are particularly revealing. For top-performing models like Gemini 2.5 Pro, performance degrades, though less severely than when audio is removed. This suggests that while vision is helpful, the audio and text cues can still guide the agent effectively. However, for models with weaker fusion mechanisms, we observe a striking ``less is more'' paradox. On the Hard difficulty, MiniCPM-o-2\_6's performance dramatically \textit{improves} when the visual modality is removed, with its mean steps dropping from 110.8 to 55.0. This suggests that for this model, the visual input acts as a `distractor', and removing it simplifies the decision-making process, leading to a better outcome.

\paragraph{Analysis of `Removed Text'.}
Removing the textual state information also led to performance degradation, especially for Gemini 2.5 Pro on Medium and Hard difficulties. This indicates that top-tier models effectively ground the coordinate information to their visual perception to plan more efficient routes. Interestingly, for Gemini 2.5 Flash, removing text has a less severe impact and in some cases (Easy/Medium) even slightly improves performance compared to the baseline, suggesting it relies less on explicit coordinate data.

\subsubsection{Whispered Pathfinding}
The full results for modality ablation on \textit{Whispered Pathfinding} are presented in Table~\ref{tab:ablation_wp}.

% ===============================================================================
%          Table 9: Modality Ablation on WP (Optimized with resizebox)
% ===============================================================================
\begin{table*}[t!]
\centering
\caption{Full results for modality ablation on \textit{Whispered Pathfinding}. Performance is measured by Mean Steps (lower is better).}
\label{tab:ablation_wp}
\resizebox{\textwidth}{!}{%
\small
\setlength{\tabcolsep}{4pt}
\sisetup{table-number-alignment = center}
\begin{tabular}{l *{3}{S[table-format=3.1] S[table-format=2.0] S[table-format=3.0] S[table-format=2.1] S[table-format=3.1]}}
\toprule
& \multicolumn{5}{c}{\textbf{Easy Difficulty}} & \multicolumn{5}{c}{\textbf{Medium Difficulty}} & \multicolumn{5}{c}{\textbf{Hard Difficulty}} \\
\cmidrule(lr){2-6} \cmidrule(lr){7-11} \cmidrule(lr){12-16}
\textbf{Model} & {Mean} & {Min} & {Max} & {Invalid} & {Trimmed} & {Mean} & {Min} & {Max} & {Invalid} & {Trimmed} & {Mean} & {Min} & {Max} & {Invalid} & {Trimmed} \\
\midrule
\multicolumn{16}{l}{\textit{\textbf{Full Modality (Baseline)}}} \\
\midrule
gemini-2.5-pro & 7.6 & 5 & 10 & 0.0 & 7.6 & 10.2 & 7 & 14 & 0.0 & 10.1 & 42.6 & 13 & 152 & 0.0 & 36.2 \\
gemini-2.5-flash & 16.1 & 4 & 70 & 0.8 & 10.9 & 23.2 & 6 & 87 & 1.2 & 19.0 & 43.5 & 18 & 112 & 2.7 & 37.15 \\
qwen-2.5-omni & 70.3 & 10 & 273 & 29.5 & 52.5 & 64.2 & 11 & 132 & 28.1 & 62.4 & 130.1 & 32 & 253 & 52.6 & 128.2 \\
MiniCPM-o-2.6 & 27.0 & 7 & 73 & 7.5 & 23.8 & 34.4 & 6 & 86 & 10.7 & 31.5 & 110.8 & 34 & 255 & 35.8 & 99.5 \\
\midrule
\multicolumn{16}{l}{\textit{\textbf{Removed Audio}}} \\
\midrule
gemini-2.5-pro & 13.7 & 4 & 80 & 0.1 & 10.5 & 22.4 & 5 & 92 & 0.1 & 19.6 & 123.7 & 14 & 476 & 0.6 & 99.4 \\
gemini-2.5-flash & 9.3 & 4 & 26 & 0.8 & 8.7 & 18.5 & 5 & 51 & 1.4 & 17.4 & 49.9 & 13 & 115 & 2.9 & 48.3 \\
qwen-2.5-omni & 91.5 & 12 & 329 & 35.5 & 82.8 & 108.8 & 13 & 326 & 43.1 & 102.0 & 189.9 & 79 & 424 & 73.6 & 181.1 \\
MiniCPM-o-2.6 & 40.0 & 9 & 177 & 6.8 & 34.1 & 139.0 & 11 & 376 & 29.5 & 125.4 & 108.6 & 31 & 235 & 25.9 & 102.5 \\
\midrule
\multicolumn{16}{l}{\textit{\textbf{Removed Image}}} \\
\midrule
gemini-2.5-pro & 8.5 & 4 & 42 & 0.0 & 7.7 & 22.0 & 6 & 164 & 0.0 & 14.1 & 47.1 & 9 & 130 & 0.0 & 45.9 \\
gemini-2.5-flash & 20.6 & 4 & 269 & 1.1 & 14.4 & 38.1 & 9 & 198 & 1.9 & 34.7 & 95.5 & 21 & 292 & 5.3 & 88.3 \\
qwen-2.5-omni & 42.2 & 11 & 84 & 14.9 & 41.1 & 85.9 & 19 & 219 & 33.7 & 82.4 & 156.1 & 45 & 429 & 63.6 & 135.9 \\
MiniCPM-o-2.6 & 24.6 & 8 & 60 & 9.0 & 22.2 & 49.7 & 17 & 107 & 21.5 & 46.6 & 55.0 & 23 & 97 & 12.0 & 45.0 \\
\midrule
\multicolumn{16}{l}{\textit{\textbf{Removed Text}}} \\
\midrule
gemini-2.5-pro & 34.5 & 11 & 115 & 0.0 & 31.4 & 147.3 & 13 & 461 & 0.0 & 111.4 & 124.2 & 45 & 241 & 0.0 & 118.9 \\
gemini-2.5-flash & 9.2 & 4 & 22 & 1.8 & 9.0 & 34.1 & 5 & 371 & 7.6 & 23.1 & 42.9 & 6 & 120 & 6.7 & 41.5 \\
qwen-2.5-omni & 29.1 & 10 & 49 & 8.2 & 29.0 & 53.8 & 11 & 133 & 23.7 & 49.2 & 89.3 & 22 & 228 & 30.2 & 80.4 \\
MiniCPM-o-2.6 & 19.0 & 6 & 53 & 5.3 & 18.2 & 29.8 & 11 & 58 & 8.7 & 28.6 & 70.6 & 34 & 114 & 13.8 & 68.3 \\
\bottomrule
\end{tabular}%
}
\end{table*}

\subsubsection{Myriad Echoes}
The full results for modality ablation on \textit{Myriad Echoes} are presented in Table~\ref{tab:ablation_me_combined}.

\paragraph{Analysis of Ablation Conditions.}
For this memory- and parsing-intensive task, the results show a more complex pattern. For the top-performing Gemini 2.5 Pro, removing any single modality leads to a severe drop in performance across all metrics (Success Rate, Mean Score, etc.), especially on higher difficulties. This underscores that its superhuman ability is contingent on successfully fusing information from the entire multi-modal stream. For other models, the performance is already very low in the baseline condition, making the impact of ablation less pronounced. However, we can observe that for weaker models, removing the audio or image modality can sometimes slightly reduce the `Parse Failure Rate', suggesting that a simpler set of inputs, even if incomplete, is less likely to confuse their parsing mechanisms.

% ===============================================================================
%          Table 10: Modality Ablation on ME (Optimized with resizebox)
% ===============================================================================
\begin{table*}[t!]
\centering
\caption{Full results for modality ablation on \textit{Myriad Echoes} across all difficulties.}
\label{tab:ablation_me_combined}
\resizebox{\textwidth}{!}{%
\small
\setlength{\tabcolsep}{4pt}
\sisetup{table-number-alignment = center}
\begin{tabular}{l *{3}{S[table-format=2] S[table-format=2.1] S[table-format=2.1] S[table-format=2.1] S[table-format=2]}}
\toprule
& \multicolumn{5}{c}{\textbf{Easy Difficulty}} & \multicolumn{5}{c}{\textbf{Medium Difficulty}} & \multicolumn{5}{c}{\textbf{Hard Difficulty}} \\
\cmidrule(lr){2-6} \cmidrule(lr){7-11} \cmidrule(lr){12-16}
\textbf{Model} & {Succ.(\%)} & {M.Score} & {Coord.} & {Icon} & {ParseF(\%)} & {Succ.(\%)} & {M.Score} & {Coord.} & {Icon} & {ParseF(\%)} & {Succ.(\%)} & {M.Score} & {Coord.} & {Icon} & {ParseF(\%)} \\
\midrule
\multicolumn{16}{l}{\textit{\textbf{Full Modality (Baseline)}}} \\
\midrule
MiniCPM-o-2.6 & 0 & 0.1 & 0.2 & 0.2 & 50 & 0 & 0.1 & 0.1 & 0.3 & 20 & 0 & 0 & 0 & 0 & 40 \\
gemini-2.5-flash & 0 & 0 & 0 & 0 & 0 & 0 & 0 & 0 & 0 & 0 & 0 & 1.9 & 4.5 & 1.6 & 0 \\
gemini-2.5-pro & 70 & 4.7 & 4.8 & 4.8 & 0 & 10 & 4 & 5.7 & 5.6 & 0 & 60 & 10.2 & 13.5 & 13.5 & 0 \\
qwen-2.5-omni & 0 & 0.1 & 0.1 & 0.3 & 60 & 0 & 0 & 0 & 0 & 50 & 0 & 0 & 0 & 0.1 & 60 \\
\midrule
\multicolumn{16}{l}{\textit{\textbf{Removed Audio}}} \\
\midrule
MiniCPM-o-2.6 & 0 & 0.3 & 0.4 & 0.2 & 50 & 0 & 0.1 & 0.1 & 0 & 20 & 0 & 0 & 0 & 0.1 & 40 \\
gemini-2.5-flash & 0 & 0 & 0 & 0 & 0 & 0 & 0 & 0 & 0 & 0 & 0 & 0.3 & 1.5 & 1.5 & 0 \\
gemini-2.5-pro & 70 & 4.7 & 5.8 & 5.8 & 0 & 0 & 1.9 & 3.4 & 3.4 & 0 & 40 & 8.8 & 10.8 & 10.8 & 0 \\
qwen-2.5-omni & 0 & 0.3 & 0.4 & 0.4 & 20 & 0 & 0 & 0 & 0.1 & 40 & 0 & 0.2 & 0.2 & 0.1 & 10 \\
\midrule
\multicolumn{16}{l}{\textit{\textbf{Removed Image}}} \\
\midrule
MiniCPM-o-2.6 & 0 & 0.3 & 0.3 & 0.1 & 60 & 0 & 0.2 & 0.2 & 0.1 & 30 & 0 & 0.1 & 0.1 & 0 & 20 \\
gemini-2.5-flash & 0 & 0.2 & 0.2 & 0 & 0 & 0 & 0.3 & 1 & 1 & 0 & 0 & 1.7 & 3.1 & 3.1 & 0 \\
gemini-2.5-pro & 30 & 3.2 & 3.7 & 3.7 & 0 & 30 & 5.1 & 7.7 & 7.7 & 0 & 50 & 9.1 & 10.5 & 10.5 & 0 \\
qwen-2.5-omni & 0 & 0.3 & 0.5 & 0.6 & 10 & 0 & 0 & 0 & 0.3 & 30 & 0 & 0.1 & 0.1 & 0.2 & 40 \\
\midrule
\multicolumn{16}{l}{\textit{\textbf{Removed Text}}} \\
\midrule
MiniCPM-o-2.6 & 0 & 0 & 0 & 0 & 60 & 0 & 0.1 & 0.1 & 0 & 40 & 0 & 0.1 & 0.4 & 0.4 & 0 \\
gemini-2.5-flash & 0 & 0 & 0 & 0 & 0 & 0 & 0.2 & 2 & 0 & 0 & 0 & 0.1 & 1.2 & 0 & 0 \\
gemini-2.5-pro & 0 & 0.6 & 3.6 & 3.6 & 0 & 0 & 1.1 & 9 & 9 & 0 & 0 & 1 & 15 & 15 & 0 \\
qwen-2.5-omni & 0 & 0 & 0 & 0 & 40 & 0 & 0.3 & 0.4 & 0.2 & 60 & 0 & 0 & 0 & 0 & 50 \\
\bottomrule
\end{tabular}%
}
\end{table*}

%--------------------------------------------------------
\subsection{Robustness to Noise}
%--------------------------------------------------------
We investigated the models' resilience to non-ideal sensory inputs by introducing noise into the visual and auditory modalities in the \textit{Phantom Soldiers in the Fog} environment (Medium difficulty). The full results are presented in Table~\ref{tab:noise_results}.

\begin{table*}[t!]
\small % To maintain consistency
\centering
\caption{Performance on \textit{Phantom Soldiers in the Fog} (Medium) under different noise conditions.}
\label{tab:noise_results}
% Use tabular* to create a table with the full text width
\begin{tabular*}{\textwidth}{@{\extracolsep{\fill}} l S[table-format=2.2] S[table-format=1.3] S[table-format=2.1] S[table-format=1.3] S[table-format=2.1] S[table-format=1.3]}
\toprule
& \multicolumn{2}{c}{\textbf{Baseline (No Noise)}} & \multicolumn{2}{c}{\textbf{Audio Noise}} & \multicolumn{2}{c}{\textbf{Image Noise}} \\
\cmidrule(lr){2-3} \cmidrule(lr){4-5} \cmidrule(lr){6-7}
\textbf{Model} & {Score} & {Succ. Rate} & {Score} & {Succ. Rate} & {Score} & {Succ. Rate} \\
\midrule
gemini-2.5-pro & 78.81 & 0.880 & 46.6 & 0.850 & 14.2 & 0.275 \\
gemini-2.5-flash & 31.20 & 0.570 & 20.5 & 0.483 & 2.9 & 0.050 \\
qwen-2.5-omni & 23.74 & 0.465 & 21.7 & 0.533 & 0 & 0 \\
MiniCPM-o-2.6 & 11.60 & 0.200 & 14.2 & 0.290 & 9.6 & 0.185 \\
\bottomrule
\end{tabular*}
\end{table*}

\begin{figure}[h!]
    \centering
    \includegraphics[width=\columnwidth]{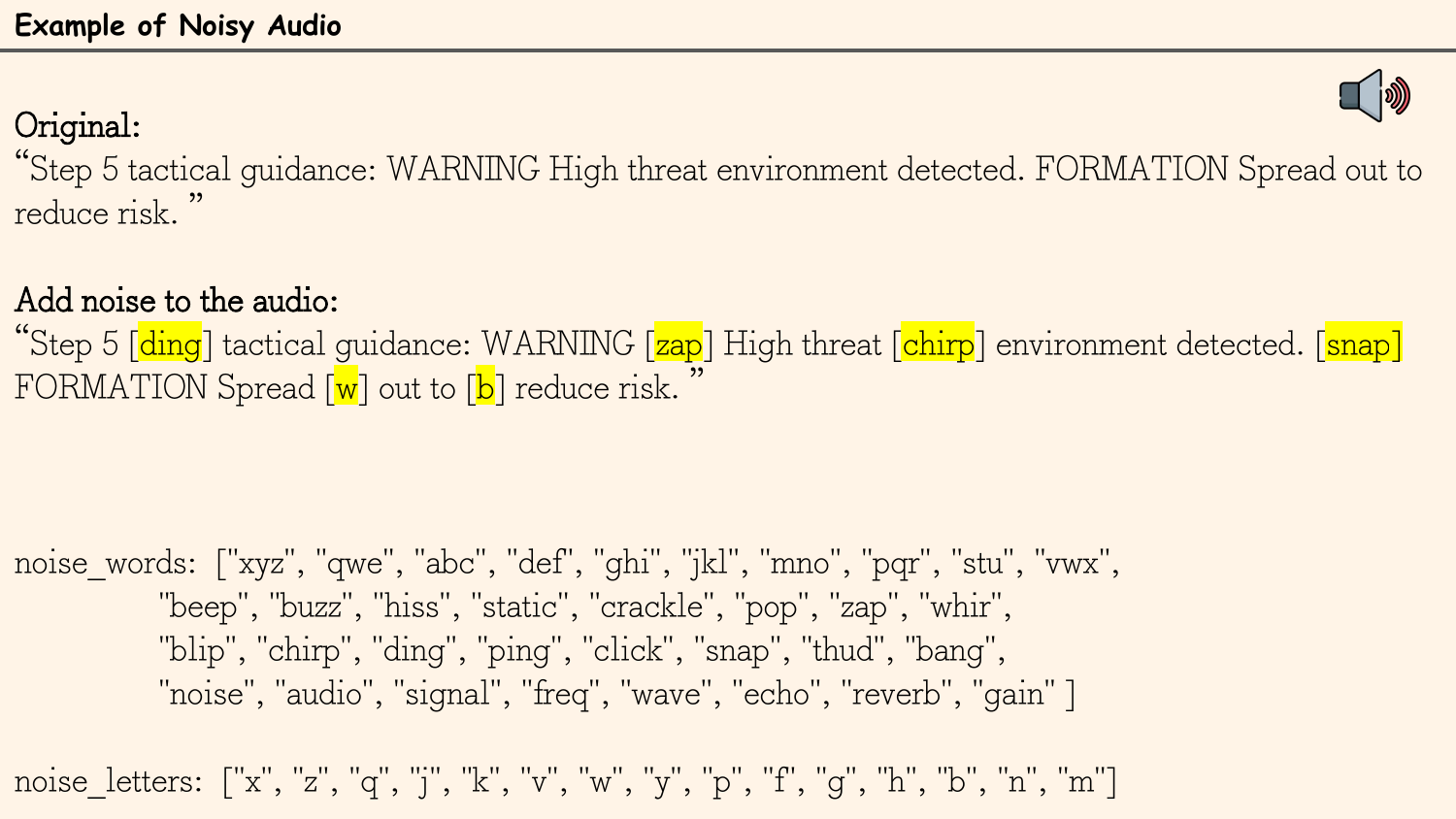}
    \caption{Illustration of audio noise injection via corruption of the source text in \textit{Phantom Soldiers in the Fog}.}
    \label{fig:noisy_audio}
\end{figure}

\paragraph{Audio Noise Injection and Analysis.}
For the audio modality, our goal was to simulate a noisy communication channel. We achieved this by corrupting the transcribed tactical guidance text before it was synthesized into speech. Specifically, we randomly inserted meaningless `noise words' (e.g., `zap', `chirp') and `noise letters' (e.g., `w', `b') into the original guidance sentences, as illustrated in Figure~\ref{fig:noisy_audio}.

The results show that while audio noise does degrade performance, most models exhibit a degree of resilience. For example, Gemini 2.5 Pro's score dropped from 78.81 to 46.6, a significant but not catastrophic decrease. Interestingly, MiniCPM-o-2\_6's performance slightly improved, which may suggest that its baseline audio processing is already suboptimal, and the noise did not significantly alter its behavior. Overall, the impact of this semantic-level audio noise was less severe than the visual noise.

\begin{figure}[h!]
    \centering
    \includegraphics[width=\columnwidth]{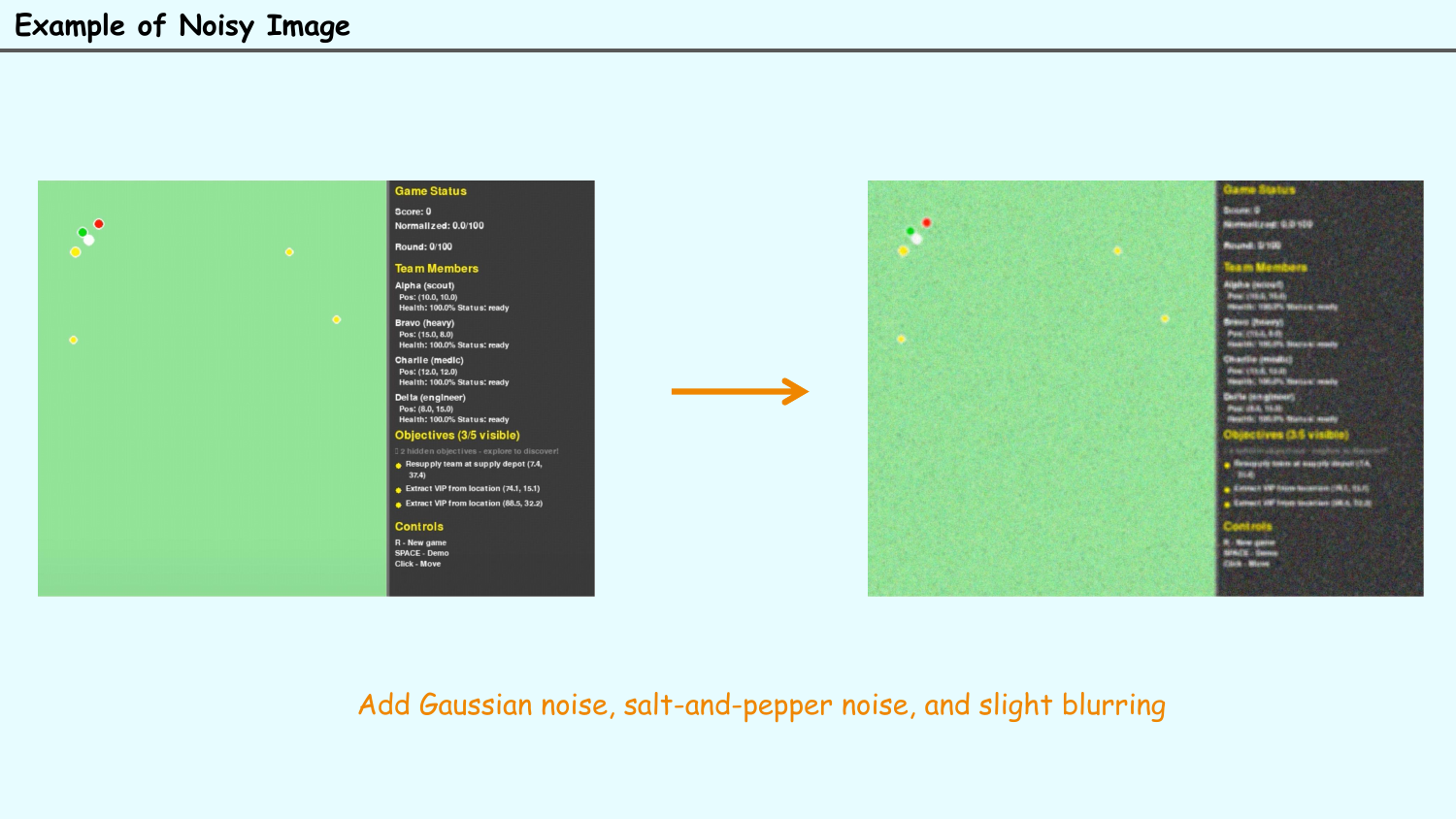}
    \caption{Illustration of visual noise injection in \textit{Phantom Soldiers in the Fog}.}
    \label{fig:noisy_image}
\end{figure}

\paragraph{Image Noise Injection and Analysis.}
For the visual modality, we simulated sensor degradation or poor visibility. We applied a combination of three common types of image noise to the top-down video feed: Gaussian noise, salt-and-pepper noise, and a slight blurring filter. The effect of this transformation is shown in Figure~\ref{fig:noisy_image}.

The impact of visual noise was far more severe and universally detrimental. As seen in Table~\ref{tab:noise_results}, all models experienced a catastrophic drop in performance. The top-performing Gemini 2.5 Pro was the most fragile, with its score plummeting by over 80\% from 78.81 to 14.2. Gemini 2.5 Flash's score dropped by nearly 90\%, and qwen-2.5-omni's performance collapsed entirely to zero. This extreme fragility strongly supports our main-text conclusion that the models may be learning superficial statistical correlations from visual data rather than robust, semantically-grounded representations. Their inability to handle minor pixel-level perturbations reveals a critical lack of generalization and robustness in their visual processing pipeline.
%--------------------------------------------------------
\subsection{Aided Reasoning via Prompting}
%--------------------------------------------------------
In contrast to penalizing models with noisy or conflicting data, this set of experiments investigates their ability to leverage helpful, explicit guidance provided directly within the prompt. We selected \textit{Myriad Echoes} and \textit{The Alchemist's Melody} for this analysis.

\begin{figure}[h!]
    \centering
    \includegraphics[width=\columnwidth]{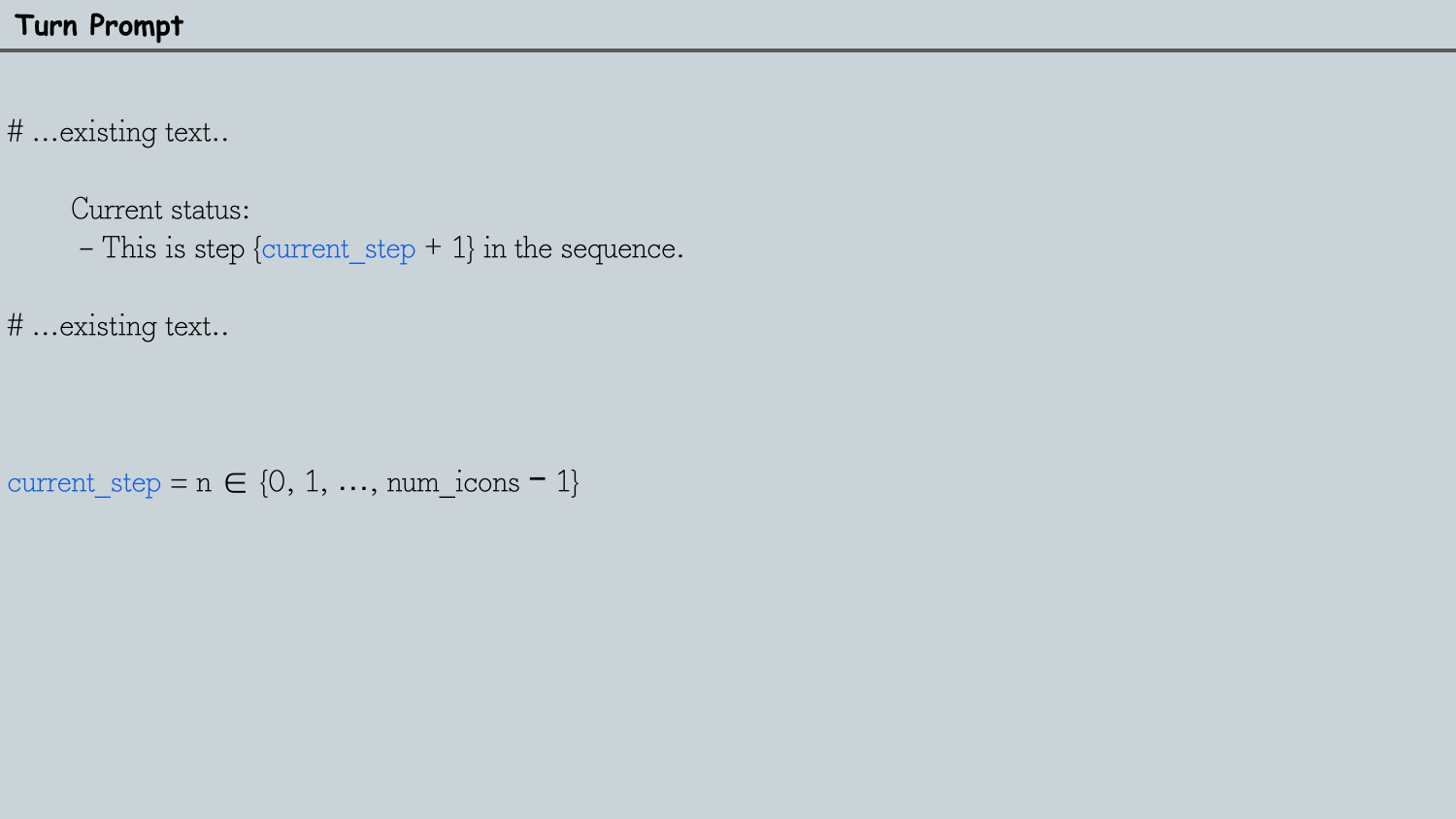}
    \caption{Augmented Turn Prompt for \textit{Myriad Echoes}, providing the agent with its current step in the sequence.}
    \label{fig:me_add_reasoning}
\end{figure}

\paragraph{Myriad Echoes.}
In the standard version of this task, the agent must internally track its progress through the execution sequence. In the `Aided Reasoning' version, we augmented the Turn Prompt for Phase 2 with an explicit status update, telling the agent which step of the sequence it was currently on, as shown in Figure~\ref{fig:me_add_reasoning}. The goal was to test if this explicit state information could reduce errors in long-sequence execution.

The results, presented in Table~\ref{tab:reasoning_me}, show that this hint significantly benefited the top-performing models. For instance, Gemini 2.5 Pro's success rate on the Hard difficulty increased from 60\% to 80\%, and its mean score improved. In contrast, weaker models were unable to effectively utilize this information, with their performance remaining largely unchanged.

% ===============================================================================
%          使用下面的代码替换你原来的 Table 12
% ===============================================================================

\begin{table*}[t!]
\centering
\caption{Performance comparison on \textit{Myriad Echoes} with and without aided reasoning prompts.}
\label{tab:reasoning_me}
\resizebox{\textwidth}{!}{%
\small
\setlength{\tabcolsep}{4pt}
\sisetup{table-number-alignment = center}
\begin{tabular}{l *{3}{S[table-format=2] S[table-format=2.1] S[table-format=2.1] S[table-format=2.1] S[table-format=2]}}
\toprule
& \multicolumn{5}{c}{\textbf{Easy Difficulty}} & \multicolumn{5}{c}{\textbf{Medium Difficulty}} & \multicolumn{5}{c}{\textbf{Hard Difficulty}} \\
\cmidrule(lr){2-6} \cmidrule(lr){7-11} \cmidrule(lr){12-16}
\textbf{Model} & {Succ.(\%)} & {M.Score} & {Coord.} & {Icon} & {ParseF(\%)} & {Succ.(\%)} & {M.Score} & {Coord.} & {Icon} & {ParseF(\%)} & {Succ.(\%)} & {M.Score} & {Coord.} & {Icon} & {ParseF(\%)} \\
\midrule
\multicolumn{16}{l}{\textit{\textbf{Baseline}}} \\
\midrule
gemini-2.5-pro & 70 & 4.7 & 4.8 & 4.8 & 0 & 10 & 4.0 & 5.7 & 5.6 & 0 & 60 & 10.2 & 13.5 & 13.5 & 0 \\
gemini-2.5-flash & 0 & 0 & 0 & 0 & 0 & 0 & 0 & 0 & 0 & 0 & 0 & 1.9 & 4.5 & 1.6 & 0 \\
qwen-2.5-omni & 0 & 0.1 & 0.1 & 0.3 & 60 & 0 & 0 & 0 & 0 & 50 & 0 & 0 & 0 & 0.1 & 60 \\
MiniCPM-o-2.6 & 0 & 0.1 & 0.2 & 0.2 & 50 & 0 & 0.1 & 0.1 & 0.3 & 20 & 0 & 0 & 0 & 0 & 40 \\
\midrule
\multicolumn{16}{l}{\textit{\textbf{With Aided Prompt}}} \\
\midrule
gemini-2.5-pro & 90 & 5.5 & 5.5 & 5.5 & 0 & 70 & 7.0 & 7.0 & 7.0 & 0 & 80 & 12.0 & 12.0 & 10.7 & 0 \\
gemini-2.5-flash & 0 & 0 & 0 & 0 & 0 & 10 & 1.0 & 1.0 & 1.0 & 0 & 0 & 1.0 & 1.0 & 1.5 & 0 \\
qwen-2.5-omni & 0 & 0.5 & 0.3 & 0 & 60 & 0 & 0.4 & 0.3 & 0.2 & 40 & 0 & 0 & 0 & 0 & 60 \\
MiniCPM-o-2.6 & 0 & 0.2 & 0.4 & 0.3 & 20 & 0 & 0.2 & 0.2 & 0.2 & 0 & 0 & 0 & 0 & 0 & 30 \\
\bottomrule
\end{tabular}%
}
\end{table*}

\begin{figure}[h!]
    \centering
    \includegraphics[width=\columnwidth]{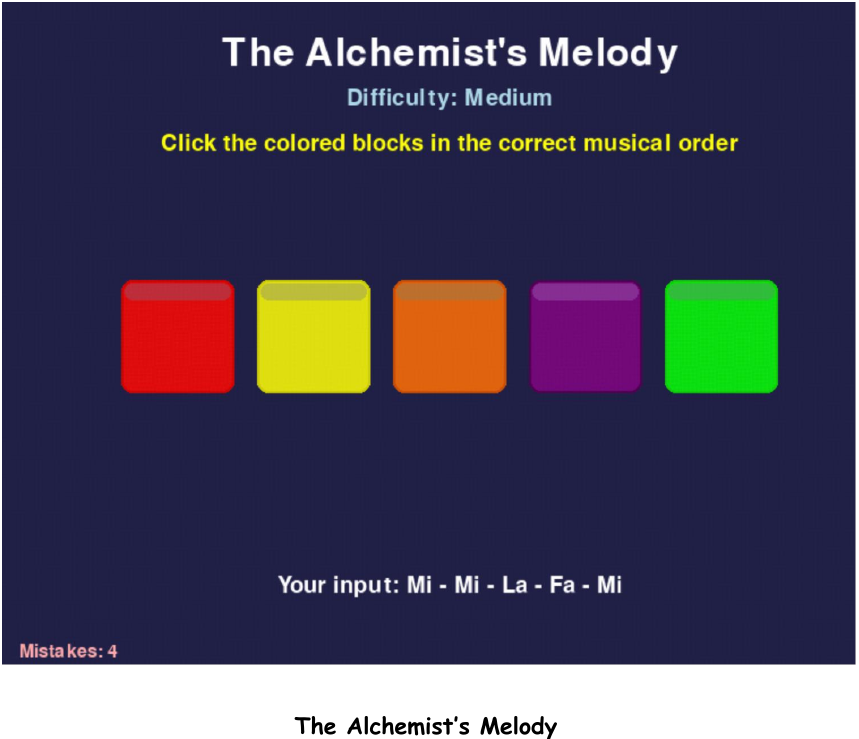}
    \caption{Augmented Turn Prompt for \textit{The Alchemist's Melody}, providing the agent with its learned color-note mapping.}
    \label{fig:am_add_reasoning}
\end{figure}

\paragraph{The Alchemist's Melody.}
This game requires the agent to deduce a color-note mapping. In the `Aided Reasoning' condition, we made this task significantly easier by directly providing the agent with its currently learned `color-note mapping' within the Turn Prompt, as shown in Figure~\ref{fig:am_add_reasoning}. This tests the agent's ability to directly apply provided knowledge rather than discover it.

The results in Table~\ref{tab:reasoning_am} are striking. The proprietary models, Gemini 2.5 Pro and Flash, demonstrated a remarkable ability to utilize this hint, with their completion rates jumping from 20\% and 0\% to a perfect 100\%. Their scores improved dramatically accordingly. In stark contrast, all tested open-source models failed to leverage this explicit information, showing no significant improvement in performance. This highlights a significant gap in advanced instruction-following and knowledge application capabilities between proprietary and open-source models.

\begin{table}[h!]
\small % To maintain consistency with other tables
\centering
\caption{Performance comparison on \textit{The Alchemist's Melody} with and without aided reasoning prompts.}
\label{tab:reasoning_am}
% Use tabular* to create a table with the full column width
\begin{tabular*}{\columnwidth}{@{\extracolsep{\fill}} l S[table-format=2.3] S[table-format=3, table-space-text-post=\%] S[table-format=2.3] S[table-format=3, table-space-text-post=\%]}
\toprule
& \multicolumn{2}{c}{\textbf{Baseline}} & \multicolumn{2}{c}{\textbf{With Aided Prompt}} \\
% Use cmidrule for partial horizontal lines under multicolumn headers
\cmidrule(lr){2-3} \cmidrule(lr){4-5}
\textbf{Model} & {Score} & {Comp. Rate} & {Score} & {Comp. Rate} \\
\midrule
gemini-2.5-pro      & 43.154 & 20\% & 73.104 & 100\% \\
gemini-2.5-flash    & 32.048 & 0\%  & 62.096 & 100\% \\
MiniCPM-o-2.6       & 30.294 & 0\%  & 32.798 & 0\% \\
VITA-1.5            & 20.010 & 0\%  & 18.896 & 0\% \\
Baichuan-Omni-1.5   & 31.823 & 0\%  & 33.548 & 0\% \\
qwen-2.5-omni       & 31.234 & 0\%  & 32.722 & 0\% \\
\bottomrule
\end{tabular*}
\end{table}
%--------------------------------------------------------
\subsection{Task Simplification}
%--------------------------------------------------------
To validate the designed complexity of our benchmark, we conducted a task simplification experiment on \textit{Myriad Echoes}. The goal was to understand if the primary difficulty lay in the multi-modal parsing phase or the long-sequence execution phase.

\paragraph{Methodology.}
We modified the original two-phase task into a single-phase perception task. In this simplified version, the agent still observes the full multi-modal sequence (video and audio) as in Phase 1. However, instead of proceeding to a second execution phase, the agent's task is to directly output the symbolic sequence it perceived. The performance is then measured by a final score which is a weighted average of the coordinate accuracy (50\%) and the icon accuracy (50\%) of its output.

\paragraph{Results.}
The results of this experiment are presented in Table~\ref{tab:simplification_me}, compared against the baseline performance on the original task. As expected, all models showed performance gains on the simplified task, as it removes the challenging long-horizon execution and action grounding components. However, even on this simplified perception-only task, the weaker open-source models still struggled to achieve high accuracy, particularly on the Hard difficulty. This confirms that the benchmark's core challenges are substantial and distributed across both its perception and action phases.

\begin{table*}[t!]
\centering
\caption{Performance comparison on \textit{Myriad Echoes} between the original task and the simplified (perception-only) task.}
\label{tab:simplification_me}
\resizebox{\textwidth}{!}{%
\small
\setlength{\tabcolsep}{4pt}
\sisetup{table-number-alignment = center}
% Column format is identical to Table 10 and 12 for consistency
\begin{tabular}{l *{3}{S[table-format=2] S[table-format=2.2] S[table-format=2.1] S[table-format=2.1] S[table-format=2]}}
\toprule
& \multicolumn{5}{c}{\textbf{Easy Difficulty}} & \multicolumn{5}{c}{\textbf{Medium Difficulty}} & \multicolumn{5}{c}{\textbf{Hard Difficulty}} \\
\cmidrule(lr){2-6} \cmidrule(lr){7-11} \cmidrule(lr){12-16}
\textbf{Model} & {Succ.(\%)} & {M.Score} & {Coord.} & {Icon} & {ParseF(\%)} & {Succ.(\%)} & {M.Score} & {Coord.} & {Icon} & {ParseF(\%)} & {Succ.(\%)} & {M.Score} & {Coord.} & {Icon} & {ParseF(\%)} \\
\midrule
\multicolumn{16}{l}{\textit{\textbf{Original Task (Baseline)}}} \\
\midrule
gemini-2.5-pro & 70 & 4.70 & 4.8 & 4.8 & 0 & 10 & 4.00 & 5.7 & 5.6 & 0 & 60 & 10.20 & 13.5 & 13.5 & 0 \\
gemini-2.5-flash & 0 & 0 & 0 & 0 & 0 & 0 & 0 & 0 & 0 & 0 & 0 & 1.90 & 4.5 & 1.6 & 0 \\
qwen-2.5-omni & 0 & 0.10 & 0.1 & 0.3 & 60 & 0 & 0 & 0 & 0 & 50 & 0 & 0 & 0 & 0.1 & 60 \\
MiniCPM-o-2.6 & 0 & 0.10 & 0.2 & 0.2 & 50 & 0 & 0.10 & 0.1 & 0.3 & 20 & 0 & 0 & 0 & 0 & 40 \\
\midrule
\multicolumn{16}{l}{\textit{\textbf{Simplified Task}}} \\
\midrule
gemini-2.5-pro & 90 & 5.60 & 5.6 & 5.6 & 0 & 60 & 6.00 & 6.0 & 6.0 & 0 & 70 & 10.60 & 10.6 & 10.6 & 0 \\
gemini-2.5-flash & 0 & 0 & 0 & 0 & 0 & 10 & 1.00 & 1.0 & 1.0 & 0 & 10 & 1.55 & 1.5 & 1.6 & 0 \\
qwen-2.5-omni & 0 & 0.25 & 0.1 & 0.4 & 50 & 0 & 0.20 & 0.2 & 0.2 & 50 & 0 & 0.05 & 0 & 0.1 & 70 \\
MiniCPM-o-2.6 & 0 & 0.10 & 0.1 & 0.1 & 30 & 0 & 0 & 0 & 0 & 10 & 0 & 0 & 0 & 0.4 & 10 \\
\bottomrule
\end{tabular}%
}
\end{table*}

%--------------------------------------------------------
\subsection{Modality Substitution}
%--------------------------------------------------------
This final diagnostic experiment investigates the models' ability to generalize across different modality representations of the same semantic information. Specifically, we tested if agents perform better when complex information is presented as structured text versus synthesized audio.

\paragraph{Methodology.}
We used the \textit{Phantom Soldiers in the Fog} environment (Medium difficulty) for this experiment. In the baseline condition, the agent receives tactical guidance via the audio channel (as Text-to-Speech). In the `Modality Substitution' condition, we disabled the audio channel entirely. Instead, the exact same structured textual guidance that would have been converted to speech was appended directly to the main text prompt. The agent's task was then to complete the mission using only the visual (video) and augmented textual modalities.

\paragraph{Results.}
The results, presented in Table~\ref{tab:substitution_ps}, reveal a strong and consistent trend. Most models, particularly the high-performing proprietary ones, showed a significant performance \textit{increase} when the auditory information was substituted with its textual equivalent. For example, Gemini 2.5 Pro's score improved from 78.81 to 86.78, and Gemini 2.5 Flash's score more than doubled from 31.2 to 70.2. This unexpected result reinforces our core finding about brittle fusion for non-textual information. It suggests that for current models, well-structured and unambiguous text is a more reliable and easier-to-process source of information than synthesized audio, even when the underlying semantic content is identical. The performance drop for MiniCPM-o-2\_6 is an anomaly that warrants further investigation, but may point to architectural differences in how it handles combined textual inputs.

\begin{table}[h!]
\small % To maintain consistency with other tables
\centering
\caption{Performance comparison on \textit{Phantom Soldiers in the Fog} (Medium) with and without Modality Substitution.}
\label{tab:substitution_ps}
% Use tabular* to create a table with the full column width
\begin{tabular*}{\columnwidth}{@{\extracolsep{\fill}} l S[table-format=2.2] S[table-format=1.3] S[table-format=2.2] S[table-format=1.3]}
\toprule
& \multicolumn{2}{c}{\textbf{Baseline}} & \multicolumn{2}{c}{\textbf{Substituted}} \\
% Key Change: Use a second row for sub-headers to handle long text
& \multicolumn{2}{c}{\small(Video+Text+Audio)} & \multicolumn{2}{c}{\small(Video+Text only)} \\
\cmidrule(lr){2-3} \cmidrule(lr){4-5}
\textbf{Model} & {Score} & {Succ. Rate} & {Score} & {Succ. Rate} \\
\midrule
gemini-2.5-pro & 78.81 & 0.880 & 86.78 & 0.920 \\
gemini-2.5-flash & 31.20 & 0.570 & 70.20 & 0.860 \\
MiniCPM-o-2.6 & 11.60 & 0.200 & 1.80 & 0.035 \\
qwen-2.5-omni & 23.74 & 0.465 & 25.90 & 0.470 \\
\bottomrule
\end{tabular*}
\end{table}

%%%%%%%%%%%%%%%%%%%%%%%%%%%%%%%%%%%%%%%%%%%%%%%%%%%%%%%%%
% Appendix G: Qualitative Case Studies
%%%%%%%%%%%%%%%%%%%%%%%%%%%%%%%%%%%%%%%%%%%%%%%%%%%%%%%%%
\section{Qualitative Case Studies}
\label{app:qualitative_cases}

To provide deeper, qualitative insights into the quantitative results presented in the main text, this section presents detailed case studies for three noteworthy phenomena observed during our evaluation.

%--------------------------------------------------------
\subsection{Case Study: Superhuman Memory in Myriad Echoes}
%--------------------------------------------------------

\paragraph{Phenomenon.}
As noted in the main text, top-tier models like Gemini 2.5 Pro exhibit superhuman performance on the \textit{Myriad Echoes} task, particularly on Hard difficulty where the sequence length is long. This case study analyzes the cognitive and architectural differences between the AI agent and a human player that lead to this performance gap.

\begin{figure}[h!]
    \centering
    \includegraphics[width=\columnwidth]{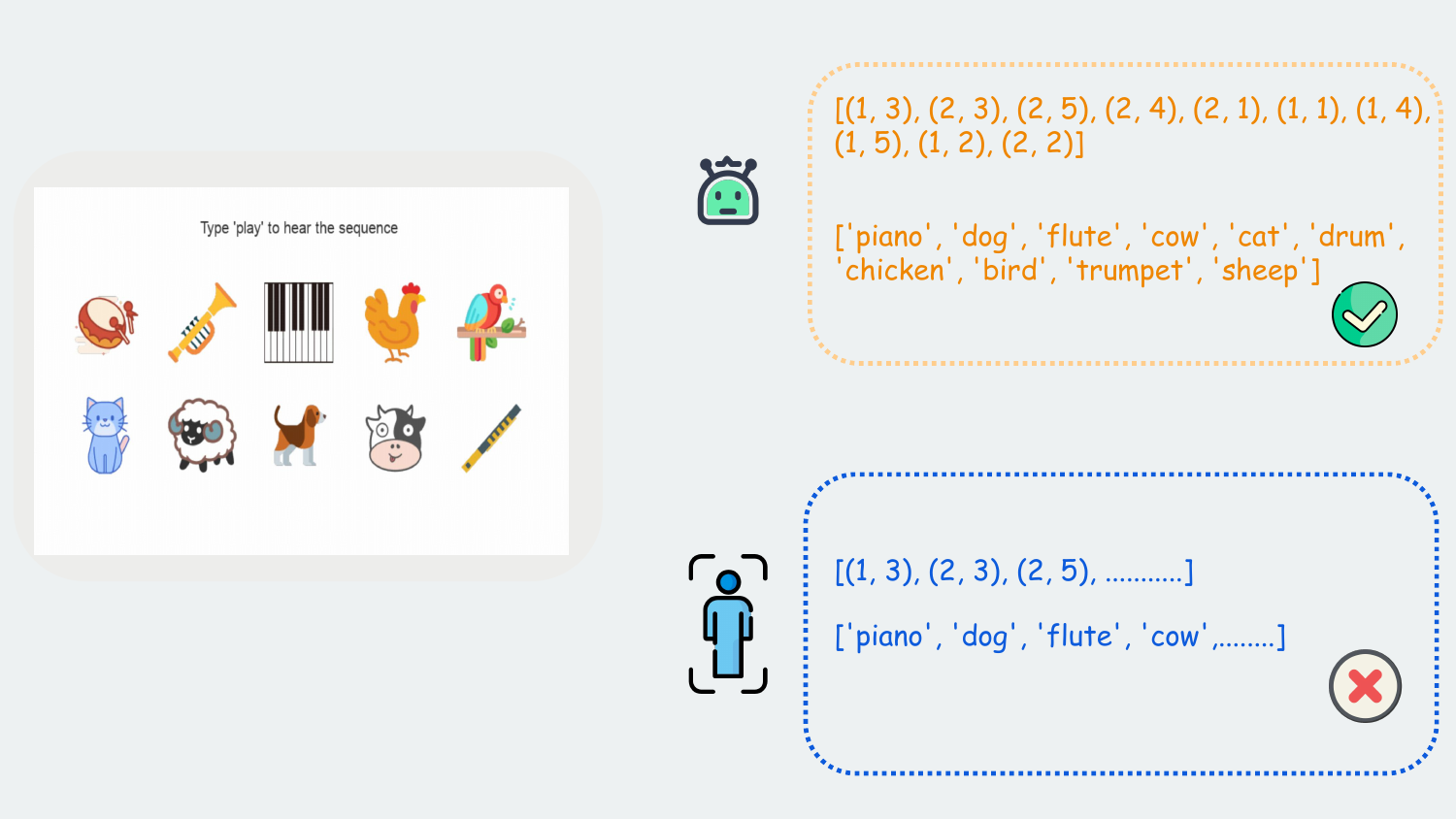}
    \caption{Illustration of the performance gap in \textit{Myriad Echoes}. The AI agent (top) can perfectly recall and transcribe the long, complex audio-visual sequence. The human player (bottom) is limited by their working memory and cannot reliably recall the full sequence.}
    \label{fig:case_superhuman}
\end{figure}

\paragraph{Analysis.}
The core challenge of \textit{Myriad Echoes} is twofold: high-bandwidth, cross-modal information encoding followed by precise, long-sequence symbolic execution. As illustrated in Figure~\ref{fig:case_superhuman}, the model is presented with a rapid, lengthy sequence of icon-sound pairs.
\begin{itemize}
    \item \textbf{Model's Advantage:} A large omni-modal model like Gemini 2.5 Pro functions as a near-perfect information transducer. Its vast parameter space and attention mechanisms allow it to faithfully transcribe the high-throughput audio-visual stream into a precise internal symbolic representation with minimal loss. In the execution phase, it can recall and act upon this long sequence with near-perfect accuracy, as its `working memory' is not biologically constrained.
    \item \textbf{Human's Limitation:} In contrast, a human player's performance is fundamentally limited by the capacity of their cognitive working memory (typically cited as 7±2 items). It is cognitively impossible for a human to perfectly memorize a rapid sequence of 10 or more arbitrary audio-visual pairs. Humans must resort to chunking or other heuristics, which are prone to error and forgetting, leading to a much lower performance ceiling.
\end{itemize}
\textbf{Conclusion:} This task highlights a domain where current AI excels: high-fidelity, short-term memory and precise symbolic manipulation. The observed superhuman performance is an expected outcome of the architectural differences between the model and the human brain, rather than an indication of superior general reasoning.

%--------------------------------------------------------
\subsection{Case Study: Anomalous Winning Strategy in Blasting Showdown}
%--------------------------------------------------------

\paragraph{Phenomenon.}
During the AI-vs-AI tournament in \textit{Blasting Showdown}, we observed cases where MiniCPM-o-2\_6 won matches despite having a Kill/Death (K/D) ratio of zero. This case study examines the unusual, passive strategy that led to this counter-intuitive success.

\begin{figure}[h!]
    \centering
    \includegraphics[width=\columnwidth]{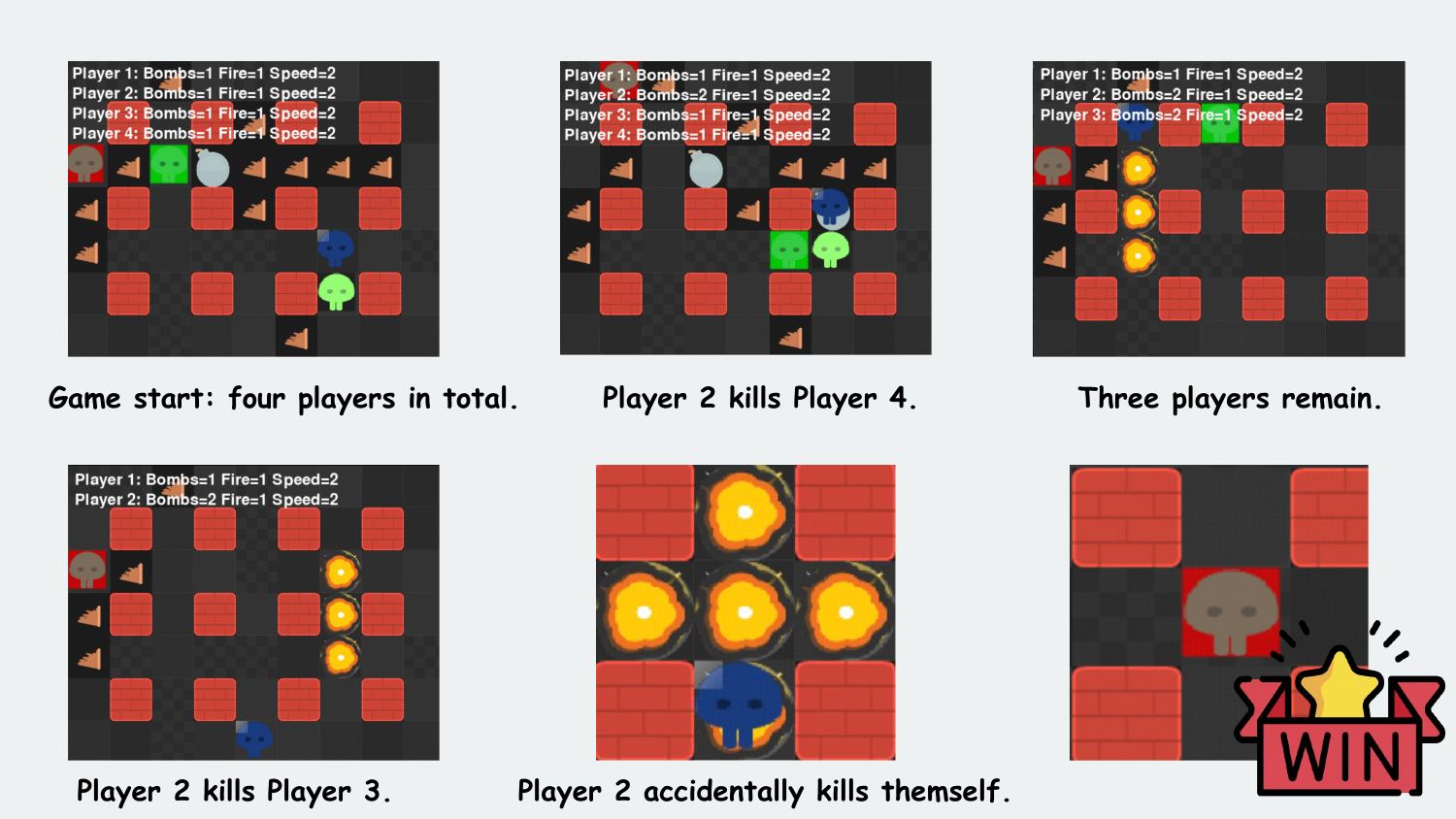}
    \caption{A step-by-step replay of a match won by MiniCPM-o-2\_6 (Player 1, red). Player 1 remains passive while the other agents eliminate each other through aggressive play and miscalculation, leading to an accidental victory.}
    \label{fig:case_passive_win}
\end{figure}

\paragraph{Analysis.}
The model's winning strategy can be characterized as extreme \textbf{risk aversion} or \textbf{passive survival}. Figure~\ref{fig:case_passive_win} depicts a representative match.
\begin{itemize}
    \item \textbf{Observed Behavior:} Throughout the match, MiniCPM-o-2\_6 (represented by the red player, Player 1) exhibited a very low tendency to place bombs or engage opponents. Its primary behavior consisted of reactive movements to evade bombs placed by other, more aggressive agents.
    \item \textbf{Environmental Dynamics:} The other three agents (Players 2, 3, and 4) actively engaged in combat. This created a chaotic and dangerous environment where players were eliminated not just by direct attacks, but also by chain reactions, self-elimination (getting trapped by their own bomb), or being caught in crossfire. In the depicted sequence, Player 2 eliminates Players 4 and 3, but then accidentally traps and eliminates itself.
    \item \textbf{Attribution:} It is unlikely that the model devised a sophisticated, deliberate strategy of `waiting out the storm'. A more plausible explanation is that its capacity for proactive, strategic planning is underdeveloped, causing it to default to the simplest possible policy: stay alive by avoiding immediate threats. In the chaotic context of a 4-player free-for-all, this simple, passive policy coincidentally proved to be highly effective.
\end{itemize}
\textbf{Conclusion:} This case study is a crucial reminder that in complex multi-agent systems, a successful outcome does not necessarily imply intelligent strategy. It highlights the importance of analyzing an agent's behavioral traces, not just its win rate, to accurately assess its planning and reasoning capabilities.

%--------------------------------------------------------
\subsection{Case Study: Systematic Failure in Myriad Echoes}
%--------------------------------------------------------

\paragraph{Phenomenon.}
A peculiar and consistent failure mode was observed for Gemini 2.5 Flash in the \textit{Myriad Echoes} task. On Easy and Medium difficulties, its performance on all sequence-related metrics was consistently zero.

\begin{figure}[h!]
    \centering
    \includegraphics[width=\columnwidth]{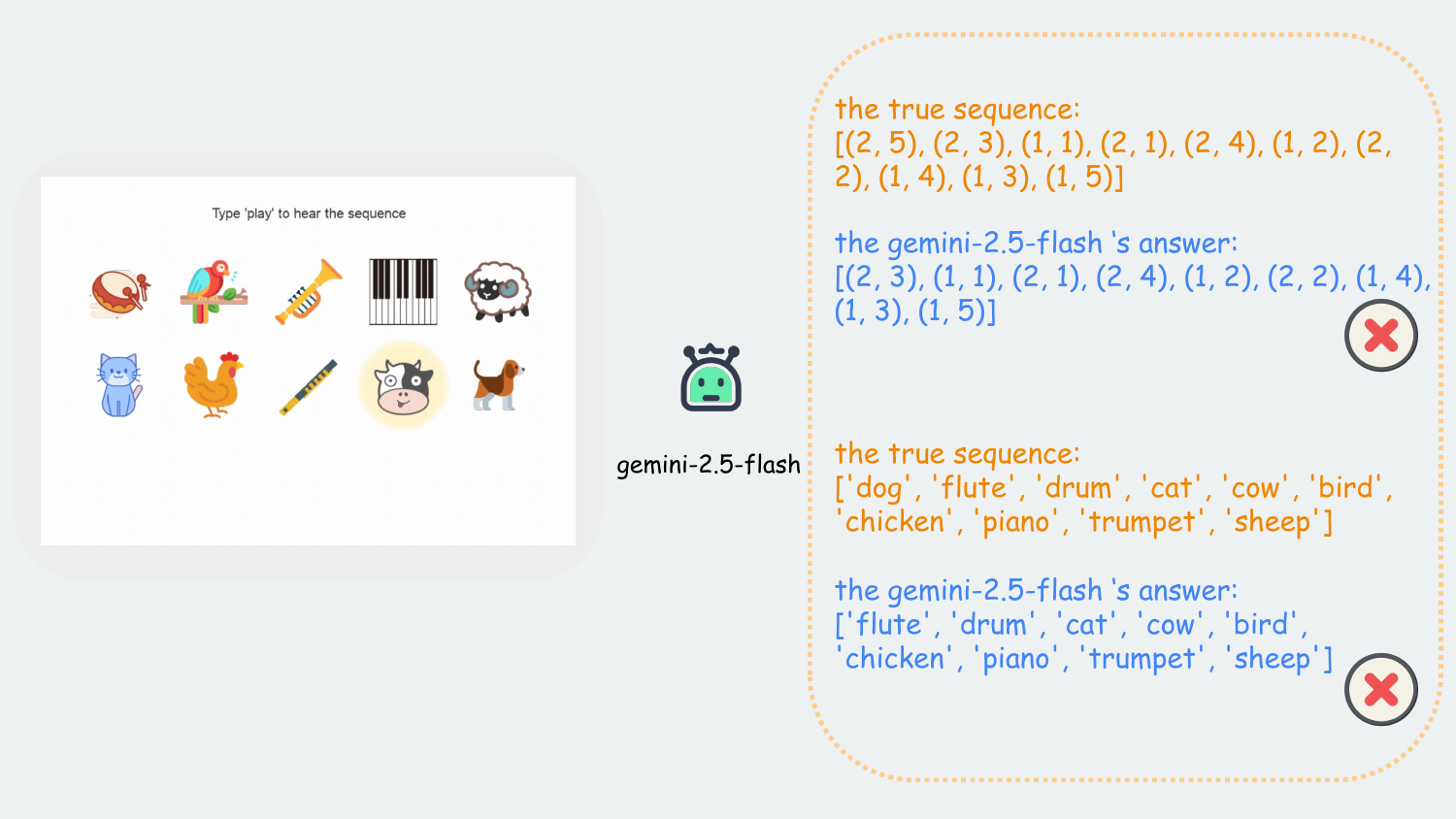}
    \caption{An example of Gemini 2.5 Flash's systematic `off-by-one' error. The model correctly identifies most of the sequence but consistently omits the first element, resulting in a complete task failure.}
    \label{fig:case_off_by_one}
\end{figure}

\paragraph{Analysis.}
The root cause of this total failure is a systematic \textbf{off-by-one error} in its sequence generation. Figure~\ref{fig:case_off_by_one} provides a clear example.
\begin{itemize}
    \item \textbf{The Task:} The agent is presented with a true sequence of a specific length (e.g., 10 items in the example).
    \item \textbf{The Error:} When prompted to reproduce the sequence, Gemini 2.5 Flash consistently outputs a sequence that is correct in content and relative order, but is missing the first element. It always generates a sequence of length N-1 when the correct length is N.
    \item \textbf{Attribution:} This behavior points to a subtle but critical flaw in how the model handles sequence boundaries or follows length constraints. It is a classic example of a \textbf{format following failure}. The model understands the core task of identifying the items, but fails on the crucial meta-task of adhering to the sequence's structural integrity (in this case, its length).
\end{itemize}
\textbf{Conclusion:} This case demonstrates that even highly capable models can harbor specific, systematic bugs in their reasoning or generation processes. It highlights the value of diagnostic benchmarks like OmniPlay, which can surface these otherwise hidden, granular failure modes that would be missed by evaluations that only measure average performance. For tasks requiring high precision, such a systematic error is a critical failure.

%====================================================================================
% APPENDIX H: DIAGNOSTIC EXPERIMENTS
%====================================================================================
\section{Detailed Statistical Results for Diagnostic Experiments}
\label{app:diagnostic_stats}

This section provides the full statistical data for the diagnostic experiments presented in Section~5.2, including the modality conflict and modality ablation studies. All experiments were conducted over N=50 independent runs to ensure statistical robustness.

\subsection{Modality Conflict (Supporting Figure~\ref{fig:conflict_results})}

Table~\ref{tab:appendix_conflict_stats} contains the detailed statistical results for the modality conflict experiment performed on \textit{Whispered Pathfinding (Hard)}. These results are visualized in Figure~\ref{fig:conflict_results} in the main text. We report the mean efficiency score, the standard deviation (SD) to show the raw performance volatility, and the standard error of the mean (SEM) used for generating the error margins in the figure.

\begin{table*}[h!]
\small % To maintain consistency
\centering
\caption{Full statistical results for the Modality Conflict experiment on \textit{Whispered Pathfinding (Hard)}. The data corresponds to Figure~\ref{fig:conflict_results}. All metrics are based on N=50 runs.}
\label{tab:appendix_conflict_stats}
% Use tabular* to create a table with the full text width
\begin{tabular*}{\textwidth}{@{\extracolsep{\fill}} l l S[table-format=2.1] S[table-format=2.1] S[table-format=1.1]}
\toprule
\textbf{Model} & \textbf{Condition} & {\textbf{Mean Score (\%)}} & {\textbf{SD}} & {\textbf{SEM}} \\
\midrule
\textbf{Gemini 2.5 Pro} & No Conflict      & 89.4 & 5.7 & 0.8 \\
                        & Audio Conflict   & 43.3 & 16.3 & 2.3 \\
                        & Text Conflict    & 32.2 & 17.7 & 2.5 \\
\midrule
\textbf{Gemini 2.5 Flash} & No Conflict      & 89.0 & 6.4 & 0.9 \\
                          & Audio Conflict   & 88.6 & 7.1 & 1.0 \\
                          & Text Conflict    & 48.1 & 20.5 & 2.9 \\
\midrule
\textbf{Qwen-2.5-Omni (7B)} & No Conflict      & 45.9 & 22.6 & 3.2 \\
                            & Audio Conflict   & 0.0 & 0.0 & 0.0 \\
                            & Text Conflict    & 37.6 & 24.0 & 3.4 \\
\midrule
\textbf{MiniCPM-o-2.6 (8B)} & No Conflict      & 59.5 & 19.1 & 2.7 \\
                            & Audio Conflict   & 32.2 & 24.7 & 3.5 \\
                            & Text Conflict    & 56.8 & 21.2 & 3.0 \\
\bottomrule
\end{tabular*}
\end{table*}

\subsection{Modality Ablation (Supporting Figure~\ref{fig:ablation_experiment})}

Table~\ref{tab:appendix_ablation_stats} provides the detailed statistical results for the modality ablation experiment, corresponding to Figure~\ref{fig:ablation_experiment} in the main text. The experiment was conducted on the 'Hard' difficulty for two distinct tasks: \textit{Whispered Pathfinding} and \textit{Myriad Echoes}. All metrics are based on N=50 independent runs.

\begin{table*}[h!]
\small % To maintain consistency
\centering
\caption{Full statistical results for the Modality Ablation experiment. The data corresponds to Figure~\ref{fig:ablation_experiment}.}
\label{tab:appendix_ablation_stats}
\sisetup{table-number-alignment = center}
\begin{tabular*}{\textwidth}{@{\extracolsep{\fill}} l l l S[table-format=2.2] S[table-format=2.2] S[table-format=1.2]}
\toprule
\textbf{Task} & \textbf{Model} & \textbf{Condition} & {\textbf{Mean Score}} & {\textbf{SD}} & {\textbf{SEM}} \\
\midrule
% --- Whispered Pathfinding Data ---
\multirow{16}{*}{\begin{tabular}[c]{@{}l@{}}\textit{Whispered Pathfinding}\\(Efficiency Score, \%)\end{tabular}} 
 & \textbf{Gemini 2.5 Pro} & Full Modality   & 86.7 & 8.5 & 1.2 \\
 &                         & Removed Audio   & 48.9 & 17.0 & 2.4 \\
 &                         & Removed Image   & 80.8 & 9.9 & 1.4 \\
 &                         & Removed Text    & 37.2 & 18.4 & 2.6 \\
\cmidrule(l){2-6}
 & \textbf{Gemini 2.5 Flash} & Full Modality   & 86.0 & 9.2 & 1.3 \\
 &                         & Removed Audio   & 79.4 & 12.0 & 1.7 \\
 &                         & Removed Image   & 55.5 & 17.7 & 2.5 \\
 &                         & Removed Text    & 83.5 & 10.6 & 1.5 \\
\cmidrule(l){2-6}
 & \textbf{Qwen-2.5-Omni (7B)} & Full Modality   & 31.6 & 21.2 & 3.0 \\
 &                         & Removed Audio   & 0.0 & 0.0 & 0.0 \\
 &                         & Removed Image   & 27.0 & 22.6 & 3.2 \\
 &                         & Removed Text    & 60.2 & 24.0 & 3.4 \\
\cmidrule(l){2-6}
 & \textbf{MiniCPM-o-2.6 (8B)} & Full Modality   & 48.8 & 25.5 & 3.6 \\
 &                         & Removed Audio   & 47.0 & 26.9 & 3.8 \\
 &                         & Removed Image   & 81.4 & 11.3 & 1.6 \\
 &                         & Removed Text    & 67.5 & 21.9 & 3.1 \\
\midrule
\midrule
% --- Myriad Echoes Data ---
\multirow{16}{*}{\begin{tabular}[c]{@{}l@{}}\textit{Myriad Echoes}\\(Weighted Score)\end{tabular}} 
 & \textbf{Gemini 2.5 Pro} & Full Modality   & 11.85 & 1.50 & 0.21 \\
 &                         & Removed Audio   & 9.80 & 2.12 & 0.30 \\
 &                         & Removed Image   & 9.80 & 2.12 & 0.30 \\
 &                         & Removed Text    & 8.00 & 2.47 & 0.35 \\
\cmidrule(l){2-6}
 & \textbf{Gemini 2.5 Flash} & Full Modality   & 2.48 & 3.18 & 0.45 \\
 &                         & Removed Audio   & 0.90 & 1.77 & 0.25 \\
 &                         & Removed Image   & 2.38 & 2.97 & 0.42 \\
 &                         & Removed Text    & 0.35 & 1.13 & 0.16 \\
\cmidrule(l){2-6}
 & \textbf{Qwen-2.5-Omni (7B)} & Full Modality   & 0.03 & 0.14 & 0.02 \\
 &                         & Removed Audio   & 0.18 & 0.42 & 0.06 \\
 &                         & Removed Image   & 0.13 & 0.35 & 0.05 \\
 &                         & Removed Text    & 0.00 & 0.00 & 0.00 \\
\cmidrule(l){2-6}
 & \textbf{MiniCPM-o-2.6 (8B)} & Full Modality   & 0.00 & 0.00 & 0.00 \\
 &                         & Removed Audio   & 0.03 & 0.14 & 0.02 \\
 &                         & Removed Image   & 0.08 & 0.28 & 0.04 \\
 &                         & Removed Text    & 0.40 & 0.99 & 0.14 \\
\bottomrule
\end{tabular*}
\end{table*}

%%%%%%%%%%%%%%%%%%%%%%%%%%%%%%%%%%%%%%%%%%%%%%%%%%%%%%%%%
% Appendix I: Full Performance Results
%%%%%%%%%%%%%%%%%%%%%%%%%%%%%%%%%%%%%%%%%%%%%%%%%%%%%%%%%
\section{Full Performance Results}
\label{sec:appendix_full_results}

This appendix provides the complete, unabridged performance data for all models and baselines across all tasks and difficulty levels from our main evaluation. We first present the summary statistics for all NPS-benchmarked tasks, which directly support the main findings in Section~\ref{sec:overall_performance}. Following this, we provide the detailed, task-specific raw metrics for each game environment.

%--------------------------------------------------------
\subsection{Summary of NPS-Benchmarked Tasks}
%--------------------------------------------------------
Table~\ref{tab:full_nps_results} provides the complete statistical data corresponding to the results visualized in Figure~\ref{fig:radar_chart} and summarized in Table~\ref{tab:dichotomy}. For each model and task, we conducted 50 independent runs with different random seeds. We report the Mean Normalized Performance Score (NPS), the Standard Deviation (SD) to show performance volatility, and the Standard Error of the Mean (SEM) to indicate the confidence in our estimation of the mean.

\begin{table*}[h!]
\centering
\caption{Full statistical results for all NPS-benchmarked tasks (N=50 runs). Data is presented as Mean NPS.}
\label{tab:full_nps_results}
\resizebox{\textwidth}{!}{%
\begin{tabular}{l l c c c}
\toprule
\textbf{Task (Grouped by Game)} & \textbf{Model} & \textbf{Mean NPS} & \textbf{Standard Deviation (SD)} & \textbf{Standard Error (SEM)} \\
\midrule
\textit{Whispered Pathfinding} (Easy) & Gemini 2.5 Pro & 98.2 & 4.2 & 0.6 \\
& Gemini 2.5 Flash & 95.9 & 7.1 & 1.0 \\
& Qwen-2.5-Omni & 66.6 & 21.2 & 3.0 \\
& MiniCPM-o-2.6 & 86.8 & 15.6 & 2.2 \\
& Baichuan-Omni-1.5 & 88.0 & 14.1 & 2.0 \\
& VITA-1.5 & 78.6 & 17.7 & 2.5 \\
\midrule
\textit{Whispered Pathfinding} (Medium) & Gemini 2.5 Pro & 99.2 & 3.5 & 0.5 \\
& Gemini 2.5 Flash & 95.7 & 8.5 & 1.2 \\
& Qwen-2.5-Omni & 78.6 & 19.8 & 2.8 \\
& MiniCPM-o-2.6 & 90.8 & 12.0 & 1.7 \\
& Baichuan-Omni-1.5 & 93.2 & 9.9 & 1.4 \\
& VITA-1.5 & 84.4 & 16.3 & 2.3 \\
\midrule
\textit{Whispered Pathfinding} (Hard) & Gemini 2.5 Pro & 95.2 & 8.5 & 1.2 \\
& Gemini 2.5 Flash & 95.0 & 9.2 & 1.3 \\
& Qwen-2.5-Omni & 75.6 & 23.3 & 3.3 \\
& MiniCPM-o-2.6 & 81.7 & 18.4 & 2.6 \\
& Baichuan-Omni-1.5 & 85.0 & 16.3 & 2.3 \\
& VITA-1.5 & 82.7 & 19.1 & 2.7 \\
\midrule
\textit{Myriad Echoes} (Easy) & Gemini 2.5 Pro & 114.1 & 12.0 & 1.7 \\
& Gemini 2.5 Flash & -7.7 & 31.8 & 4.5 \\
& Qwen-2.5-Omni & -3.8 & 36.1 & 5.1 \\
& MiniCPM-o-2.6 & -3.8 & 38.2 & 5.4 \\
& Baichuan-Omni-1.5 & -4.5 & 37.5 & 5.3 \\
& VITA-1.5 & -6.1 & 41.0 & 5.8 \\
\midrule
\textit{Myriad Echoes} (Medium) & Gemini 2.5 Pro & 157.0 & 18.4 & 2.6 \\
& Gemini 2.5 Flash & -2.5 & 33.2 & 4.7 \\
& Qwen-2.5-Omni & -2.5 & 39.6 & 5.6 \\
& MiniCPM-o-2.6 & 2.5 & 43.1 & 6.1 \\
& Baichuan-Omni-1.5 & 0.0 & 40.3 & 5.7 \\
& VITA-1.5 & -2.3 & 44.5 & 6.3 \\
\midrule
\textit{Myriad Echoes} (Hard) & Gemini 2.5 Pro & \textbf{399.2} & 25.5 & 3.6 \\
& Gemini 2.5 Flash & 81.4 & 29.7 & 4.2 \\
& Qwen-2.5-Omni & -1.7 & 34.6 & 4.9 \\
& MiniCPM-o-2.6 & -2.5 & 38.9 & 5.5 \\
& Baichuan-Omni-1.5 & -2.5 & 37.5 & 5.3 \\
& VITA-1.5 & -2.5 & 41.7 & 5.9 \\
\midrule
\textit{The Alchemist's Melody} (Default) & Gemini 2.5 Pro & 28.4 & 33.2 & 4.7 \\
& Gemini 2.5 Flash & 10.5 & 38.9 & 5.5 \\
& Qwen-2.5-Omni & 9.2 & 41.0 & 5.8 \\
& MiniCPM-o-2.6 & 7.7 & 43.1 & 6.1 \\
& Baichuan-Omni-1.5 & 10.2 & 39.6 & 5.6 \\
& VITA-1.5 & -8.9 & 48.1 & 6.8 \\
\midrule
\textit{Phantom Soldiers} (Easy) & Gemini 2.5 Pro & 88.6 & 19.8 & 2.8 \\
& Gemini 2.5 Flash & 86.5 & 22.6 & 3.2 \\
& Qwen-2.5-Omni & -25.6 & 45.2 & 6.4 \\
& MiniCPM-o-2.6 & -28.4 & 48.1 & 6.8 \\
& Baichuan-Omni-1.5 & 16.5 & 38.9 & 5.5 \\
& VITA-1.5 & -38.1 & 53.0 & 7.5 \\
\midrule
\textit{Phantom Soldiers} (Medium) & Gemini 2.5 Pro & 73.6 & 26.9 & 3.8 \\
& Gemini 2.5 Flash & 6.3 & 36.8 & 5.2 \\
& Qwen-2.5-Omni & -9.1 & 43.1 & 6.1 \\
& MiniCPM-o-2.6 & -42.2 & 50.9 & 7.2 \\
& Baichuan-Omni-1.5 & -35.4 & 49.5 & 7.0 \\
& VITA-1.5 & -69.3 & 35.4 & 5.0 \\
\midrule
\textit{Phantom Soldiers} (Hard) & Gemini 2.5 Pro & \textbf{87.5} & 24.7 & 3.5 \\
& Gemini 2.5 Flash & 54.5 & 33.2 & 4.7 \\
& Qwen-2.5-Omni & 11.2 & 41.0 & 5.8 \\
& MiniCPM-o-2.6 & -21.5 & 49.5 & 7.0 \\
& Baichuan-Omni-1.5 & 8.3 & 43.8 & 6.2 \\
& VITA-1.5 & \textbf{-49.2} & 42.4 & 6.0 \\
\bottomrule
\end{tabular}
}
\end{table*}

\clearpage % 确保表格不会和下面的内容挤在一起

% ===============================================================================
%          NEW TABLE: Human Expert Performance (Raw Scores)
% ===============================================================================
\begin{table*}[h!]
\small
\centering
\caption{Summary of Human Expert Performance (N=12 participants). Data is presented as the mean raw score used for NPS calculation, along with the standard deviation (SD) and standard error of the mean (SEM) measuring inter-player variability. Scoring rules are detailed in Appendix~\ref{app:exp_params_metrics}.}
\label{tab:human_raw_scores}
\sisetup{table-number-alignment = center}
\begin{tabular*}{\textwidth}{@{\extracolsep{\fill}} l S[table-format=3.2] S[table-format=2.2] S[table-format=1.2]}
\toprule
\textbf{Task (Grouped by Game)} & {\textbf{Mean Raw Score}} & {\textbf{SD (Inter-Player)}} & {\textbf{SEM (Inter-Player)}} \\
\midrule
% --- Whispered Pathfinding (Moderate Variance) ---
% Scores are 1/Trimmed_Steps. A small, unit-less value.
\textit{Whispered Pathfinding} (Easy) & 0.20 & 0.03 & 0.009 \\
\textit{Whispered Pathfinding} (Medium) & 0.13 & 0.02 & 0.006 \\
\textit{Whispered Pathfinding} (Hard) & 0.07 & 0.01 & 0.003 \\
\midrule
% --- Myriad Echoes (Low Variance) ---
% Scores are 0.5*M.Score + 0.25*Coord + 0.25*Icon
\textit{Myriad Echoes} (Easy) & 4.20 & 0.38 & 0.11 \\
\textit{Myriad Echoes} (Medium) & 3.10 & 0.28 & 0.08 \\
\textit{Myriad Echoes} (Hard) & 3.03 & 0.36 & 0.10 \\
\midrule
% --- Alchemist's Melody (Low Variance) ---
\textit{The Alchemist's Melody} (Default) & 87.66 & 6.14 & 1.77 \\
\midrule
% --- Phantom Soldiers (Moderate Variance) ---
% Scores are 0.5 * Normalized Score + 0.5 * (Success Rate * 100)
\textit{Phantom Soldiers} (Easy) & 100.00 & 4.50 & 1.30 \\
\textit{Phantom Soldiers} (Medium) & 98.80 & 9.39 & 2.71 \\
\textit{Phantom Soldiers} (Hard) & 96.75 & 12.58 & 3.63 \\
\bottomrule
\end{tabular*}
\end{table*}

\subsection{Statistical Analysis of Win Rates in Blasting Showdown}
%----------------------------------------------------------------
Table~\ref{tab:stats_bs_winrate} provides the statistical analysis for the win rates reported in Figure~4. The analysis is based on the outcomes of a 50-game tournament. We report the raw number of wins, the mean win rate (p), the standard deviation (SD) calculated as $\sqrt{p(1-p)}$, and the standard error of the mean (SEM) calculated as $SD/\sqrt{N}$, where N=50.

\begin{table*}[h!]
\small % To maintain consistency
\centering
\caption{Statistical analysis of win rates for the AI-vs-AI evaluation on \textit{Blasting Showdown} (N=50 games).}
\label{tab:stats_bs_winrate}
% Use tabular* to create a table with the full text width
\begin{tabular*}{\textwidth}{@{\extracolsep{\fill}} l c S[table-format=2.1, table-space-text-post=\%] S[table-format=1.3] S[table-format=1.1, table-space-text-post=\%]}
\toprule
\textbf{Model} & \textbf{Wins / Total} & {\textbf{Win Rate (\%)}} & {\textbf{SD}} & {\textbf{SEM (\%)}} \\
\midrule
Gemini 2.5 Pro      & 18 / 50 & 36.1\% & 0.480 & 6.8\% \\
Gemini 2.5 Flash    & 14 / 50 & 28.9\% & 0.453 & 6.4\% \\
MiniCPM-o-2.6       & 10 / 50 & 19.4\% & 0.395 & 5.6\% \\
Baichuan-Omni-1.5   & 9 / 50  & 17.7\% & 0.382 & 5.4\% \\
Qwen-2.5-Omni       & 6 / 50  & 11.8\% & 0.323 & 4.6\% \\
VITA-1.5            & 4 / 50  & 7.4\%  & 0.262 & 3.7\% \\
\bottomrule
\end{tabular*}
\end{table*}

%--------------------------------------------------------
\subsection{Task-Specific Raw Metrics: Whispered Pathfinding}
%--------------------------------------------------------
Table~\ref{tab:full_results_wp} presents the detailed performance metrics for the \textit{Whispered Pathfinding} task. The primary metric for this navigation task is `Mean Steps', where a lower value indicates better performance. We also report the trimmed mean, which is less sensitive to outliers.

% ===============================================================================
%          使用下面的代码替换你原来的 Table 18
% ===============================================================================

\begin{table*}[h!]
\centering
% Caption is outside resizebox to maintain a standard font size
\caption{Full performance results for \textit{Whispered Pathfinding} across all difficulties.}
\label{tab:full_results_wp}

% Force the table to fit within the text width
\resizebox{\textwidth}{!}{%
\footnotesize % Use a smaller font inside
\renewcommand{\arraystretch}{0.9} % Reduce row height
\setlength{\tabcolsep}{4pt} % Adjust column spacing for the original content
\sisetup{table-number-alignment = center}
\begin{tabular}{l *{3}{S[table-format=3.1] S[table-format=3] S[table-format=3] S[table-format=2.1] S[table-format=3.1]}}
\toprule
% Combined header row
\textbf{Model} & 
\multicolumn{5}{c}{\textbf{Easy Difficulty}} & 
\multicolumn{5}{c}{\textbf{Medium Difficulty}} & 
\multicolumn{5}{c}{\textbf{Hard Difficulty}} \\
\cmidrule(lr){2-6} \cmidrule(lr){7-11} \cmidrule(lr){12-16}
% Abbreviated sub-header row
& {Mean} & {Min} & {Max} & {Inv.} & {Trim.} & {Mean} & {Min} & {Max} & {Inv.} & {Trim.} & {Mean} & {Min} & {Max} & {Inv.} & {Trim.} \\
\midrule
human & 5.2 & 3 & 8 & 0.0 & 5.1 & 8.3 & 6 & 10 & 0.0 & 8.0 & 15.6 & 10 & 27 & 0.0 & 13.9 \\
\midrule
gemini-2.5-pro & 7.6 & 5 & 10 & 0.0 & 7.6 & 10.2 & 7 & 14 & 0.0 & 10.1 & 42.6 & 13 & 152 & 0.0 & 36.2 \\
gemini-2.5-flash & 16.1 & 4 & 70 & 0.8 & 10.9 & 23.2 & 6 & 87 & 1.2 & 19.0 & 43.5 & 18 & 112 & 2.7 & 37.15 \\
qwen-2.5-omni & 70.3 & 10 & 273 & 29.5 & 52.5 & 64.2 & 11 & 132 & 28.1 & 62.4 & 130.1 & 32 & 253 & 52.6 & 128.2 \\
MiniCPM-o-2.6 & 27.0 & 7 & 73 & 7.5 & 23.8 & 34.4 & 6 & 86 & 10.7 & 31.5 & 110.8 & 34 & 255 & 35.8 & 99.5 \\
VITA-1.5 & 36.1 & 13 & 70 & 15.0 & 35.5 & 52.8 & 8 & 162 & 21.9 & 47.8 & 106.5 & 23 & 343 & 35.7 & 94.8 \\
Baichuan-Omni-1.5 & 23.1 & 11 & 47 & 6.3 & 22.2 & 31.0 & 12 & 67 & 7.8 & 25.3 & 89.3 & 27 & 236 & 20.2 & 84.7 \\
\midrule
random & 193.2 & 25 & 500 & 0.0 & 147.0 & 277.8 & 125 & 477 & 0.0 & 262.3 & 413.4 & 119 & 500 & 0.0 & 482.7 \\
\bottomrule
\end{tabular}%
}
\end{table*}
%--------------------------------------------------------
\subsection{Task-Specific Raw Metrics: Myriad Echoes}
%--------------------------------------------------------
Table~\ref{tab:full_results_me} presents the detailed performance metrics for the \textit{Myriad Echoes} task. This task assesses both multi-modal parsing (Coord. Acc., Icon Acc.) and execution (Mean Score).

\begin{table*}[h!]
\centering
\caption{Full performance results for \textit{Myriad Echoes} across all difficulties.}
\label{tab:full_results_me}

% Force the table to fit within the text width
\resizebox{\textwidth}{!}{%
\footnotesize % Use a smaller font inside
\renewcommand{\arraystretch}{0.9} % Reduce row height
\setlength{\tabcolsep}{4pt} % Adjust column spacing
\sisetup{table-number-alignment = center}
\begin{tabular}{l *{3}{S[table-format=2] S[table-format=2.2] S[table-format=2.2] S[table-format=2.2] S[table-format=2]}}
\toprule
% Combined header row
\textbf{Model} & 
\multicolumn{5}{c}{\textbf{Easy Difficulty}} & 
\multicolumn{5}{c}{\textbf{Medium Difficulty}} & 
\multicolumn{5}{c}{\textbf{Hard Difficulty}} \\
\cmidrule(lr){2-6} \cmidrule(lr){7-11} \cmidrule(lr){12-16}
% Abbreviated sub-header row
& {Succ(\%)} & {M.Score} & {Coord.} & {Icon} & {ParseF(\%)} & {Succ(\%)} & {M.Score} & {Coord.} & {Icon} & {ParseF(\%)} & {Succ(\%)} & {M.Score} & {Coord.} & {Icon} & {ParseF(\%)} \\
\midrule
human & {-} & 3.70 & 3.60 & 5.80 & {-} & {-} & 2.50 & 2.60 & 4.80 & {-} & {-} & 2.60 & 2.30 & 4.60 & {-} \\
\midrule
gemini-2.5-pro & 70 & 4.70 & 4.80 & 4.80 & 0 & 10 & 4.00 & 5.70 & 5.60 & 0 & 60 & 10.20 & 13.50 & 13.50 & 0 \\
gemini-2.5-flash & 0 & 0 & 0 & 0 & 0 & 0 & 0 & 0 & 0 & 0 & 0 & 1.90 & 4.50 & 1.60 & 0 \\
qwen-2.5-omni & 0 & 0.10 & 0.10 & 0.30 & 60 & 0 & 0 & 0 & 0 & 50 & 0 & 0 & 0 & 0.10 & 60 \\
MiniCPM-o-2.6 & 0 & 0.10 & 0.20 & 0.20 & 50 & 0 & 0.10 & 0.10 & 0.30 & 20 & 0 & 0 & 0 & 0 & 40 \\
VITA-1.5 & 0 & 0.10 & 0.05 & 0 & 0 & 0 & 0 & 0 & 0.02 & 20 & 0 & 0 & 0 & 0 & 30 \\
Baichuan-Omni-1.5 & 0 & 0.20 & 0.10 & 0 & 50 & 0 & 0.10 & 0.10 & 0 & 70 & 0 & 0 & 0 & 0 & 60 \\
\midrule
random & 0 & 0.55 & 0.08 & 0.02 & 0 & 0 & 0.05 & 0.05 & 0.15 & 0 & 0 & 0.10 & 0.07 & 0.03 & 0 \\
\bottomrule
\end{tabular}%
}
\end{table*}

%--------------------------------------------------------
\subsection{Task-Specific Raw Metrics: The Alchemist's Melody}
%--------------------------------------------------------
Table~\ref{tab:full_results_am} presents the detailed performance metrics for the \textit{The Alchemist's Melody} task.

\begin{table*}[h!]
\small
\centering
\caption{Full performance results for \textit{The Alchemist's Melody}.}
\label{tab:full_results_am}
\begin{tabular*}{\textwidth}{@{\extracolsep{\fill}} l S[table-format=2.2] S[table-format=3, table-space-text-post=\%]}
\toprule
\textbf{Model} & {\textbf{Score}} & {\textbf{Completion Rate (\%)}} \\
\midrule
human & 87.66 & 100\% \\
\midrule
gemini-2.5-pro & 43.15 & 20\% \\
gemini-2.5-flash & 32.05 & 0\% \\
qwen-2.5-omni & 31.23 & 0\% \\
MiniCPM-o-2.6 & 30.29 & 0\% \\
VITA-1.5 & 20.01 & 0\% \\
Baichuan-Omni-1.5 & 31.82 & 0\% \\
\midrule
random & 25.51 & 0\% \\
\bottomrule
\end{tabular*}
\end{table*}

%--------------------------------------------------------
\subsection{Task-Specific Raw Metrics: Phantom Soldiers in the Fog}
%--------------------------------------------------------
Table~\ref{tab:full_results_ps} presents the detailed performance metrics for the \textit{Phantom Soldiers in the Fog} task across all difficulties.

\begin{table*}[h!]
\small
\centering
\caption{Full performance results for \textit{Phantom Soldiers in the Fog} across all difficulties.}
\label{tab:full_results_ps}
\begin{tabular*}{\textwidth}{@{\extracolsep{\fill}} l S[table-format=3.1] S[table-format=2.2] S[table-format=2.2] S[table-format=1.2] S[table-format=1.2] S[table-format=1.3]}
\toprule
& \multicolumn{3}{c}{\textbf{Score}} & \multicolumn{3}{c}{\textbf{Success Rate}} \\
\cmidrule(lr){2-4} \cmidrule(lr){5-7}
\textbf{Model} & {\textbf{Easy}} & {\textbf{Medium}} & {\textbf{Hard}} & {\textbf{Easy}} & {\textbf{Medium}} & {\textbf{Hard}} \\
\midrule
human & 100.0 & 99.60 & 98.50 & 1.00 & 0.98 & 0.950 \\
\midrule
gemini-2.5-pro & 83.51 & 78.81 & 91.62 & 1.00 & 0.88 & 0.857 \\
gemini-2.5-flash & 80.39 & 31.20 & 73.54 & 1.00 & 0.57 & 0.610 \\
qwen-2.5-omni & 5.13 & 23.74 & 23.34 & 0.13 & 0.465 & 0.550 \\
MiniCPM-o-2.6 & 3.10 & 11.60 & 8.93 & 0.11 & 0.20 & 0.270 \\
VITA-1.5 & 0 & 0 & 0 & 0 & 0 & 0 \\
Baichuan-Omni-1.5 & 29.15 & 11.50 & 19.60 & 0.50 & 0.28 & 0.550 \\
\midrule
random & 25.20 & 22.86 & 17.80 & 0.30 & 0.58 & 0.460 \\
\bottomrule
\end{tabular*}
\end{table*}

%--------------------------------------------------------
\subsection{Task-Specific Raw Metrics: Blasting Showdown}
%--------------------------------------------------------
Table~\ref{tab:full_results_bs} presents the full tournament results for the \textit{Blasting Showdown} task. As this is a competitive multi-agent environment, performance is measured by win rates and combat effectiveness metrics rather than a normalized score.

\begin{table*}[h!]
\small
\centering
\caption{Full tournament results for the AI-vs-AI evaluation on \textit{Blasting Showdown}.}
\label{tab:full_results_bs}
\begin{tabular*}{\textwidth}{@{\extracolsep{\fill}} l c c S[table-format=2.2, table-space-text-post=\%] c c S[table-format=1.2]}
\toprule
\textbf{Model} & \textbf{Games Played} & \textbf{Wins} & {\textbf{Win Rate (\%)}} & \textbf{Kills} & \textbf{Deaths} & {\textbf{K/D Ratio}} \\
\midrule
gemini-2.5-pro & 36 & 13 & 36.11\% & 93 & 39 & 2.38 \\
gemini-2.5-flash & 38 & 11 & 28.95\% & 68 & 41 & 1.66 \\
MiniCPM-o-2.6 & 31 & 6 & 19.35\% & 0 & 72 & 0.00 \\
Baichuan-Omni-1.5 & 34 & 6 & 17.65\% & 31 & 55 & 0.56 \\
qwen-2.5-omni & 34 & 4 & 11.76\% & 13 & 53 & 0.25 \\
VITA-1.5 & 27 & 2 & 7.41\% & 0 & 42 & 0.00 \\
\bottomrule
\end{tabular*}
\end{table*}
\end{document}